\pdfoutput=1
\documentclass[11pt,dvipsnames]{article}
\usepackage[utf8]{inputenc}

\def\noannotate{1}

\usepackage[margin=1.0in]{geometry}
\usepackage[english]{babel}
\usepackage{verbatim}
\usepackage{authblk}

  \date{}   %

\usepackage[citestyle=alphabetic, bibstyle=alphabetic, natbib=true, maxcitenames=2, maxbibnames=99, sorting=ynt]{biblatex}
\usepackage{csquotes}
\usepackage{titletoc}
\usepackage{longtable}

\usepackage{titlesec}
\usepackage[noend]{algpseudocode}
\usepackage{algorithm}

\usepackage{adjustbox}

\usepackage{afterpage}

\def\comments{0}
\newcommand{\kw}[1]{\textit{#1}}
\usepackage{standalone}

\usepackage[shortcuts]{extdash}

\newcommand{\colortext}[2]{{\color{#1} #2}}

\ifdefined\noannotate
  \newcommand{\revOne}[1]{#1}
  \newcommand{\revTwo}[1]{#1}
\else
  \newcommand{\revOne}[1]{#1}
  \newcommand{\revTwo}[1]{\colortext{ForestGreen}{{#1}}}
\fi

\usepackage{amsmath}

\DeclareMathOperator*{\argmin}{arg\,min}

\usepackage{amsfonts}  %
\usepackage{amssymb}   %
\usepackage{bbm}       %
\usepackage{mathrsfs}  %
\usepackage{relsize}   %

\usepackage{icomma}

\usepackage{mathtools} %
\usepackage{etoolbox}
\newcommand\swapifbranches[3]{#1{#3}{#2}}
\makeatletter
\MHInternalSyntaxOn
\patchcmd{\DeclarePairedDelimiter}{\@ifstar}{\swapifbranches\@ifstar}{}{}
\MHInternalSyntaxOff
\makeatother
\DeclarePairedDelimiter{\sbrack}{\lbrack}{\rbrack}

\DeclarePairedDelimiter{\ceil}{\lceil}{\rceil}
\DeclarePairedDelimiter{\abs}{\lvert}{\rvert}

\DeclarePairedDelimiter{\norm}{\lVert}{\rVert}
\usepackage{bm}
\DeclarePairedDelimiterX\myset[1]\lbrace\rbrace{#1}
\DeclarePairedDelimiterX\setbuild[2]\lbrace\rbrace{#1 \bm: #2}

\newcommand{\setint}[1]{{\sbrack{#1}}}
\newcommand{\powerSet}[1]{2^{#1}}
\newcommand{\powerSetInt}[1]{\powerSet{\setint{#1}}}

\newcommand{\func}[3]{{#1:#2\rightarrow#3}}
\newcommand{\defeq}{\coloneqq}
\newcommand{\fedeq}{\eqqcolon}

\newcommand{\expectS}[2]{\mathbb{E}_{#1}\sbrack{#2}}

\newcommand{\ind}[1]{{\mathbbm{1}\sbrack{#1}}}
\newcommand{\indBig}[1]{{\mathbbm{1}\big[ #1 \big]}}

\newcommand{\eqsmall}[1]{{\small #1}}

\newcommand{\real}{\mathbb{R}}
\newcommand{\realnn}{\real_{{\geq}0}}  %

\newcommand{\transpose}{^{\intercal}}

\newcommand{\bigO}[1]{{\mathcal{O}(#1)}}

\usepackage{array}  %
\usepackage{arydshln}  %
\usepackage{bigdelim}
\usepackage{booktabs}
\usepackage{multirow}
\usepackage{makecell}  %

\usepackage{amsthm}
\newtheoremstyle{remark}  %
{7pt}   %
{3pt}   %
{}   %
{}      %
{\bfseries} %
{:}     %
{.5em}  %
{}      %
\theoremstyle{remark}
\newtheorem{remark}{Remark}
\usepackage{tikz}

\usepackage{pgfplots}
\pgfplotsset{compat=1.13}
\usepgfplotslibrary{colorbrewer}
\usepgfplotslibrary{fillbetween}
\usepgfplotslibrary{statistics}
\usepackage{pgfplotstable}
\usepackage{graphicx} %
\graphicspath{{img/}} %

\usepackage{xifthen}
\newcommand{\ifempty}[3]{%
  \ifthenelse{\isempty{#1}}{#2}{#3}%
}

\newcommand{\dotprod}[2]{{#1\transpose \, #2}}

\newcolumntype{H}{>{\setbox0=\hbox\bgroup}c<{\egroup}@{}}

\newcommand\labelAndRemember[2]
  {\expandafter\gdef\csname labeled:#1\endcsname{#2}%
   \label{#1}#2}
\newcommand\recallLabel[1]
   {\csname labeled:#1\endcsname\tag{\ref{#1}}}

\newcommand{\citepos}[1]{\citeauthor{#1}'s \citep{#1}}

\newcommand{\teStr}{\text{te}}
\newcommand{\trainSet}{\mathcal{D}}
\newcommand{\trainSetAlt}{D}
\newcommand{\trainSetAltB}{\trainSetAlt'}
\newcommand{\trainSetAltPlusZi}{\trainSetAlt \cup \zI}
\newcommand{\nTr}{n}

\newcommand{\trainSetCore}{\trainSetAlt_{\textnormal{CS}}}

\newcommand{\testSet}{\trainSet_{\teStr}}

\newcommand{\trIdx}{i}
\newcommand{\trIdxAlt}{j}

\newcommand{\zeroVec}{\vec{0}}

\newcommand{\domainZ}{\mathcal{Z}}
\newcommand{\domainX}{\mathcal{X}}
\newcommand{\domainY}{\mathcal{Y}}

\newcommand{\hatFunc}[1]{\widehat{#1}}

\newcommand{\zSym}{z}
\newcommand{\zIBase}[1][\trIdx]{\zSym_{#1}}
\newcommand{\zI}{\zIBase}
\newcommand{\zIAlt}{\zIBase[\trIdxAlt]}
\newcommand{\zIOne}{\zIBase[\trIdx - 1]}
\newcommand{\zOne}{\zIBase[1]}
\newcommand{\zFin}{\zIBase[\nTr]}
\newcommand{\setZi}{\zI}  %
\newcommand{\X}{x}
\newcommand{\xI}[1][\trIdx]{\X_{#1}}
\newcommand{\Y}{y}

\newcommand{\yHat}{\hatFunc{\Y}}
\newcommand{\yHatI}[1][\trIdx]{\yHat_{#1}}
\newcommand{\yI}[1][\trIdx]{\Y_{#1}}
\newcommand{\dimX}{d}

\newcommand{\zTe}{\zSym_{\teStr}}
\newcommand{\xTe}{\X_{\teStr}}
\newcommand{\yTe}{\Y_{\teStr}}

\newcommand{\dec}{f}
\newcommand{\decFunc}[2]{{\dec( #1 ; #2)}}

\newcommand{\dsSub}[2]{{#1_{\trainSet^{\setminus #2}}}}
\newcommand{\trainSetNoZiBase}[1]{#1^{\setminus \setZi}}
\newcommand{\trainSetNoZi}{\trainSetNoZiBase{\trainSet}}

\newcommand{\W}{\theta}
\newcommand{\wT}[1][\itr]{\itrSup{\W}{#1}}
\newcommand{\wTOne}{\wT[\itr - 1]}
\newcommand{\wZero}{\wT[0]}
\newcommand{\wFin}{\wT[\nItr]}

\newcommand{\wOpt}{\W^{*}}

\newcommand{\wTSubBase}[2][\itr]{\wT[#1]_{#2}}
\newcommand{\wFinSubBase}[1]{\wTSubBase[\nItr]{#1}}
\newcommand{\wTSub}[2][\itr]{\dsSub{\wT[#1]}{#2}}
\newcommand{\wFinSub}{\wTSub[\nItr]{\zI}}

\newcommand{\dimW}{p}
\newcommand{\domainW}{\real^{\dimW}}

\newcommand{\batch}{\mathcal{B}}

\newcommand{\batchT}[1][\itr]{\itrSup{\batch}{#1}}
\newcommand{\batchOne}{\batchT[1]}
\newcommand{\batchFin}{\batchT[\nItr]}

\newcommand{\batchSizeT}{\abs{\batchT}}

\newcommand{\batchSizeSym}{b}
\newcommand{\batchSizeBase}[1]{\itrSup{\batchSizeSym}{#1}}
\newcommand{\batchSizeOne}{\batchSizeBase{1}}
\newcommand{\batchSizeTSym}{\batchSizeBase{\itr}}
\newcommand{\batchSizeFin}{\batchSizeBase{\nItr}}

\newcommand{\loss}{\ell}
\newcommand{\lFunc}[2]{{\loss ( #1 , #2 )}}
\newcommand{\lDecFunc}[3]{{\loss ( \decFunc{#1}{#2} , #3 )}}

\newcommand{\riskSym}{\mathcal{L}}
\newcommand{\risk}[2]{{\riskSym( #1 ; #2 )}}

\newcommand{\riskIFin}{\risk{\zI}{\wFin}}

\newcommand{\riskIT}{\risk{\zI}{\wT}}

\newcommand{\itr}{t}
\newcommand{\nItr}{T}

\newcommand{\itrSup}[2]{#1^{(#2)}}

\newcommand{\gradTe}{{\gradW \risk{\zTe}{\wTOne}}}
\newcommand{\gradI}{{\gradW \risk{\zI}{\wTOne}}}

\newcommand{\gradW}{\nabla_{\W}}
\newcommand{\gradWSq}{\gradW^{2}}

\newcommand{\hess}{H_{\W}}
\newcommand{\hessT}[1][\itr]{\itrSup{\hess}{#1}}
\newcommand{\hessFin}{\hessT[\nItr]}

\newcommand{\invHessT}{{(\hessT)^{-1}}}
\newcommand{\invHessFin}{{(\hessFin)^{-1}}}

\newcommand{\relatIfTerm}{\invHessFin \, \gradW \riskIFin}

\newcommand{\wdecay}{\lambda}  %
\newcommand{\lr}{\eta}

\newcommand{\lrT}[1][\itr]{\itrSup{\lr}{#1}}
\newcommand{\lrOne}{\lrT[1]}
\newcommand{\lrFin}{\lrT[\nItr]}

\newcommand{\trainParams}{\Theta}

\newcommand{\subsetItr}{\mathcal{T}}

\newcommand{\influence}{\mathcal{I}}
\newcommand{\baseInfFunc}[3]{{#1 \left( #2 , #3 \right)}}
\newcommand{\infFunc}{\baseInfFunc{\influence}}

\newcommand{\infEstSym}{\widehat{\influence}}
\newcommand{\infEstFunc}{\baseInfFunc{\infEstSym}}

\newcommand{\memName}{\textsc{Mem}}
\newcommand{\mem}[1]{{\memName(#1)}}

\newcommand{\infCook}{\influence_{\text{Cook}}}
\newcommand{\infCookFunc}[1]{{\infCook\mathopen{}\left( #1 \right)\mathclose{}}}

\newcommand{\infLoo}{\influence_{\textsc{LOO}}}
\newcommand{\infLooITe}{\baseInfFunc{\infLoo}{\zI}{\zTe}}

\newcommand{\infEstFeld}{\infEstSym_{\textsc{Down}}}

\newcommand{\infSV}{\influence_{\text{SV}}}
\newcommand{\infST}{\influence_{\text{ST}}}
\newcommand{\infKnnSV}{\influence_{\kNeigh\text{NN-SV}}}
\newcommand{\infBanzhaf}{\influence_{\text{Banzhaf}}}

\newcommand{\shapValSym}{\nu}
\newcommand{\shapValFunc}[1]{{\shapValSym( #1 )}}
\newcommand{\shapValComplete}[2]{{\mathcal{V}( #1 ; #2)}}

\newcommand{\shapCoalition}{A}

\newcommand{\infEstIF}{\infEstSym_{\textsc{IF}}}
\newcommand{\infEstRelatIF}{\infEstSym_{\text{\relatif}}}

\newcommand{\infFuncSTest}{s_{\text{test}}}

\newcommand{\infRepPt}{\influence_{\textsc{RP}}}
\newcommand{\infEstRepPt}{\infEstSym_{\textsc{RP}}}

\newcommand{\infTracInIdeal}{\influence_{\text{\tracin{}}}}
\newcommand{\infEstTracIn}{\infEstSym_{\text{\tracin{}}}}
\newcommand{\infEstTracInCP}{\infEstSym_{\text{\tracinCP}}}
\newcommand{\infEstGas}{\infEstSym_{\gas}}
\newcommand{\infEstRenormIF}{\infEstSym_{\text{RenormIF}}}

\newcommand{\infEstHydra}{\infEstSym_{\hydra}}

\newcommand{\feldIdx}{k}
\newcommand{\feldIdxAlt}{\feldIdx'}
\newcommand{\nFeldModel}{K}
\newcommand{\nFeldModelI}[1][\trIdx]{\nFeldModel_{#1}}

\newcommand{\feldDataset}{\trainSetAlt}
\newcommand{\feldDatasetK}[1][\feldIdx]{\feldDataset^{#1}}
\newcommand{\feldDatasetKAlt}{\feldDatasetK[\feldIdxAlt]}

\newcommand{\feldDatasetSize}{m}

\newcommand{\wFeldK}[1][\feldDatasetK]{\wFinSubBase{#1}}
\newcommand{\wFeldKAlt}{\wFeldK[\feldDatasetKAlt]}

\newcommand{\cScoreName}{C\=/score}
\newcommand{\trainDataDist}{\mathcal{P}}
\newcommand{\cTrSize}{\feldDatasetSize}
\newcommand{\consistencySym}{C}
\newcommand{\cProfile}[1][\zTe]{{\consistencySym_{\cTrSize,\trainSet}(#1)}}
\newcommand{\cScore}[1][\zTe]{{\consistencySym_{\trainSet}(#1)}}

\newcommand{\kNeigh}{k}
\newcommand{\neighAlt}[2]{{\text{Neigh}(#1;#2)}}

\newcommand{\simN}[1]{\stackrel{#1}{\sim}}
\newcommand{\simOne}{\sim}  %

\newcommand{\bigOpPln}{$\bigO{\nTr + \dimW}$}
\newcommand{\bigOn}{$\bigO{\nTr}$}
\newcommand{\bigOnSq}{$\bigO{\nTr^2}$}
\newcommand{\bigOnT}{$\bigO{\nTr\nItr}$}
\newcommand{\bigOnSqT}{$\bigO{\nTr^2 \nItr}$}
\newcommand{\bigOnP}{$\bigO{\nTr \dimW}$}

\newcommand{\bigOkT}{$\bigO{\nFeldModel\nItr}$}
\newcommand{\bigOk}{$\bigO{\nFeldModel}$}

\newcommand{\bigOT}{$\bigO{\nItr}$}

\newcommand{\bigOp}{$\bigO{\dimW}$}
\newcommand{\bigOpT}{$\bigO{\dimW\nItr}$}
\newcommand{\bigOnpT}{$\bigO{\nTr\dimW\nItr}$}
\newcommand{\bigOpTpn}{$\bigO{\dimW\nItr + \nTr\dimW}$}

\newcommand{\bigOnp}{$\bigO{\nTr\dimW}$}

\newcommand{\epsI}{\epsilon_{\trIdx}}

\newcommand{\wTEpsI}[1][\itr]{\wT[#1]_{+\epsI}}
\newcommand{\wTOneEpsI}{\wT[\itr - 1]_{+\epsI}}
\newcommand{\wFinEpsI}{\wFin_{+\epsI}}

\newcommand{\vertEpsIzero}{\Big\vert_{\epsI = 0}}

\newcommand{\kernel}{\mathcal{K}}
\newcommand{\kernelFunc}[3]{{\kernel(#1, #2,#3)}}
\newcommand{\kernelITe}{{\kernel(\xI, \xTe, \rpValI)\teDimVert}}

\newcommand{\rpValSym}{\alpha}
\newcommand{\rpValI}{\rpValSym_{\trIdx}}

\newcommand{\teDimVert}{\,\vert_{\yTe}}
\newcommand{\rpLossDerivI}{\frac{\partial \lFunc{\yHatI}{\yI}}{\partial \yHat}}

\newcommand{\featSym}{\mathbf{f}}

\newcommand{\wFinBegin}{\itrSup{\dot{\W}}{\nItr}}
\newcommand{\wFinLast}{\itrSup{\ddot{\W}}{\nItr}}

\newcommand{\featI}[1][\trIdx]{\featSym_{#1}}
\newcommand{\featTe}{\featI[\teStr]}

\newcommand{\regularizerSym}{r}
\newcommand{\regFunc}[1]{{\regularizerSym ( #1 )}}

\newcommand{\lOne}{$L_1$}
\newcommand{\lTwo}{$L_2$}

\newcommand{\hyperGradTI}[1][\itr]{\widetilde{h}^{(#1)}_{\trIdx}}
\newcommand{\hyperGradZeroI}{\hyperGradTI[0]}
\newcommand{\hyperGradTOneI}{\hyperGradTI[\itr - 1]}
\newcommand{\hyperGradFinI}{\hyperGradTI[\nItr]}

\newcommand{\gradT}[1][\itr]{\itrSup{g}{#1}}
\newcommand{\gradTOne}{\gradT[\itr - 1]}

\newcommand{\riskBatchT}{\risk{\batchT}{\wTOne}}
\newcommand{\riskBatchTOne}{\risk{\batchT}{\wTOne}}

\newcommand{\riskBatchTOneEpsI}{\risk{\batchT}{\wTOneEpsI}}

\newcommand{\riskBatchFinOneEpsI}{\risk{\batchFin}{\wTEpsI[\nItr - 1]}}

\newcommand{\knn}{$\kNeigh$NN}
\newcommand{\loo}{LOO}
\newcommand{\knnLOO}{\knn{}~\loo{}}
\newcommand{\leafRefit}{\textsc{LeafRefit}}

\newcommand{\sv}{SV}
\newcommand{\gShap}{G\=/Shapley}
\newcommand{\mcShap}{TMC\=/Shapley}
\newcommand{\knnShap}{\knn{}~Shapley}
\newcommand{\betaShap}{Beta Shapley}

\newcommand{\ame}{AME}

\newcommand{\SHAP}{SHAP}

\newcommand{\relatif}{RelatIF}
\newcommand{\fastif}{\textsc{FastIF}}
\newcommand{\leafInfluence}{\textsc{LeafInfluence}}

\newcommand{\rpLocalJacobian}{RPS\=/LJE}
\newcommand{\trex}{TREX}

\newcommand{\tracin}{TracIn}
\newcommand{\tracinCP}{\tracin{}CP}
\newcommand{\tracinAD}{\tracin{}AD}
\newcommand{\tracinWE}{\tracin{}WE}
\newcommand{\vaeTracIn}{VAE\=/\tracin{}}
\newcommand{\gas}{\textsc{GAS}}
\newcommand{\boostin}{BoostIn}

\newcommand{\tracinLast}{\tracin{}\=/Last}

\newcommand{\tracinRP}{\tracin{}RP}
\newcommand{\tracinRpDim}{d}

\newcommand{\hydra}{\textsc{HyDRA}}

\newcommand{\feldman}{\textsc{Downsampling}}
\newcommand{\feldmanShort}{\textsc{Downsamp}.}

\ifnum\comments=1
    \newcommand{\zayd}[1]{\marginpar{\tiny\color{orange}{ZH: #1}}}
    \newcommand{\daniel}[1]{\marginpar{\tiny\color{blue}{DL: #1}}}
\else
    \newcommand{\zayd}[1]{}
    \newcommand{\daniel}[1]{}
\fi

\newcommand{\SupplementaryMaterialsTitle}{%
  \vbox{
    \hrule height 4pt
    \vskip 0.05in
    \vskip -\parskip%
    \begin{center}
      {\Large\bf \titleTextBreak{} \par}

      \vspace{8pt}
      {\large Supplemental Materials \par}
    \end{center}
    \vskip 0.15in
    \vskip -\parskip
    \hrule height 1pt
    \vskip 0.09in%
  }
}

\newcommand{\AlgFontSize}{\footnotesize}
\newcommand{\EqFontSize}{\small}
\newcommand{\nomenclatureTableFontSize}{\small}

\newcommand{\algInputDelim}{;}

\usepackage{setspace}
\newcommand{\algSetStretch}{\setstretch{1.22}}

\newcommand{\algMiniPageWidth}{0.72\textwidth}

\newcommand{\algcomment}[1]{\hfill$\triangleright$~#1}

\providecommand{\keywords}[1]
{
  \small
  \noindent
  \textbf{\textit{Keywords}}:~#1
}

\newcommand{\relatedHeading}[1]{%
  \noindent%
  \revOne{%
    \textbf{#1}\hspace{6pt}%
  }%
}
\usepackage{forest}
\useforestlibrary{edges}

\colorlet{linecol}{black!75}

\forestset{
  nice empty nodes/.style={
      for tree={},
      delay={where content={}{shape=coordinate,for siblings={anchor=north}}{}}
  },
}

\newcommand{\linRegCleanLineColor}{ForestGreen}
\newcommand{\linRegAllLineColor}{BrickRed}

\usepackage[colorlinks=false]{hyperref}

  \newcommand{\titleText}{Training Data Influence Analysis and Estimation: A Survey}
\newcommand{\titleTextBreak}{Training Data Influence Analysis \\ and Estimation: A Survey}
\newcommand{\pdfKeywords}{%
  Influence analysis,
  influence estimation,
  training data attribution,
  data valuation,
  influence functions,
  \tracin{},
  Shapley value
}

\RequirePackage{graphicx, xcolor}
\definecolor[named]{ACMBlue}{cmyk}{1,0.1,0,0.1}
\definecolor[named]{ACMYellow}{cmyk}{0,0.16,1,0}
\definecolor[named]{ACMOrange}{cmyk}{0,0.42,1,0.01}
\definecolor[named]{ACMRed}{cmyk}{0,0.90,0.86,0}
\definecolor[named]{ACMLightBlue}{cmyk}{0.49,0.01,0,0}
\definecolor[named]{ACMGreen}{cmyk}{0.20,0,1,0.19}
\definecolor[named]{ACMPurple}{cmyk}{0.55,1,0,0.15}
\definecolor[named]{ACMDarkBlue}{cmyk}{1,0.58,0,0.21}
\hypersetup{colorlinks,
  linkcolor=ACMRed,
  citecolor=ACMPurple,
  urlcolor=ACMDarkBlue,
  filecolor=ACMDarkBlue,
  pdftitle={\titleText},
  pdfauthor={Zayd Hammoudeh and Daniel Lowd},
  pdfkeywords={\pdfKeywords},
}

\title{\textbf{\titleTextBreak}}

\author[1,2]{Zayd Hammoudeh%
    \footnote{%
      Correspondence to \href{mailto:zayd@cs.uoregon.edu}{zayd@cs.uoregon.edu}.
      Work primarily done while at the University of Oregon.
      This paper is published in journal \href{https://link.springer.com/article/10.1007/s10994-023-06495-7}{\textit{Machine Learning}}~\citep{Hammoudeh:2024:InfluenceSurvey}.%
    }%
}%
\author[1]{Daniel Lowd}
\affil[1]{University of Oregon}
\affil[2]{Qualtrics AI}

\makeatletter%
\begin{document}
\maketitle

\begin{abstract}
Good models require good training data.
For overparameterized deep models, the causal relationship between training data and model predictions is increasingly opaque and poorly understood.
Influence analysis partially demystifies training's underlying interactions by quantifying the amount each training instance alters the final model.
Measuring the training data's influence exactly can be provably hard in the worst case;
this has led to the development and use of influence estimators, which only approximate the true influence.
This paper provides the first comprehensive survey of training data influence analysis and estimation.
We begin by formalizing the various, and in places orthogonal, definitions of training data influence.
We then organize state-of-the-art influence analysis methods into a taxonomy;
we describe each of these methods in detail and compare their underlying assumptions, asymptotic complexities, and overall strengths and weaknesses.
Finally, we propose future research directions to make influence analysis more useful in practice as well as more theoretically and empirically sound.
A curated, up-to-date list of resources related to influence analysis is available at \url{https://github.com/ZaydH/influence_analysis_papers}.
 \end{abstract}


\section{Introduction}\label{sec:Intro}

Machine learning is built on training data~\citep{Reddi:2021:DataEngineering}.
Without good training data, nothing else works.
How modern models learn from and use training data is increasingly opaque~\citep{Koh:2017:Understanding,Zhang:2021:UnderstandingGeneralizationStillRequires,Xiang:2022:IncreasinglyUnexplainableAI}.
Regarding state-of-the-art black-box models, \citet{Yampolskiy:2020:UnexplainableAI} notes, ``If all we have is a `black box' it is impossible to understand causes of failure and improve system safety.''

Large modern models require tremendous amounts of training data~\citep{Black:2021:LeaveOneOut}.
Today's uncurated, internet-derived datasets
commonly contain numerous anomalous instances~\citep{Pleiss:2020:IdentifyingMislabeledData}.
These anomalies can arise from multiple potential sources.
For example, training data anomalies may have a natural cause such as
distribution shift~\citep{Rousseeuw:1997:RobustRegressionOutlierDetection,Yang:2021:OodSurvey},
measurement error,
or
non-representative samples drawn from the tail of the data distribution
\citep{Huber:1981:RobustStatistics,Feldman:2020:LearningRequiresMemorization,Jiang:2021:CScore}.
Anomalous training instances also occur due to human or algorithmic labeling errors -- even on well-known, highly-curated datasets~\citep{Ekambaram:17:LabelNoiseDatasets}.
Malicious adversaries can insert anomalous poison instances into the training data with the goal of manipulating specific model predictions~\citep{Biggio:2012:Poisoning,Chen:2017:Targeted,Shafahi:2018:PoisonFrogs,Hammoudeh:2023:CertifiedRegression}.
Regardless of the cause, anomalous training instances degrade a model's overall generalization performance.

Today's large datasets
also generally overrepresent established and dominant viewpoints~\citep{Bender:2021:StochasticParrots}.
Models trained on these huge public datasets encode and exhibit biases based on protected characteristics, including gender, race, religion, and disability~\citep{Basta:2019:GenderBiasWordEmbeddings,Kurita:2019:BiasedWordEmbeddings,Tan:2019:WordRepresentationBias,Zhang:2020:BiasClinicalWordEmbedding,Hutchingson:2020:NlpBiasDisabilities}.
These training data biases can translate into real-world harm, where, as an example, a recidivism model falsely flagged black defendants as high risk at twice the rate of white defendants~\citep{Angwin:2016:CompasPropublica}.

Understanding the data and its relationship to trained models is essential for building trustworthy ML systems.
However, it can be very difficult to answer even basic questions about the relationship between training data and model predictions; for example:
\begin{enumerate}
  \setlength{\itemsep}{0pt}
  \item Is a prediction well-supported by the training data, or was the prediction just random?
  \item Which portions of the training data improve a prediction? Which portions make it worse?
  \item Which instances in the training set caused the model to make a specific prediction?
\end{enumerate}

One strategy to address basic questions like those above is to render them moot by exclusively using simple, transparent model classes~\citep{Lipton:2019:MythosOfInterpretability}.
Evidence exists that this ``interpretable-only'' strategy may be appropriate in some settings~\citep{Knight:2017:DarkSecretAI}.
However, even interpretable model classes can be grossly affected by training data issues~\citep{Huber:1981:RobustStatistics,Cook:1982:NoteOutlierModel,Cook:1982:InfluenceRegression}.
Moreover, as the performance penalty of interpretable models grows, their continued use becomes harder to justify.
With the growing use of black-box models,
we need better methods to analyze and understand black-box model decisions.
Otherwise, society must carry the burden of black-box failures.

\subsection{Relating Models and Their Training Data}

All model decisions are rooted in the training data. %
Training data \textit{influence analysis} \revOne{(also known as \kw{data valuation}~\citep{Ghorbani:2019:DataShapley,Jia:2019:KnnShapley,Ki:2023:DataValuation} and \kw{data attribution}~\citep{Park:2023:Trak,Nguyen:2023:BayesianDataAttribution,Dai:2023:DataAttributionDiffsionModels})} partially demystifies the relationship between training data and model predictions by determining how to apportion credit (and blame) for specific model behavior to the training instances~\citep{Shapley:1988:ShapleyValue,Koh:2017:Understanding,Yeh:2018:Representer,Pruthi:2020:TracIn}.
Essentially, \revTwo{influence analysis's objective is to answer the question}: \textit{What is each training instance's effect on a model}?
An instance's ``effect'' is with respect to some specific perspective.
For example, an instance's effect may be quantified as the change in model performance when some instance is deleted from the training data.  %
The effect can also be relative, e.g., whether one training instance changes the model more than another.

Influence analysis emerged alongside the initial study of linear models and regression~\citep{Jaeckel:1972:InfinitesimalJackknife,Cook:1982:InfluenceRegression}.
This early analysis focused on quantifying how worst-case perturbations to the training data affected the final model parameters.
The insights gained from early influence analysis contributed to the development of numerous methods that improved model robustness and reduced model sensitivity to training outliers~\citep{Hogg:1979:StatisticalRobustness,Rousseeuw:1994:LeastMedianSquares}.

Since these early days, machine learning models have grown substantially in complexity and opacity~\citep{Devlin:2019:Bert,Krizhevsky:2012,Dosovitskiy:2021:VisionTransformer}.
Training datasets have also exploded in size~\citep{Black:2021:LeaveOneOut}.
These factors combine to make training data influence analysis significantly more challenging where, for multilayer parametric models (e.g., neural networks), determining a single training instance's exact effect can be NP\=/complete in the worst case~\citep{Blum:1992:TrainingNpComplete}.

In practice, influence may not need to be measured exactly.
\kw{Influence estimation} methods provide an approximation of training instances' true influence.
Influence estimation is generally much more computationally efficient and is now the approach of choice~\citep{Schioppa:2022:ScaledUpInfluenceFunctions}.
However, modern influence estimators achieve their efficiency via various assumptions about the model's architecture and learning environment~\citep{Koh:2017:Understanding,Yeh:2018:Representer,Ghorbani:2019:DataShapley}.
These varied assumptions result in influence estimators having different advantages and disadvantages as well as in some cases, even orthogonal perspectives on the definition of influence itself~\citep{Pruthi:2020:TracIn}.

\subsection{Our Contributions}

To the extent of our knowledge, there has not yet been a comprehensive review of these differing perspectives of training data influence, much less of the various methods themselves.
This paper fills in that gap by providing the first comprehensive survey of existing influence analysis techniques.
We describe how these various methods overlap and, more importantly, the consequences -- both positive and negative -- that arise out of their differences.
We provide this broad and nuanced understanding of influence analysis so that ML researchers and practitioners can better decide which influence analysis method best suits their specific application objectives~\citep{Schioppa:2022:ScaledUpInfluenceFunctions}.

\revOne{%
Although we aim to provide a comprehensive survey of influence analysis, we cannot cover every method in detail.
Instead, we focus on the most impactful methods so as not to distract from the key takeaways.
In particular, we concentrate on influence analysis methods that are general~\citep{Cook:1982:InfluenceRegression,Ghorbani:2019:DataShapley,Feldman:2020:InfluenceEstimation} or targeted towards parametric models \citep{Koh:2017:Understanding,Yeh:2018:Representer,Pruthi:2020:TracIn,Chen:2021:Hydra} with less emphasis on non-parametric methods~\citep{Sharchilev:2018:TreeInfluence,Jia:2019:KnnShapley,Brophy:2023:TreeInfluence}.
Multiple other research areas are based on ranking and subsampling training instances including data pruning~\citep{Yang:2023:DatasetPruning}, coreset selection~\citep{Bachem:2017:PracticalCoresetConstruction,Feldman:2020:CoresetsSurvey}, active learning~\citep{Ren:2021:ActiveLearning}, and submodular dataset selection~\citep{Wei:2015:SubmodularDatasetselection}, but these topics are beyond the scope of this work.%
}%
\footnote{%
  \revTwo{%
  Section~\ref{sec:EstimatorOverview:RelatedTopics} briefly contrasts how the objectives of these related areas align with influence analysis's objectives.%
  }%
}%

In the remainder of this paper, we first standardize the general notation used throughout this work (Sec.~\ref{sec:Preliminaries}).
Section~\ref{sec:EstimatorOverview} reviews the various general formulations through which training data influence is viewed.
We also categorize and summarize the properties of the seven most impactful influence analysis methods.
Sections~\ref{sec:Estimators:RetrainBased} and~\ref{sec:Estimators:GradientBased} describe these foundational influence methods in detail.
For each method, we (1)~formalize the associated definition of influence and how it is measured, (2)~detail the formulation's strengths and weaknesses, (3)~enumerate any related or derivative methods, and (4)~explain the method's time, space, and storage complexities.
Section~\ref{sec:Applications} reviews various learning tasks where influence analysis has been applied.
We provide our perspective on future directions for influence analysis research in Section~\ref{sec:FutureDirections}.
\section{General Notation}%
\label{sec:Preliminaries}

This section details our primary notation.
In cases where a single influence method requires custom nomenclature, we introduce the unique notation alongside discussion of that method.\footnote{Supplemental Table~\ref{tab:App:Nomenclature:MethodParams} provides a reference for all notation specific to a single influence analysis method.}
Supplemental Section~\ref{sec:App:Nomenclature} provides a full nomenclature reference.

Let $\setint{r}$ denote the set of integers $\myset{1, \ldots, r}$.
${A \simN{m} B}$ denotes that the cardinality of set~$A$ is~$m$ and that $A$ is drawn uniformly at random (u.a.r.)\ from set~$B$.
For \kw{singleton} set~$A$ (i.e., ${\abs{A} = 1}$), the sampling notation is simplified to ${A \simOne B}$.
Let $\powerSet{A}$ denote the \kw{power set} of any set~$A$.
Set subtraction is denoted ${A \setminus B}$.
For singleton~${B = \myset{b}}$, set subtraction is simplified to ${A \setminus b}$.

The \kw{zero vector} is denoted~$\zeroVec$ with the vector's dimension implicit from context.
$\ind{a}$ is the \kw{indicator function}, where ${\ind{a} = 1}$ if predicate~$a$ is true and~0 otherwise.

Let ${\X \in \domainX \subseteq \real^{\dimX}}$ denote an arbitrary \kw{feature vector}, and let ${\Y \in \domainY}$ be a \kw{dependent value} (e.g.,~label, target).
\kw{Training set}, ${\trainSet \defeq \myset{\zI}_{\trIdx = 1}^{\nTr}}$, consists of $\nTr$~training instances where each instance is a tuple, ${\zI \defeq (\xI, \yI) \in \domainZ}$ and ${\domainZ \defeq \domainX \times \domainY}$.
(Arbitrary) test instances are denoted ${\zTe \defeq (\xTe, \yTe) \in \domainZ}$.
Note that $\yTe$~need not be $\xTe$'s~true dependent value; $\yTe$~can be any value in~$\domainY$.
Throughout this work, subscripts~``$\trIdx$'' and~``$\teStr$'' entail that the corresponding symbol applies to an arbitrary training and test instance, respectively.

\kw{Model} $\func{\dec}{\domainX}{\domainY}$ is parameterized by ${\W \in \domainW}$, where ${\dimW \defeq \abs{\W}}$; $\dec$ is trained on (a subset of) dataset~$\trainSet$.
Most commonly, $\dec$~performs either classification or regression, although more advanced model classes (e.g.,~generative models) are also considered.
Model performance is evaluated using a \kw{loss function} $\func{\loss}{\domainY \times \domainY}{\real}$.
Let \eqsmall{${\risk{\zSym}{\W} \defeq \loss\big(\decFunc{\X}{\W}, \Y \big)}$} denote the \kw{empirical risk} of instance~${\zSym = (\X, \Y)}$ w.r.t.\ parameters~$\W$.
By convention, a smaller risk is better. %

This work primarily focuses on overparameterized models with ${\dimW \gg \dimX}$, \revOne{where $\dimX$~is the data dimension}.
Such models are almost exclusively trained using first-order optimization algorithms (e.g.,~gradient descent), which proceed iteratively over ${\nItr}$~iterations.
Starting from \kw{initial parameters}~${\wZero}$, the optimizer returns at the end of each iteration, ${\itr \in \setint{\nItr}}$, updated model parameters~$\wT$, where $\wT$ is generated from previous parameters~$\wTOne$, loss function~$\ell$, \kw{batch} ${\batchT \subseteq \trainSet}$, \kw{learning rate} ${\lrT > 0}$, and \kw{weight decay}~(\lTwo) strength ${\wdecay \geq 0}$.
Training gradients are denoted
${\gradW \risk{\zI}{\wT}}$.
The training set's \kw{empirical risk Hessian} for iteration $\itr$ is denoted \eqsmall{${\hessT \defeq \frac{1}{\nTr} \sum_{\zI \in \trainSet} \gradWSq \risk{\zI}{\wT}}$}, with the corresponding \kw{inverse risk Hessian} denoted~\eqsmall{$\invHessT$}.
Throughout this work, superscript~``${(\itr)}$'' entails that the corresponding symbol applies to training iteration~$\itr$.

Some models may be trained on data other than full training set~$\trainSet$, e.g., subset~${\trainSet \setminus \setZi}$.
Let ${\trainSetAlt \subset \trainSet}$ denote an alternate training set, and denote model parameters trained on~$\trainSetAlt$ as~$\wTSubBase{\trainSetAlt}$.
For example, $\wFinSub$ are the final parameters for a model trained on all of~$\trainSet$ except training instance~${\zI \in \trainSet}$.
When training on all of the training data, subscript~$\trainSet$ is dropped, i.e., ${\wT \equiv \wTSubBase{\trainSet}}$.
\section{Overview of Influence and Influence Estimation}%
\label{sec:EstimatorOverview}

\revTwo{As Section~\ref{sec:Intro} explains, training data influence's objective is to quantify the ``effect'' of one or more training instances on a model.}
This effect's scope can be as localized as an individual model prediction, e.g., $\decFunc{\xTe}{\wFin}$;
the effect's scope can also be so broad as to encompass the entire test data distribution.

Positive influence entails that the training instance(s) improve some quality measure, e.g., risk $\risk{\zTe}{\wFin}$.
Negative influence means that the training instance(s) make the quality measure worse.
Training instances with positive influence are referred to as \kw{proponents} or \kw{excitatory examples}.
Training instances with negative influence are called \kw{opponents} or \kw{inhibitory examples}~\citep{Koh:2017:Understanding,Yeh:2018:Representer}.

Highly expressive, overparameterized models remain functionally black boxes~\citep{Koh:2017:Understanding}.
Understanding why a model behaves in a specific way remains a significant challenge~\citep{Belle:2021:PrinciplesXAI},
and
the inclusion or removal of even a single training instance can drastically change a trained model's behavior~\citep{Rousseeuw:1994:LeastMedianSquares,Black:2021:LeaveOneOut}.
In the worst case, quantifying one training instance's influence may require repeating all of training.

Since measuring influence \textit{exactly} may be intractable or unnecessary, \textit{influence estimators} -- which only approximate the true influence -- are commonly used in practice.
As with any approximation, influence estimation requires making trade-offs, and the various influence estimators balance these design choices differently.
This in turn leads influence estimators to make different assumptions and rely on different mathematical formulations.

When determining which influence analysis methods to highlight in this work, we relied on two primary criteria: (1)~a method's overall impact and (2)~the method's degree of novelty in relation to other approaches.
\revOne{%
In particular, we concentrate on influence analysis methods that are either model architecture agnostic or that are targeted towards parametric models (e.g., neural networks).
Nonetheless, we briefly discuss non-parametric methods as well.
}

The remainder of this section considers progressively more general definitions of influence.

\subsection{Pointwise Training Data Influence}%
\label{sec:EstimatorOverview:PointwiseInfluence}

\kw{Pointwise influence} is the simplest and most commonly studied definition of influence.
It quantifies how a single training instance affects a model's prediction on a single test instance according to some quality measure (e.g.,~test loss).
Formally, a pointwise influence analysis method is a function $\func{\influence}{\domainZ \times \domainZ}{\real}$ with the pointwise influence of training instance~$\zI$ on test instance~$\zTe$ denoted $\infFunc{\zI}{\zTe}$.
Pointwise influence estimates are denoted ${\infEstFunc{\zI}{\zTe} \in \real}$.
\revOne{%
  Note that the model architecture, training algorithm, full training set~($\trainSet$), and even the random seed can (significantly) affect an instance's influence.
  To improve clarity and readability, we treat these parameters as fixed and implicit in our nomenclature for $\influence$ and~$\infEstSym$.
}%

Below, we briefly review early pointwise influence analysis contributions and then transition to a discussion of more recent pointwise methods.

\subsubsection{Early Pointwise Influence Analysis}%
\label{sec:EstimatorOverview:PointwiseInfluence:EarlyMethods}

\begin{figure}[t]
  \centering
\begin{tikzpicture}
    \begin{axis}
      [
        hide axis,
        xmin=0,
        xmax=1,
        ymin=0,
        ymax=1,
        scale only axis,width=1mm, %
        legend cell align={left},              %
        legend style={font=\footnotesize},
        legend columns=5,
        legend image post style={scale=0.65}, %
        legend style={/tikz/every even column/.append style={column sep=0.5cm}}
      ]
      \addlegendimage{only marks,mark=square*,blue!30,draw=blue,mark size=3.5pt}
      \addlegendentry{Inlier}

      \addlegendimage{only marks,mark=*,red!30,draw=red,mark size=3.5pt,line width=0.6pt}
      \addlegendentry{Outlier}

      \addlegendimage{dashed, domain=0:5, smooth, color=\linRegCleanLineColor, line width=1.2pt, opacity=1}
      \addlegendentry{Least Squares Inliers Only}

      \addlegendimage{domain=0:5, smooth, color=\linRegAllLineColor, line width=1.2pt, opacity=1}
      \addlegendentry{Least Squares All}
    \end{axis}%
\end{tikzpicture}%
 
  \vspace{8pt}
\newcommand{\plotWidth}{2in}
\newcommand{\plotHeight}{\plotWidth}
\newcommand{\plotFontSize}{\footnotesize}

\newcommand{\linRegMinX}{0.5}
\newcommand{\linRegMaxX}{5.5}
\newcommand{\linRegLineWidth}{0.7pt}
\begin{tikzpicture}
  \pgfplotstableread[col sep=comma] {plots/src/data/linear_reg.csv}\thedata
  \begin{axis}
     [
       scale only axis,%
       width={\plotWidth},
       height={\plotHeight},
       xmin={\linRegMinX},
       xmax={\linRegMaxX},
       xtick={1,2,3,4,5},
       x tick label style={font=\plotFontSize,align=center},
       xlabel={\plotFontSize$\X$},
       xmajorgrids,
       ymin=0.5,
       ymax=4.5,
       ylabel={\plotFontSize$\Y$},
       ytick distance={1},
       ymajorgrids,
       y tick label style={font=\plotFontSize,align=center},
       point meta=explicit,
       clip mode=individual,                %
       scatter/classes={
         0={mark=square*,blue!30,draw=blue,mark size=2.0pt},
         1={mark=*,red!30,draw=red,mark size=2.0pt,line width=0.6pt}
       },
     ]

     \addplot[dashed, domain=\linRegMinX:\linRegMaxX, smooth, color=\linRegCleanLineColor, line width=\linRegLineWidth, opacity=1] { 1.997 * x + 0.007 } {};    %
     \label{leg:EstimatorOverview:PointwiseInfluence:EarlyMethods:LSq:InDist}

     \addplot[domain=\linRegMinX:\linRegMaxX, smooth, color=\linRegAllLineColor, line width=\linRegLineWidth, opacity=1] { -0.3797 * x + 3.477 } {};    %
     \label{leg:EstimatorOverview:PointwiseInfluence:EarlyMethods:LSq:All}

     \addplot[
       scatter,
       only marks,
       line width=0.2,
       mark size=0.7pt,
     ]
     table[x index=0,y index=1, meta index=2] {\thedata};
     \label[0]{leg:EstimatorOverview:PointwiseInfluence:EarlyMethods:Clean}
     \label[1]{leg:EstimatorOverview:PointwiseInfluence:EarlyMethods:Outlier}
  \end{axis}
\end{tikzpicture}
   \caption{%
    \textbf{Outlier Pointwise Influence on Least-Squares Regression}:
    Influence of a single outlier
    (\ref{leg:EstimatorOverview:PointwiseInfluence:EarlyMethods:Outlier})
    on a least-squares model where in-distribution data
    (\ref{leg:EstimatorOverview:PointwiseInfluence:EarlyMethods:Clean})
    are generated from
    linear distribution ${\Y = 2 \X}$.
    The single outlier sample
    (${\X = 5}$ \& ${\Y = 1.2}$)
    influences the inlier-only least-squares linear model
    (\ref{leg:EstimatorOverview:PointwiseInfluence:EarlyMethods:LSq:InDist})
    substantially such that a least-squares model trained on all instances
    (\ref{leg:EstimatorOverview:PointwiseInfluence:EarlyMethods:LSq:All})
    predicts all training $\Y$~values poorly.
    Adapted from \citet[Fig.~2(b)]{Rousseeuw:1997:RobustRegressionOutlierDetection}.
  }
  \label{fig:EstimatorOverview:PointwiseInfluence:EarlyMethods:LeastSquares}
\end{figure}

The earliest notions of pointwise influence emerged out of robust statistics -- specifically the analysis of training outliers' effects on linear regression models~\citep{Cook:1977:DetectionInfluence}.
Given training set~$\trainSet$, the least-mean squares linear model parameters are%
\footnote{%
  For simplicity, the bias term is considered part of $\W$.%
}%
\begin{equation}%
  \label{eq:EstimatorOverview:LeastSquares}
  \wOpt
    \defeq
      \argmin_{\W}
        \frac{1}%
             {\abs{\trainSet}}
        \sum_{(\xI,\yI) \in \trainSet}
          \left( \yI - \W\transpose\xI \right)^2
  \text{.}
\end{equation}
\noindent
Observe that this least-squares estimator has a \kw{breakdown point} of~0~\citep{Rousseeuw:1994:LeastMedianSquares}.
This means that least-squares regression is completely non-robust where a single training data outlier can shift model parameters~$\wOpt$ arbitrarily.
For example, Figure~\ref{fig:EstimatorOverview:PointwiseInfluence:EarlyMethods:LeastSquares} visualizes how a single training data outlier~(\ref{leg:EstimatorOverview:PointwiseInfluence:EarlyMethods:Outlier}) can induce a nearly orthogonal least-squares model.
Put simply, an outlier training instance's potential pointwise influence on a least-squares model is unbounded.
Unbounded influence on the model parameters equates to unbounded influence on model predictions.

Early influence analysis methods sought to identify the training instance that was most likely to be an outlier~\citep{Srikantan:1961:SingleOutlier,Tietjen:1973:TestingSingleOutlier}.
A training outlier can be defined as the training instance with the largest negative influence on prediction $\decFunc{\xTe}{\wOpt}$.
Intuitively, each training instance's pointwise influence can be measured by training ${\nTr \defeq \abs{\trainSet}}$ models, where each model's training set leaves out a different training instance.
These $\nTr$~models would then be compared to identify the outlier.%
\footnote{%
  Section~\ref{sec:Estimators:RetrainBased:LeaveOneOut} formalizes how to measure pointwise influence by repeatedly retraining with a different training instance left out of the training set each time.%
}
However, such repeated retraining is expensive
and so more efficient pointwise influence analysis methods were studied.

Different assumptions about the training data distribution lead to different definitions of the most likely outlier.
For example, \citet{Srikantan:1961:SingleOutlier}, \citet{Snedecor:1968:StatisticalMethods}, and \citet{Ellenberg:1976:SingleOutlierRegression} all assume that training data outliers arise from mean shifts in normally distributed training data.
Under this constraint, their methods all identify the maximum likelihood outlier as the training instance with the largest absolute \kw{residual},~$\abs{\yI - {\wOpt}\transpose \xI}$.
However, \citet{Cook:1982:InfluenceRegression} prove that under different distributional assumptions (e.g.,~a variance shift instead of a mean shift), the maximum likelihood outlier may not have the largest residual.

These early influence analysis results demonstrating least-squares fragility spurred development of more robust regressors.
For instance, \citet{Rousseeuw:1994:LeastMedianSquares} replaces Eq.~\eqref{eq:EstimatorOverview:LeastSquares}'s mean operation with median;
this simple change increases the breakdown point of model parameters~$\wOpt$ to the maximum value,~50\%.
In addition, multiple robust loss functions have been proposed that constrain or cap outliers' pointwise influence~\citep{Huber:1964:RobustEstimation,Beaton:1974:TukeyBiweight,Dennis:1978:WelschLoss,Leclerc:1989}.

As more complex models grew in prevalence, influence analysis methods similarly grew in complexity.
In recent years, numerous influence analysis methods targeting deep models have been proposed.
We briefly review the most impactful, modern pointwise influence analysis methods next.

\subsubsection{Modern Pointwise Influence Analysis}%
\label{sec:EstimatorOverview:PointwiseInfluence:ModernMethods}

\begin{figure}[t]
  \centering
  \makebox[\textwidth][c]{%
  \newcommand{\citeFig}[1]{}
  {
    \tikzset{
      my rounded corners/.append style={rounded corners=2pt},
    }
    \scriptsize
    \begin{forest}
      /tikz/every node/.append style={font=\scriptsize},
      for tree={
        line width=1pt,
        if={level()<4}{
          my rounded corners,
          draw=linecol,
        }{},
        if level=0{%
          forked edges,
          l sep+=0.35cm,  %
          s sep+=0.35cm,  %
          align=center,
        }{%
          if level=1{%
            l sep+=0.35cm,
            s sep+=8pt,   %
            parent anchor=south,
            child anchor=north,
            align=center,
            for descendants={
              align=center,
            },
          }{
            if level=2{
              l sep+=0.35cm,
              s sep+=8pt,   %
            }{
              if level=3{
                tier=parting ways,  %
                parent anchor=south west,
                 for descendants={
                   child anchor=west,
                   parent anchor=west,
                   anchor=west,
                   align=left,
                 },
              }{
                if level=4{
                  shape=coordinate,
                  no edge,
                  grow'=0,
                  calign with current edge,
                  xshift=15pt,
                  for descendants={
                    parent anchor=south west,
                  },
                  for children={
                    s sep-=10pt,
                    edge path={
                      \noexpand\path[\forestoption{edge}] (!to tier=parting ways.parent anchor) |- (.child anchor)\forestoption{edge label};
                    },
                    for descendants={
                      no edge,
                    },
                  },
                }{
                },
              },
            },
          },
        },%
      },
      [Influence Analysis Methods
        [Retraining-Based %
          [{},nice empty nodes,calign=child, calign child=2
             [ Leave-One-Out\citeFig{Cook:1982:InfluenceRegression}
                [
                  [ \knnLOO{}\citeFig{Jia:2021:ScalabilityVsUtility} ]
                  [ \leafRefit{}\citeFig{Sharchilev:2018:TreeInfluence} ]
                  [ Inf.\ Sketching\citeFig{Wojnowicz:2016:InfluenceSketching} ]
                ]
             ]
             [ \feldman{}\citeFig{Feldman:2020:InfluenceEstimation}
              [
                [ C\=/Score\citeFig{Jiang:2021:CScore} ]
                [ Generative\citeFig{Van:2021:Memorization} ]
              ]
             ]
             [ Shapley Value\citeFig{Shapley:1953}
               [
                 [ Interaction Index\citeFig{Grabisch:1999:InteractionAmongPlayers} ]
                 [ Shapley-Taylor\citeFig{Sundararajan:2020:ShapleyTaylorInteraction} ]
                 [ \mcShap{}\citeFig{Ghorbani:2019:DataShapley} ]
                 [ \gShap{}\citeFig{Ghorbani:2019:DataShapley} ]
                 [ \knnShap{}\citeFig{Jia:2019:KnnShapley} ]
                 [ \betaShap{}\citeFig{Kwon:2022:BetaShapley} ]
                 [ Banzhaf Value\citeFig{Wang:2023:BanzhafValue} ]
                 [ \ame{}\citeFig{Lin:2022:InfluenceRandomizedExperiments} ]
                 [ \SHAP{}\citeFig{Lundberg:2017:SHAP} ]
                 [ Neuron Shapley\citeFig{Ghorbani:2020:NeuronShapley} ]
               ]
             ]
          ]
        ]
        [Gradient-Based
           [Static
              [ Influence Func.\citeFig{Koh:2017:Understanding}
                [
                  [ \fastif{}\citeFig{Guo:2021:FastIF} ]
                  [ Arnoldi IF\citeFig{Schioppa:2022:ScaledUpInfluenceFunctions} ]
                  [ \leafInfluence{}\citeFig{Sharchilev:2018:TreeInfluence} ]
                  [ Group IF\citeFig{Koh:2019:GroupEffects} ]
                  [ Second-Order\citeFig{Basu:2020:OnSecond} ]
                  [ RelatIF\citeFig{Barshan:2020:RelatIF} ]
                  [ Renorm.\ IF\citeFig{Hammoudeh:2022:GAS} ]
                ]
              ]
              [ Representer Pt.\citeFig{Yeh:2018:Representer}
                [
                  [ High Dim.\ Rep.\citeFig{Tsai:2023:HighDimRepPt} ]
                  [ \rpLocalJacobian\citeFig{Sui:2021:LocalJacobianRepPt} ]
                  [ \trex{}\citeFig{Brophy:2023:TreeInfluence} ]
                ]
              ]
           ]
           [Dynamic
              [\tracin{}\citeFig{Pruthi:2020:TracIn}
                [
                  [ \tracinCP{}\citeFig{Pruthi:2020:TracIn} ]
                  [ \tracinRP{}\citeFig{Pruthi:2020:TracIn} ]
                  [ \tracinLast{}\citeFig{Pruthi:2020:TracIn} ]
                  [ \vaeTracIn{}\citeFig{Kong:2021:VaeTracIn} ]
                  [ \tracinAD{}\citeFig{Thimonier:2022:TracInAD} ]
                  [ \tracinWE{}\citeFig{Yeh:2022:FirstBetterLast} ]
                  [ \boostin{}\citeFig{Brophy:2023:TreeInfluence} ]
                  [ \gas{}\citeFig{Hammoudeh:2022:GAS} ]
                ]
              ]
              [\hydra{}\citeFig{Chen:2021:Hydra}
                [
                  [ SGD\=/Influence\citeFig{Hara:2019:SgdInfluence} ]
                ]
              ]
           ]
        ]
     ]
    \end{forest}
  }
}

  \caption{%
    \textbf{Influence Analysis Taxonomy}:
    Categorization of the seven primary pointwise influence analysis methods.
    Section~\ref{sec:Estimators:RetrainBased} details the three primary retraining-based influence methods,
    leave-one-out (Sec.~\ref{sec:Estimators:RetrainBased:LeaveOneOut}),
    \feldman{} (Sec.~\ref{sec:Estimators:RetrainBased:Feldman}),
    and
    Shapley value (Sec.~\ref{sec:Estimators:RetrainBased:Shapley}).
    Section~\ref{sec:Estimators:GradientBased} details gradient-based static estimators
    influence functions (Sec.~\ref{sec:Estimators:GradientBased:Static:IF})
    and
    representer point (Sec.~\ref{sec:Estimators:GradientBased:Static:RepresenterPoint})
    as well as dynamic estimators
    \tracin{} (Sec.~\ref{sec:Estimators:GradientBased:Dynamic:TracIn}) and
    \hydra{} (Sec.~\ref{sec:Estimators:GradientBased:Dynamic:HyDRA}).
    Closely-related and derivative estimators are shown as a list below their parent method.
    See supplemental Table~\ref{tab:App:Nomenclature:InfDef} for the formal mathematical definition of all influence methods and estimators.
    Due to space, each method's citation is in supplemental Table~\ref{tab:App:Nomenclature:Methods}.
  }%
  \label{fig:Estimator:Taxonomy}
\end{figure}

Figure~\ref{fig:Estimator:Taxonomy} provides a taxonomy of the seven most impactful modern pointwise influence analysis methods.
Below each method appears a list of closely related and derivative approaches.
Modern influence analysis methods broadly categorize into two primary classes, namely:
\begin{itemize}
  \item \kw{Retraining-Based Methods}: Measure the training data's influence by repeatedly retraining model~$\dec$ using different subsets of training set~$\trainSet$.

  \item \kw{Gradient-Based Influence Estimators}: Estimate influence via the alignment of training and test instance gradients either throughout or at the end of training.
\end{itemize}
\noindent
An in-depth comparison of these influence analysis methods requires detailed analysis so we defer the extensive discussion of these two categories to Sections~\ref{sec:Estimators:RetrainBased} and~\ref{sec:Estimators:GradientBased}, respectively. %
Table~\ref{tab:InfluenceEstimators:Comparison} summarizes the key properties of Figure~\ref{fig:Estimator:Taxonomy}'s seven methods -- including comparing each method's assumptions (if any), strengths/weaknesses, and asymptotic complexities.
These three criteria are also discussed when detailing each of these methods in the later sections.
\afterpage{%
  \clearpage
  \newpage
  \begin{table}[t!]
    \centering
    \caption{%
      \textbf{Influence Analysis Method Comparison}:
      Comparison of the complexity, assumptions, strengths, and limitations of the seven primary influence estimators detailed in Sections~\ref{sec:Estimators:RetrainBased} and~\ref{sec:Estimators:GradientBased}.
      Recall that $\nTr$, $\dimW$, and $\nItr$ are the training-set size, model parameter count, and training iteration count, respectively.
      \loo{} is the leave\=/one\=/out influence.
      ``Full time complexity" denotes the time required to calculate influence for the first test instance; ``incremental complexity" is the added time required for each subsequent test instance.
      ``Storage complexity" represents the amount of memory required to persistently save any additional model parameters needed by the influence method.
      For \hydra{} and the three retraining-based methods, the storage complexity is implementation dependent and is marked with an asterisk.
      We report each method's worst-case storage complexity;
      for the retraining-based methods, this worst-case storage complexity yields the best-case incremental complexity.
      Static, gradient-based estimators -- influence functions and representer point -- require no additional storage.
      Differentiability encompasses both model~$\dec$ and loss function~$\loss$ except in the case of representer point where only the loss function must be differentiable.
      All criteria below are discussed in detail alongside the description of each method.
    }
    \label{tab:InfluenceEstimators:Comparison}
    {
      \footnotesize%
\renewcommand{\arraystretch}{1.2}
\setlength{\dashlinedash}{0.4pt}
\setlength{\dashlinegap}{1.5pt}
\setlength{\arrayrulewidth}{0.3pt}
\newcommand{\myrule}{\cdashline{1-8}}

\newcommand{\bigOexpT}{$\bigO{2^{\nTr}\,\nItr}$}
\newcommand{\bigOexp}{$\bigO{2^{\nTr}}$}
\newcommand{\bigOexpP}{$\bigO{2^{\nTr}\,\dimW}$}
\newcommand{\bigOkp}{$\bigO{\nFeldModel\dimW}$}

\newcommand{\zeroStore}{0}

\newcommand{\SectionRow}{
  Section reference
  & \ref{sec:Estimators:RetrainBased:LeaveOneOut}
  & \ref{sec:Estimators:RetrainBased:Feldman}
  & \ref{sec:Estimators:RetrainBased:Shapley}
  & \ref{sec:Estimators:GradientBased:Static:IF}
  & \ref{sec:Estimators:GradientBased:Static:RepresenterPoint}
  & \ref{sec:Estimators:GradientBased:Dynamic:TracIn}
  & \ref{sec:Estimators:GradientBased:Dynamic:HyDRA}
}

\newcommand{\ck}{\checkmark}
\begin{tabular}{@{}lccccccc@{}}
  \toprule
                            & \multicolumn{3}{c}{\multirow{2}{*}{Retraining-Based}} &  \multicolumn{4}{c}{Gradient-Based}  \\\cmidrule(lr){5-8}
                            &            &            &              & \multicolumn{2}{c}{Static} & \multicolumn{2}{c}{Dynamic} \\\cmidrule(lr){2-4}\cmidrule(lr){5-6}\cmidrule(lr){7-8}
                            & \loo{}     & \feldmanShort{} & Shapley      & Inf.\ Func.  & Rep.\ Pt.   & \tracin{} & \hydra{}      \\
  \midrule
  \SectionRow{} \\\myrule
  Full time complexity      & \bigOnT{}  & \bigOkT    & \bigOexpT{}  & \bigOnP      & \bigOn      & \bigOnpT  & \bigOnpT{}     \\\myrule
  Incremental complexity    & \bigOn{}   & \bigOk{}   & \bigOexp{}   & \bigOnP      & \bigOn      & \bigOnpT  & \bigOnp{}      \\\myrule
  Space complexity          & \bigOpPln  & \bigOpPln  & \bigOpPln    & \bigOpPln    & \bigOpPln   & \bigOpPln & \bigOnp{}      \\\myrule
  Storage complexity        & \bigOnP{}* & \bigOkp{}* & \bigOexpP{}* & \zeroStore   & \zeroStore  & \bigOpT   & \bigOpTpn{}*   \\\myrule
  Assumes differentiable?   &            &            &              & \ck{}        & \ck{}*      & \ck{}     & \ck{}          \\\myrule
  Assumes convexity?        &            &            &              & \ck{}        & \ck{}       &           &                \\\myrule
  Assumes stationarity?     &            &            &              & \ck{}        & \ck{}       &           &                \\\myrule
  Assumes linear model?     &            &            &              &              & \ck{}       &           &                \\\myrule
  Optimizer specific?       &            &            &              &              &             & \ck{}     & \ck{}          \\\myrule
  Uses Hessian?             &            &            &              & \ck{}        &             &           & \ck{}          \\\myrule
  Any model class?          & \ck{}      & \ck{}      & \ck{}        &              &             &           &                \\\myrule
  Estimates \loo{}?         & \ck{}      & \ck{}      & \ck{}        & \ck{}        &             &           & \ck{}          \\\myrule
  Group influence?          &            & \ck{}      & \ck{}        & \ck{}        &             &           &                \\\myrule
  Hyperparam.\ sensitive?   &            &            &              & \ck{}        &             &           & \ck{}          \\\myrule
  Generative models?        &            & \ck{}      & \ck{}        &              &             & \ck{}     &                \\\myrule
  High upfront cost?        & \ck{}      & \ck{}      & \ck{}        &              &             &           & \ck{}          \\\myrule
  High instance cost?       &            &            & \ck{}        & \ck{}        &             & \ck{}     &                \\\myrule
  Amortizable?              & \ck{}      & \ck{}      & \ck{}        &              &             & \ck{}     & \ck{}          \\
  \bottomrule
\end{tabular}
    }
  \end{table}
  \clearpage
  \newpage
}

\subsection{Alternative Perspectives on Influence}%
\label{sec:EstimatorOverview:AlternativeDefinitions}

Note that pointwise effects are only one perspective on how to analyze the training data's influence.
Below we briefly summarize six alternate, albeit less common, perspectives of training data influence.
While pointwise influence is this work's primary focus, later sections also contextualize existing influence methods w.r.t.\ these alternate perspectives where applicable.

\noindent%
(1)~%
Recall that pointwise influence quantifies the effect of a single training instance on a single test prediction.
In reality, multiple related training instances generally influence a prediction as a group~\citep{Feldman:2020:InfluenceEstimation}, where group members
have a total effect much larger than the sum of their individual effects~\citep{Basu:2020:OnSecond,Das:2021:DataSubsetInfluence,Hammoudeh:2022:GAS}.
\kw{Group influence} quantifies a set of training instances' total, combined influence on a specific test prediction.

We use very similar notation to denote group and pointwise influence.
The only difference is that for group influence, the first parameter of function~$\influence$ is a training (sub)set instead of an individual training instance;
the same applies to group influence estimates~$\infEstSym$.
For example, given some test instance~$\zTe$, the entire training set's group influence and group influence estimate are denoted $\infFunc{\trainSet}{\zTe}$ and $\infEstFunc{\trainSet}{\zTe}$, respectively.

In terms of magnitude,
previous work has shown that the group influence of a related set of training instances is generally lowered bounded by the sum of the set's pointwise influences~\citep{Koh:2019:GroupEffects}.
Put simply, the true influence of a coherent group is \textit{more than the sum of its parts}, or formally, for coherent ${\trainSetAlt \subseteq \trainSet}$, it often holds that
\begin{equation}%
  \label{eq:EstimatorOverview:GroupInfluence:Bound}%
  \abs{%
    \infFunc{\trainSetAlt}{\zTe}%
  }%
  >
    \sum_{\zI \in \trainSetAlt}
      \abs{%
        \infFunc{\zI}%
                {\zTe}%
      }%
  \text{.}
\end{equation}
\noindent
Existing work studying group influence is limited.
Later sections note examples where any of the seven primary pointwise influence methods have been extended to consider group effects.

\noindent%
(2)~%
\kw{Joint influence} extends influence to consider \textit{multiple test instances} collectively~\citep{Jia:2022:CertifiedKNN,Chen:2022:BaggingTrainingSetAttacks}.
These test instances may be a specific subpopulation within the test distribution -- for example in \kw{targeted data poisoning attacks}~\citep{Jagielski:2021:Subpopulation,Wallace:2021}.
The test instances could also be a representative subset of the entire test data distribution -- for example in \kw{coreset selection}~\citep{Borsos:2020:BilevelCoresetSelection} or \kw{indiscriminate poisoning attacks}~\citep{Biggio:2012:Poisoning,Fowl:2021:Adversarial}.

Most (pointwise) influence analysis methods are \kw{additive} meaning for target set~${\testSet \subseteq \domainZ}$, the joint (pointwise) influence simplifies to
\begin{equation}\label{eq:Estimators:MultitargetInfluence}
  \infFunc{\zI}{\testSet}
    =
      \sum_{\zTe \in \testSet}
        \infFunc{\zI}%
                {\zTe}
  \text{.}
\end{equation}
\noindent
Additivity is not a requirement of influence analysis, and there are provably non-additive influence estimators~\citep{Yan:2021:ShapleyCore}.

\noindent%
(3)~%
Overparameterized models like deep networks are capable of achieving near-zero training loss in most settings~\citep{Bartlett:2020:BenignOverfitting,Feldman:2020:LearningRequiresMemorization,DAmour:2020:Underspecification}.
This holds even if the training set is large and randomly labeled~\citep{Zhang:2017:RethinkingGeneralization,Arpit:2017:DeepMemorization}.
Near-zero training loss occurs because deep models often memorize some training instances.

Both \citet{Pruthi:2020:TracIn} and \citet{Feldman:2020:InfluenceEstimation} separately define a model's \kw{memorization}\footnote{\citet{Pruthi:2020:TracIn} term ``memorization'' as \kw{self-influence}. We use \citepos{Feldman:2020:InfluenceEstimation} terminology here since it is more consistent with other work~\citep{Van:2021:Memorization,Kandpal:2022:DeduplicatingPrivacyRisks}.} of training instance~$\zI$ as the pointwise influence of $\zI$ \textit{on itself}.
Formally
\begin{equation}
  \labelAndRemember{eq:EstimatorOverview:Memorization}{%
    \mem{\zI}
    \defeq
      \infFunc{\zI}{\zI}
  }
  \approx
    \infEstFunc{\zI}{\zI}
  \text{.}
\end{equation}

\noindent
(4)~%
\kw{Cook's distance} measures the effect of training instances on the model parameters themselves~\citep[Eq.~(5)]{Cook:1977:DetectionInfluence}.
Formally, the pointwise Cook's distance of ${\zI \in \trainSet}$ is
\begin{equation}
  \labelAndRemember{eq:EstimatorOverview:CooksDistance:Pointwise}{%
    \infCookFunc{\zI}
      \defeq
        \wFin
        -
        \wFinSub
  }%
  \text{.}
\end{equation}
\noindent%
Eq.~\eqref{eq:EstimatorOverview:CooksDistance:Pointwise} trivially extends to groups of training instances where for any ${\trainSetAlt \subseteq \trainSet}$
\begin{equation}
  \infCookFunc{\trainSetAlt}
    \defeq
      \wFin
      -
      \wFinSubBase{\trainSet \setminus \trainSetAlt}
  \text{.}
\end{equation}
Cook's distance is particularly relevant for interpretable model classes where feature weights are most transparent.
This includes linear regression~\citep{Rousseeuw:1997:RobustRegressionOutlierDetection,Wojnowicz:2016:InfluenceSketching} and decision trees~\citep{Brophy:2023:TreeInfluence}.

\noindent%
(5)~%
All definitions of influence above consider training instances' effects w.r.t.\ a single instantiation of a model.
Across repeated stochastic retrainings, a training instance's influence may vary -- potentially substantially~\citep{Basu:2021:InfluenceFunctionsFragile,Summers:2021:NondeterminismNeuralNetwork,Raste:2022:QuantifyingRandomness}.
\kw{Expected influence} is the average influence across all possible instantiations within a given model class~\citep{K:2021:RevisitingInfluentialExamples,Wang:2023:DataBanzhaf}.
Expected influence is particularly useful in domains where the random component of training is unknowable a priori.
For example, with poisoning and backdoor attacks, an adversary crafts malicious training instances to be highly influential in expectation across all random parameter initializations and batch orderings~\citep{Chen:2017:Targeted,Shafahi:2018:PoisonFrogs,Fowl:2021:Adversarial}.

Expected influence generalizes to consider group effects.
\revOne{%
Existing related work focuses on counterfactuals such as, ``what is the expected prediction for~$\xTe$ if a model is trained on some arbitrary subset of $\trainSet$~\citep{Ilyas:2022:Datamodels,Ki:2023:DataValuation}?''
Other work seeks to predict model parameters~$\wFin$ given an arbitrary training subset~\citep{Zeng:2023:ModelPred}.%
}

\noindent%
(6)~%
Observe that all preceding definitions view influence as a specific numerical value to measure/estimate.
Influence analysis often simplifies to a relative question of whether one training instance is more influential than another.
An \kw{influence ranking} orders (groups of) training instances from most positively influential to most negatively influential.
These rankings are useful in a wide range of applications~\citep{Kwon:2022:BetaShapley,Wang:2023:DataBanzhaf}, including data cleaning and poisoning attack defenses as discussed in Section~\ref{sec:Applications}.

\newcommand{\relatedParagraph}[1]{%
  \relatedHeading{#1}%
}

\subsection{\revTwo{Topics Related to Influence Analysis}}%
\label{sec:EstimatorOverview:RelatedTopics}

\revTwo{%
  All influence analysis methods we highlight in Sections~\ref{sec:Estimators:RetrainBased} and~\ref{sec:Estimators:GradientBased} estimate a function that quantifies the impact of specific training examples on a given model or prediction.
  For the most part, these methods take an ablation perspective, i.e., measuring how much a model changes when removing specific training examples.
  However, there are several other research areas related to analyzing the impact of different subsets of the training data, albeit with somewhat different methods or objectives.
  We briefly describe some of these topics below.%
  \footnote{%
    Section~\ref{sec:Applications} (``\nameref{sec:Applications}'') discusses additional tasks that have leveraged influence analysis to achieve their own meta-objectives (e.g., adversarial robustness, model explainability, etc.).%
  }

\newcommand{\coresetCost}[1]{\text{cost}(#1, Q)}

  \emph{Data pruning methods} such as coresets~\citep{Bachem:2017:PracticalCoresetConstruction} also consider the impact of removing examples, but they typically consider removing many examples (rather than one or a few) with the primary goal of increasing computational efficiency.
  A \kw{coreset} ${\trainSetCore \subset \trainSet}$ is a (weighted) set of points that can stand in for the overall training data when measuring a \kw{cost function}, $\coresetCost{\trainSetCore}$ where ${Q \in \mathcal{Q}}$ is a solution in solution space~$\mathcal{Q}$.
  $\trainSetCore$ is an \kw{$\varepsilon$\=/coreset} if it approximates the cost function within a factor of ${\epsilon > 0}$:
  \begin{equation}
    \abs{%
      \coresetCost{\trainSetCore}
      -
      \coresetCost{\trainSet}
    }
    \leq
    \varepsilon \,\coresetCost{\trainSet}
    \text{.}
  \end{equation}
  Several techniques exist to efficiently find~$\trainSetCore$ \citep{Mirzasoleiman:2020:Coresets,Feldman:2020:CoresetsSurvey,Tukan:2023:ProvableDataSelection}.
  These include methods loosely based on weighted \kw{importance sampling}, where each training instance's sampling probability is proportional to the instance's influence~\citep{Bachem:2017:PracticalCoresetConstruction}.

Coresets let us learn a model over a much smaller set of points, while still yielding a model that is within a factor of ${1 + \varepsilon}$ of the optimal loss on the original training set.
Therefore, a coreset represents a \emph{sufficient} set of points for a given task, while most influence estimation methods identify the most \emph{necessary} points --- the points without which performance would decline, even given many other points from the original dataset.
Another difference is that coresets are often motivated by efficiency concerns, whereas influence estimation is more motivated by the need to understand the data and its impact on a model --- specifically, a model trained on the entire training data.

Coreset construction often involves \emph{submodular optimization}~\citep{Bilmes:2022:SubmodularityInML}, so that an efficient, greedy approach finds a nearly-optimal set of points.
However, this also means that if there are multiple, equally-important points, submodular optimization will select one and skip the others as redundant.
This ``winner-take-all" approach is in stark contrast to most influence estimation methods, which tend to assign similar importance to similar points.

\emph{Active learning} seeks to \textit{maximize} a model's performance while annotating as little training data as possible~\citep{Ren:2021:ActiveLearning}.
Like influence estimation, active learning estimates the relative value different data points would have in fitting a model.
However, unlike influence estimation, this is done without knowledge of the labels of these points.
Furthermore, the goal is maximizing performance more than understanding the data, increasing efficiency, or precisely matching the loss on the full training data.

Often, the data points to label are chosen greedily by identifying the training instance whose labeling would most positively influence the model (in expectation).
Quantifying each unlabeled instance's true influence at each active learning iteration may be prohibitive, so influence \kw{estimation} techniques are often used to quantify each remaining unlabeled instance's marginal contribution at a given iteration~\citep{Liu:2021:InfluenceActiveLearning}.
}
 
With this broad perspective on influence analysis and \revTwo{related concepts} in mind, we transition to focusing on specific influence analysis methods in the next two sections.

\section{Retraining-Based Influence Analysis}%
\label{sec:Estimators:RetrainBased}

Training instances can only influence a model if they are used during training.
As Section~\ref{sec:EstimatorOverview:PointwiseInfluence:ModernMethods} describes in the context of linear regression, one method to measure influence just trains a model with and without some instance; influence is then defined as the difference in these two models' behavior.
This basic intuition is the foundation of retraining-based influence analysis, and
this simple formulation applies to any model class -- parametric or non\=/parametric.

Observe that the retraining-based framework makes no assumptions about the learning environment.
In fact, this simplicity is one of the primary advantages of retraining-based influence.
For comparison, Table~\ref{tab:InfluenceEstimators:Comparison} shows that all gradient-based influence estimators make strong assumptions -- some of which are known not to hold for deep models (e.g.,~convexity).
However, retraining's flexibility comes at the expense of high (sometimes prohibitive) computational cost.

Below, we describe three progressively more complex retraining-based influence analysis methods.
Each method mitigates weaknesses of the preceding method -- in particular, devising techniques to make retraining-based influence more viable computationally.

\begin{remark}
  This section treats model training as \kw{deterministic} where, given a fixed training set, training always yields the same output model.
  Since the training of modern models is mostly stochastic, retraining-based estimators should be represented as expectations over different random initializations and batch orderings.
  Therefore, (re)training should be repeated multiple times for each relevant training (sub)set with a probabilistic average taken over the valuation metric~\citep{Lin:2022:InfluenceRandomizedExperiments}.
  For simplicity of presentation, expectation over randomness is dropped from the influence and influence estimator definitions below.
\end{remark}

\begin{remark}
  Section~\ref{sec:Preliminaries} defines $\trainSet$ as a supervised training set.
  The three primary retraining-based influence analysis methods detailed below also generalize to unsupervised and semi-supervised training.
\end{remark}

\begin{remark}%
  \label{rem:Estimators:RetrainBased:TimeComplexity}
  When calculating retraining's time complexity below, each training iteration's time complexity is treated as a constant cost.
  This makes the time complexity of training a single model~\bigOT{}.
  Depending on the model architecture and hyperparameter settings, a training iteration's complexity may directly depend on training-set size~$\nTr$ or model parameter count~$\dimW$.
\end{remark}

\begin{remark}
  It may be possible to avoid full model retraining by using machine unlearning methods capable of certifiably ``forgetting'' training instances~\citep{Guo:2020:CertifiedRemoval,Brophy:2021:MachineUnlearningForests,Nguyen:2022:MachineUnlearningSurvey,Eisenhofer:2022:VerifiableMachineUnlearning}.
  The asymptotic complexity of such methods is model-class specific and beyond the scope of this work.
  Nonetheless, certified deletion methods can drastically reduce the overhead of retraining-based influence analysis.
\end{remark}

\subsection{Leave-One-Out Influence}%
\label{sec:Estimators:RetrainBased:LeaveOneOut}

\kw{Leave-one-out}~(\loo) is the simplest influence measure described in this work.
\loo{} is also the oldest, dating back to \citet{Cook:1982:InfluenceRegression}  %
who term it \kw{case deletion diagnostics}.

As its name indicates, \kw{leave-one-out influence} is the change in $\zTe$'s risk due to the removal of a single instance,~$\zI$, from the training set~\citep{Koh:2017:Understanding,Black:2021:LeaveOneOut,Jia:2021:ScalabilityVsUtility}.
Formally,
\begin{equation}
  \labelAndRemember{eq:Estimators:RetrainBased:LOO}
  {%
    \baseInfFunc{\infLoo}{\zI}{\zTe}
    \defeq
      \risk{\zTe}{\wFinSub}
      -
      \risk{\zTe}{\wFin}
  }%
  \text{,}
\end{equation}
\noindent
where $\wFinSub$~are the final model parameters when training on subset ${\trainSet \setminus \zI}$ and $\wFin$ are the final model parameters trained on all of~$\trainSet$.

Measuring the entire training set's \loo{} influence requires training ${(\nTr + 1)}$ models.
Given a deterministic model class and training algorithm (e.g.,~convex model optimization~\citep{Boyd:2004:ConvexOptimization}), \loo{} is one of the few influence measures that can be computed exactly in polynomial time w.r.t.\ training-set size~$\nTr$ and iteration count~$\nItr$.

\subsubsection{Time, Space, and Storage Complexity}%
\label{sec:Estimators:RetrainBased:LeaveOneOut:Complexity}

Training a single model has time complexity~\bigOT{} (see Remark~\ref{rem:Estimators:RetrainBased:TimeComplexity}).
By additivity, training ${(\nTr + 1)}$ models has total time complexity \bigOnT{}.
Since these ${(\nTr + 1)}$ models are independent, they can be trained in parallel.

Pointwise influence analysis always has space complexity of at least~\bigOn{}, i.e.,~the space taken by the $\nTr$~influence values ${\forall_{\trIdx} \, \infFunc{\zI}{\zTe}}$.
Training a single model has space complexity~\bigOp{}, where ${\dimW \defeq \abs{\W}}$; this complexity scales linearly with the number of models trained in parallel.
Table~\ref{tab:InfluenceEstimators:Comparison} treats the level of training concurrency as a constant factor, which is why \loo{}'s total space complexity is listed as~\bigOpPln{}.

A naive implementation of \loo{} would train the $\nTr$~additional models and immediately discard them after measuring $\zTe$'s~test loss.
This simple version of \loo{} has $\bigO{1}$ storage complexity.
If instead the ${(\nTr + 1)}$~models are stored, analysis of subsequent test instances requires no additional retraining.
This drastically reduces \loo{}'s incremental time complexity for subsequent instances to just \bigOn{} forward passes -- a huge saving.%
\footnote{%
  \loo{}'s incremental computational cost can be (significantly) reduced in practice via batching.%
}
Note that this amortization of the retraining cost induces an \bigOnP{} storage complexity as listed in Table~\ref{tab:InfluenceEstimators:Comparison}.

\subsubsection{Strengths and Weaknesses}

Leave-one-out influence's biggest strength is its simplicity.
\loo{} is human-intelligible -- even by laypersons.
For that reason, \loo{} has been applied to ensure the fairness of algorithmic decisions~\citep{Black:2021:LeaveOneOut}.
\revOne{%
  Moreover, like all methods in this section, \loo{}'s simplicity allows it to be combined with any model architecture.
}

\loo{}'s theoretical simplicity comes at the price of huge upfront computational cost.
Training some state-of-the-art models from scratch even once is prohibitive for anyone beyond industrial actors~\citep{Dishkin:2021:DistributedDeepLearning}.
For even the biggest players, it is impractical to train ${(\nTr + 1)}$~such models given huge modern datasets~\citep{Basu:2021:InfluenceFunctionsFragile}.
The climate effects of such retraining also cannot be ignored~\citep{Strubell:2020:EnergyDeepLearning}.

\loo{}'s simple definition in Eq.~\eqref{eq:Estimators:RetrainBased:LOO} is premised on deterministic training.
However, even when training on the \textit{same} data, modern models may have significant predictive variance for a given test instance~\citep{Black:2021:LeaveOneOut,Wang:2023:DataBanzhaf}.
This variance makes it difficult to disentangle the effect of an instance's deletion from training's intrinsic variability~\citep{Basu:2021:InfluenceFunctionsFragile}.
For a single training instance, estimating the \loo{} influence within a standard deviation of~$\sigma$ requires training ${\Omega(1 / \sigma^2)}$ models.
Therefore, estimating the entire training set's \loo{} influence requires training ${\Omega(\nTr/\sigma^2)}$~models -- further exacerbating \loo{}'s computational infeasibility~\citep{Feldman:2020:InfluenceEstimation}.

These limitations notwithstanding, \loo{}'s impact on influence analysis research is substantial.
Many pointwise influence analysis methods either directly estimate the leave-one-out influence --
e.g.,~\feldman{} (Sec.~\ref{sec:Estimators:RetrainBased:Feldman}),
influence functions (Sec.~\ref{sec:Estimators:GradientBased:Static:IF}),
\hydra{} (Sec.~\ref{sec:Estimators:GradientBased:Dynamic:HyDRA})
--
or are very similar to \loo{} --
e.g.,~Shapley value (Sec.~\ref{sec:Estimators:RetrainBased:Shapley}).

\subsubsection{Related Methods}

\relatedHeading{Efficient Nearest-Neighbor \loo}
Although \loo{} has poor upfront and storage complexities in general, it can be quite efficient for some model classes -- particularly instance-based learners~\citep{Aha:1991:InstanceBasedLearning}.
For example, \citet{Jia:2021:ScalabilityVsUtility} propose the \kw{\knn{} leave-out-one} (\knnLOO) estimator, which calculates the \loo{}~influence over a surrogate $\kNeigh$\=/nearest neighbors classifier instead of over target model~$\dec$.
\knnLOO{} relies on a simple two-step process.
First, the features of test instance~$\zTe$ and training set~$\trainSet$ are extracted using a pretrained model.
Next, a \knn{} classifier's \loo{} influence is calculated exactly~\citep[Lemma~1]{Jia:2021:ScalabilityVsUtility} using these extracted features.
\citeauthor{Jia:2021:ScalabilityVsUtility} prove that \knnLOO{} influence only requires $\bigO{\nTr \log \nTr}$~time -- significantly faster in practice than vanilla \loo{}'s \bigOnT{} complexity.
\citeauthor{Jia:2021:ScalabilityVsUtility} also demonstrate empirically that \knnLOO{} and vanilla \loo{} generate similar influence rankings across various learning domains and tasks.

\relatedHeading{Efficient \loo{} Estimation in Decision Tree Ensembles}
\citet{Sharchilev:2018:TreeInfluence} propose \leafRefit{}, an efficient \loo{} estimator for decision-tree ensembles.
\leafRefit{}'s efficiency derives from the simplifying assumption that instance deletions do not affect the trees' structure.
In cases where this assumption holds, \leafRefit{}'s tree influence estimates are exact.
To the extent of our knowledge, \leafRefit{}'s suitability for surrogate influence analysis of deep models has not yet been explored.

\relatedHeading{Cook's Distance and Linear Regression}
For least-squares linear regression, \citet{Wojnowicz:2016:InfluenceSketching} show that each training instance's \loo{} influence on the model parameters (i.e.,~Cook's distance) can be efficiently estimated by mapping training set~$\trainSet$ into a lower-dimensional subspace.
By the Johnson\=/Lindenstrauss lemma~\citep{Johnson:1984}, these influence sketches approximately preserve the pairwise distances between the training instances in~$\trainSet$ provided the projected dimension is on the order of~${\log \nTr}$.

\relatedHeading{\loo{} Group Influence}
Leave\=/one\=/out can be extended to \kw{leave\=/$m$\=/out} for any integer ${m \leq \nTr}$.%
\footnote{%
  Leave\=/$m$\=/out influence analysis is also called \kw{multiple case deletion diagnostics}~\citep{Rousseeuw:1997:RobustRegressionOutlierDetection}.
}
Leave\=/$m$\=/out has time complexity $\bigO{\binom{\nTr}{m}}$, which is exponential in the worst case.
Shapley value influence~\citep{Shapley:1953} (Sec.~\ref{sec:Estimators:RetrainBased:Shapley}) shares significant similarity with leave\=/$m$\=/out.

As mentioned above,
\loo{} influence serves as the reference influence value for multiple influence estimators including \feldman{}, which we describe next.

\subsection{\feldman{}}%
\label{sec:Estimators:RetrainBased:Feldman}

Proposed by \citet{Feldman:2020:InfluenceEstimation}, \feldman{}%
\footnote{%
  \citet{Feldman:2020:InfluenceEstimation} do not specify a name for their influence estimator.
  Previous work has referred to \citeauthor{Feldman:2020:InfluenceEstimation}'s method as ``subsampling''~\citep{Brophy:2023:TreeInfluence} and as ``counterfactual influence''~\citep{Zhang:2021:CounterfactualMemorization}.
  We use ``\feldman{}'' to differentiate \citeauthor{Feldman:2020:InfluenceEstimation}'s method from the existing, distinct task of dataset subsampling~\citep{Ting:2018:OptimalSubsampling} while still emphasizing the methods' reliance on repeated training-set sampling.%
}
mitigates leave-one-out influence's two primary weaknesses: (1)~computational complexity dependent on~$\nTr$ and (2)~instability due to stochastic training variation.

\feldman{} relies on an ensemble of $\nFeldModel$~submodels each trained on a u.a.r.\ subset of full training set~$\trainSet$.
Let ${\feldDatasetK \,\simN{\feldDatasetSize}\, \trainSet}$ be the $\feldIdx$\=/th submodel's training set where ${\forall_{\feldIdx} \, \abs{\feldDatasetK} = \feldDatasetSize < \nTr}$.%
\footnote{%
  \citet{Feldman:2020:InfluenceEstimation} propose setting ${\feldDatasetSize = \ceil{0.7\nTr}}$.%
}
Define
${\nFeldModelI \defeq \sum_{\feldIdx = 1}^{\nFeldModel} \ind{\zI \in \feldDatasetK}}$
as the number of submodels that used instance~$\zI$ during training.
The \kw{\feldman{} pointwise influence estimator}%
\footnote{%
  \citet{Feldman:2020:InfluenceEstimation} define their estimator specifically for classification.
  \feldman{}'s definition in Eq.~\eqref{eq:Estimators:RetrainBased:Feldman} uses a more general form to cover additional learning tasks such as regression.
  \citeauthor{Feldman:2020:InfluenceEstimation}'s original formulation would be equivalent to defining the risk as ${\risk{\zTe}{\wFeldK} = \indBig{\yTe \ne \decFunc{\xTe}{\wFeldK}}}$, i.e., the accuracy subtracted from one.%
}
is then
\begin{equation}%
  \labelAndRemember{eq:Estimators:RetrainBased:Feldman}{%
    \baseInfFunc{\infEstFeld}{\zI}{\zTe}
      \defeq
      \frac{1}
           {\nFeldModel - \nFeldModelI}
      \sum_{\substack{\feldIdx \\ \zI \notin \feldDatasetK}}
        \risk{\zTe}{\wFeldK}
      -
      \frac{1}
           {\nFeldModelI}
      \sum_{\substack{\feldIdxAlt \\ \zI \in \feldDatasetKAlt}}
        \risk{\zTe}{\wFeldKAlt}
  }%
  \text{.}
\end{equation}
\noindent
Intuitively, Eq.~\eqref{eq:Estimators:RetrainBased:Feldman} is the change in $\zTe$'s average risk when $\zI$ is not used in submodel training.
By holding out multiple instances simultaneously and then averaging, each \feldman{} submodel provides insight into the influence of \textit{all} training instances.
This allows \feldman{} to require (far) fewer retrainings than \loo{}.

Since each of the $\nFeldModel$~training subsets is i.i.d., then
\begin{equation}\label{eq:Estimators:RetrainBased:Feldman:Expectation}
  \lim_{\nFeldModel \rightarrow \infty}
  \baseInfFunc{\infEstFeld}{\zI}{\zTe}
    =
    \expectS{\trainSetAltB \,\simN{\feldDatasetSize}\, \trainSetNoZi}%
            {\risk{\zTe}{\wFinSubBase{\trainSetAltB}}}
    -
    \expectS{\trainSetAlt \,\simN{\feldDatasetSize - 1}\, \trainSetNoZi}%
            {\risk{\zTe}{\wFinSubBase{\trainSetAltPlusZi}}}
    \text{.}%
\end{equation}%
\noindent%
For sufficiently large~$\feldDatasetSize$ and~$\nTr$, the expected behavior of a model trained on an i.i.d.\ dataset of size~$m$ becomes indistinguishable from one trained on ${m-1}$~i.i.d.\ instances.
Applying this property along with linearity of expectation and Eq.~\eqref{eq:Estimators:RetrainBased:LOO}, Eq.~\eqref{eq:Estimators:RetrainBased:Feldman:Expectation} reformulates as
{
  \begin{align}\label{eq:Estimators:RetrainBased:Feldman:ExpectedLOO}
    \lim_{\nFeldModel,\nTr,\feldDatasetSize \rightarrow \infty}
    \baseInfFunc{\infEstFeld}{\zI}{\zTe}
      &=
      \expectS{\trainSetAltB \,\simN{\feldDatasetSize - 1}\, \trainSetNoZi}%
              {\risk{\zTe}{\wFinSubBase{\trainSetAltB}}}
      -
      \expectS{\trainSetAlt \,\simN{\feldDatasetSize - 1}\, \trainSetNoZi}%
              {\risk{\zTe}{\wFinSubBase{\trainSetAltPlusZi}}}
      \\
      &= \expectS{\trainSetAlt \,\simN{\feldDatasetSize - 1}\, \trainSetNoZi}%
                 {%
                   \risk{\zTe}{\wFinSubBase{\trainSetAlt}}
                   -
                   \risk{\zTe}{\wFinSubBase{\trainSetAltPlusZi}}
                 }  \\
      &= \expectS{\trainSetAlt \,\simN{\feldDatasetSize - 1}\, \trainSetNoZi}
                 {\infLooITe\vphantom{\bigg[}}
      \text{.}
  \end{align}%
}%

Hence, \feldman{} is a statistically \kw{consistent} estimator of the \kw{expected \loo{} influence}.
This means that \feldman{} does not estimate the influence of training instance~$\zI$ on a single model instantiation.
Rather, \feldman{} estimates $\zI$'s \emph{influence on the training algorithm and model architecture as a whole}.
By considering influence in expectation, \feldman{} addresses \loo{}'s inaccuracy caused by stochastic training's implicit variance.

In practice, $\nFeldModel$, $\nTr$, and~$\feldDatasetSize$ are finite. Nonetheless, \citet[Lemma~2.1]{Feldman:2020:InfluenceEstimation} prove that, with high probability, \feldman{}'s \loo{} influence estimation error is bounded given~$\nFeldModel$ and~$\frac{\feldDatasetSize}{\nTr}$.

\revOne{%
While \feldman{}'s formulation above is w.r.t.\ a single training instance,
\feldman{} trivially extends to estimate the \kw{expected group influence} of multiple training instances.
Observe however that the expected fraction of u.a.r.\ training subsets that either contains all instances in a group or none of a group decays geometrically with ${\frac{\feldDatasetSize}{\nTr}}$ and ${(1 - \frac{\feldDatasetSize}{\nTr})}$, respectively.
Therefore, for large group sizes, $\nFeldModel$ needs to be exponential in~$\nTr$ to cover sufficient group combinations.
}

\begin{remark}
  \label{remark:Estimators:RetrainBased:Downsampling:Assumption}
  \feldman{} trains models on data subsets of size~$\feldDatasetSize$ under the assumption that statements made about those models generalize to models trained on a dataset of size~$\nTr$ (i.e.,~all of~$\trainSet$).
  This assumption may not hold for small~$\feldDatasetSize$.
  To increase the likelihood this assumption holds, \citeauthor{Feldman:2020:InfluenceEstimation} propose fixing ${\frac{\feldDatasetSize}{\nTr} = 0.7}$.
  This choice balances satisfying the aforementioned assumption against the number of submodels since \feldman{} requires $\nFeldModel$ and ratio~$\frac{\feldDatasetSize}{\nTr}$ combined dictate $\nFeldModelI$, i.e., the number of submodels that are trained on~$\zI$.
\end{remark}

\subsubsection{Time, Space, and Storage Complexity}%
\label{sec:Estimators:RetrainBased:Feldman:Complexity}

\feldman{}'s complexity analysis is identical to that of \loo{} (Sec.~\ref{sec:Estimators:RetrainBased:LeaveOneOut:Complexity}) except, instead of the time and storage complexities being dependent on training-set size~$\nTr$, \feldman{} depends on submodel count~$\nFeldModel$.
For perspective, \citeauthor{Feldman:2020:InfluenceEstimation}'s empirical evaluation used ${\nFeldModel = 2,000}$ for ImageNet~\citep{Deng:2009:ImageNet} (${\nTr > 14\text{M}}$) as well as ${\nFeldModel = 4,000}$ for MNIST~\citep{MNIST} (${\nTr = 60,000}$) and CIFAR10~\citep{CIFAR10} (${\nTr = 50,000}$)  --  a savings of one to four orders of magnitude over vanilla~\loo{}.

\begin{remark}
  \feldman{}'s incremental time complexity is technically ${\bigO{\nFeldModel + \nTr} \in \bigO{\nTr}}$ since pointwise influence is calculated w.r.t.\ each training instance.
  The time complexity analysis above focuses on the difference in the number of forward passes required by \loo{} and \feldman{}; fewer forward passes translate to \feldman{} being much faster than \loo{} in practice.
\end{remark}

\subsubsection{Strengths and Weaknesses}

Although more complicated than \loo{}, \feldman{} is still comparatively simple to understand and implement.
\feldman{} makes only a single assumption that should generally hold in practice (see Remark~\ref{remark:Estimators:RetrainBased:Downsampling:Assumption}).
\feldman{} is also flexible and can be applied to most applications.

Another strength of \feldman{} is its low incremental time complexity.
Each test example requires only $\nFeldModel$~forward passes. %
These forward passes can use large batch sizes to further reduce the per-instance cost.
This low incremental cost allows \feldman{} to be applied at much larger scales than other methods.
For example, \citet{Feldman:2020:InfluenceEstimation} measure all pointwise influence estimates for the entire ImageNet dataset~\citep{Deng:2009:ImageNet} (${\nTr > 14\text{M}}$).
These large-scale experiments enabled \citeauthor{Feldman:2020:InfluenceEstimation} to draw novel conclusions about neural training dynamics -- including that training instance memorization~\eqref{eq:EstimatorOverview:Memorization} by overparameterized models is not a bug, but a feature, that is currently necessary to achieve state-of-the-art generalization results.

In terms of weaknesses, while \feldman{} is less computationally expensive than \loo{}, \feldman{} still has a high upfront computational cost.
Training multiple models may be prohibitively expensive even when ${\nFeldModel \ll \nTr}$.
Amortization of this upfront training overhead across multiple test instances is beneficial but by no means a panacea.

\subsubsection{Related Methods}

\feldman{} has two primary related methods.

\relatedHeading{Generative \feldman{}}%
Training instance memorization also occurs in generative models where the generated outputs are (nearly) identical copies of training instances~\citep{Kandpal:2022:DeduplicatingPrivacyRisks}.
\citet{Van:2021:Memorization} extend \feldman{} to deep generative models -- specifically, density models (e.g., variational autoencoders~\citep{Kingma:2014:VAE,Rezende:2014:DeepGenerativeModels}) that estimate posterior probability~${p(\X \vert \trainDataDist, \W)}$, where $\trainDataDist$ denotes the training data distribution.
Like \feldman{}, \citeauthor{Van:2021:Memorization}'s approach relies on training multiple submodels.%
\footnote{%
  Rather than training submodels using i.i.d.\ subsets of~$\trainSet$, \citet{Van:2021:Memorization} propose training the submodels via repeated $d$\=/fold cross-validation.
  While technically different, \citeauthor{Van:2021:Memorization}'s approach is functionally equivalent to \citepos{Feldman:2020:InfluenceEstimation} u.a.r.\ sampling procedure.%
}
The primary difference is that \citeauthor{Van:2021:Memorization} consider generative risk
\begin{equation}
  \risk{\xTe}{\wFin}
  =
    -
    \log p(\xTe \vert \trainDataDist, \wFin)
  \text{.}
\end{equation}
\noindent%
Beyond that, \citeauthor{Van:2021:Memorization}'s method is the same as \feldman{} as both methods consider the \loo{} influence~\eqref{eq:Estimators:RetrainBased:Feldman}.

\relatedHeading{Consistency Profile and Score}%
  \feldman{}'s second closely related method is \citepos{Jiang:2021:CScore} \kw{consistency profile}, defined formally as%
\footnote{%
  \citet{Jiang:2021:CScore} define their estimator specifically for classification. We present their method more generally to apply to other losses/domains (e.g.,~regression).
  As with \feldman{}, it is trivial to map between Eq.~\eqref{eq:Estimators:Retrain:Feldman:ConsistencyProfile}'s formulation and that of \citeauthor{Jiang:2021:CScore}%
}
\begin{equation}\label{eq:Estimators:Retrain:Feldman:ConsistencyProfile}
  \cProfile
    \defeq
      - \expectS{\trainSetAlt \,\simN{\cTrSize}\, \trainSet}{\risk{\zTe}{\wFinSubBase{\trainSetAlt}}}
      \text{.}
\end{equation}
\noindent
By negating the risk in Eq.~\eqref{eq:Estimators:Retrain:Feldman:ConsistencyProfile}, a higher expected risk corresponds to a lower consistency profile.
Consistency profile differs from \feldman{} in two ways. (1)~\feldman{} implicitly considers a single submodel training-set size~$\feldDatasetSize$ while
consistency profile disentangles the estimator from $\feldDatasetSize$.
(2)~\feldman{} estimates $\zI$'s influence on $\zTe$ while consistency profile considers all of $\trainSet$ as a group and estimates
the expected group influence of a random subset ${\trainSetAlt \subseteq \trainSet}$ given ${\feldDatasetSize \defeq \abs{\trainSetAlt}}$.

\citet{Jiang:2021:CScore} also propose the \kw{consistency score} (\cScoreName), defined formally as
\begin{equation}\label{eq:Estimators:Retrain:Feldman:ConsistencyScore}
  \cScore
  \defeq
    \expectS{\cTrSize \,\simOne\, \setint{\nTr}}%
            {\cProfile}
  \text{,}
\end{equation}
\noindent
where $\cTrSize$ is drawn uniformly from set~$\setint{\nTr}$.
By taking the expectation over all training-set sizes, \cScoreName{} provides a total ordering over all test instances.
A large \cScoreName{} entails that $\zTe$ is harder for the model to confidently predict.
Large \cScoreName{}s generally correspond to rare/atypical test instances from the tails of the data distribution.
Since \feldman{} considers the effect of each training instance individually, \feldman{} may be unable to identify these hard\=/to\=/predict test instances -- in particular if $\feldDatasetSize$ is large enough to cover most data distribution modes.

The next section introduces the Shapley value, which in essence merges the ideas of \feldman{} and \cScoreName{}.

\subsection{Shapley Value}%
\label{sec:Estimators:RetrainBased:Shapley}

Derived from cooperative game theory,
\kw{Shapley value}~(\sv{}) quantifies the increase in value when a group of players cooperates to achieve some shared objective \citep{Shapley:1953,Shapley:1988:ShapleyValue}.
Given $\nTr$~total players, \kw{characteristic function} $\func{\shapValSym}{\powerSetInt{\nTr}}{\real}$ defines the value of any player coalition ${\shapCoalition \subseteq \setint{\nTr}}$
\citep{Dubey:1975:ShapleyValueUniqueness}.
By convention, a larger $\shapValFunc{\shapCoalition}$ is better.
Formally, player $\trIdx$'s Shapley value w.r.t.\ $\shapValSym$ is
\begin{equation}%
  \label{eq:Estimators:RetrainBased:Shapley:Base}
  \shapValComplete{\trIdx}%
                  {\shapValSym}%
    \defeq
      \frac{1}%
           {\nTr}
      \sum_{\shapCoalition \subseteq \setint{\nTr} \setminus \trIdx}
          \frac{1}%
               {\binom{\nTr - 1}{\abs{\shapCoalition}}}
          \sbrack{
              \shapValFunc{\shapCoalition \cup \trIdx}
              -
              \shapValFunc{\shapCoalition}%
              \vphantom{\Big\vert}
          }
  \text{,}
\end{equation}
\noindent
where $\binom{\nTr - 1}{\abs{\shapCoalition}}$ is the binomial coefficient.
\citet{Ghorbani:2019:DataShapley} adapt \sv{} to model training by treating the $\nTr$~instances in training set $\trainSet$ as $\nTr$~cooperative players with the shared objective of training the ``best'' model.%
\footnote{%
  ``Best'' here is w.r.t.\ some data valuation measure of interest, with different use cases potentially defining ``best'' differently.%
}
For any ${\shapCoalition \subseteq \trainSet}$, let ${\shapValFunc{\shapCoalition} \defeq -\risk{\zTe}{\wFinSubBase{\shapCoalition}}}$ where the negation is needed because more ``valuable'' training subsets have lower risk.
Then, $\zI$'s \kw{Shapley value pointwise influence} on $\zTe$ is
\begin{equation}%
  \labelAndRemember{eq:Estimators:RetrainBased:ShapleyInfluence}{%
    \baseInfFunc{\infSV}{\zI}{\zTe}
        \defeq
        \frac{1}%
             {\nTr}
        \sum_{\trainSetAlt \subseteq \trainSetNoZi}
          \frac{1}%
               {%
                  \binom{\nTr - 1}%
                        {\abs{\trainSetAlt}}%
               }
               \sbrack{
                  \risk{\zTe}%
                       {\wFinSubBase{\trainSetAlt}}
                  -
                  \risk{\zTe}%
                       {\wFinSubBase{\trainSetAltPlusZi}}%
                  \vphantom{\Big\vert}
               }%
  }
  \text{.}
\end{equation}
\noindent%
More intuitively, \sv{} is the weighted change in $\zTe$'s risk when $\zI$ is added to a random training subset; the weighting ensures all training subset sizes~($\abs{\trainSetAlt}$) are prioritized equally.
Eq.~\eqref{eq:Estimators:RetrainBased:ShapleyInfluence} can be viewed as generalizing the leave\=/one\=/out influence, where rather than considering only full training set~$\trainSet$, Shapley value averages the \loo{} influence across all possible subsets of~$\trainSet$.

\revOne{%
  There exists multiple extensions of Shapley value to the group context~\citep{Bordt:2023:ShapleyValuesGAM,Tsai:2023:FaithShap,Grabisch:1999:InteractionAmongPlayers,Sundararajan:2020:ShapleyTaylorInteraction}.%
\footnote{%
  Most of these works were proposed in the context of studying the interaction between groups of features.
  Directly adapting these ideas to groups of training instances is straightforward.%
}
The most well-known method is \citepos{Grabisch:1999:InteractionAmongPlayers} \kw{Shapley interaction index}, which for any subset ${\shapCoalition \subseteq \trainSet}$, is defined as
\begin{equation}
  \labelAndRemember{eq:Estimators:RetrainBased:Shapley:InteractionIndex}%
  {%
    \baseInfFunc{\infSV}{A}{\zTe}
        \defeq
        -
        \sum_{\trainSetAlt \subseteq \trainSet \setminus A}
          \frac{(\nTr - \abs{A} - \abs{\trainSetAlt})! \,\, \abs{\trainSetAlt}!}%
               {(\nTr - \abs{A} + 1)!}
          \sum_{\trainSetAltB \subseteq A}
            \left(
              -1
            \right)^{\abs{A} - \lvert \trainSetAltB \rvert}
              \risk{\zTe}%
                   {\wFinSubBase{\trainSetAlt \cup \trainSetAltB}}
  }
  \text{.}
\end{equation}
\noindent
To intuitively understand Shapley interaction index's inner summation, consider when $A$ consists of only two instances (e.g., ${A = \myset{\zI, \zIAlt}}$), then
\begin{align}%
  \sum_{\trainSetAltB \subseteq \myset{\zI, \zIAlt}}
    \left(
      -1
    \right)^{2 - \lvert \trainSetAltB \rvert}
      \risk{\zTe}%
           {\wFinSubBase{\trainSetAlt \cup \trainSetAltB}}
  =
    &\risk{\zTe}%
         {\wFinSubBase{\trainSetAlt}}
    -
    \risk{\zTe}%
         {\wFinSubBase{\trainSetAlt \cup \zI}}
    \\\nonumber
    &-
    \risk{\zTe}%
         {\wFinSubBase{\trainSetAlt \cup \zIAlt}}
    +
    \risk{\zTe}%
         {\wFinSubBase{\trainSetAlt \cup \myset{\zI, \zIAlt}}}
  \text{.}
\end{align}
\noindent%
\citet{Grabisch:1999:InteractionAmongPlayers} explain that a positive interaction index entails that the combined group influence of the training instances in ${A \subseteq \trainSet}$ exceeds the sum of their marginal influences;
similarly, a negative interaction index means that $A$'s group influence is less than the sum of their marginal influences.
If the interaction index is zero, $A$'s members have no net interaction. %
\citet{Sundararajan:2020:ShapleyTaylorInteraction} propose an alternate Shapley group formulation they term the \kw{Shapley-Taylor interaction index}, which is defined as
\begin{equation}%
  \labelAndRemember{eq:Estimators:RetrainBased:Shapley:Taylor}%
  {%
    \baseInfFunc{\infST}{A}{\zTe}
        \defeq
        -
        \frac{\abs{A}}%
             {\nTr}
        \sum_{\trainSetAlt \subseteq \trainSet \setminus A}
          \frac{1}%
               {%
                  \binom{\nTr - 1}%
                        {\abs{\trainSetAlt}}%
               }%
          \sum_{\trainSetAltB \subseteq A}
            \left(
              -1
            \right)^{\abs{A} - \lvert \trainSetAltB \rvert}
              \risk{\zTe}%
                   {\wFinSubBase{\trainSetAlt \cup \trainSetAltB}}
  }
  \text{.}
\end{equation}
\noindent%
The Shapley and Shapley-Taylor interaction indices provide different mathematical guarantees the details of which extend beyond the scope of this work.
We refer the reader to \citet{Sundararajan:2020:ShapleyTaylorInteraction} for additional discussion.%
}

\subsubsection{Time, Space, and Storage Complexity}%
\label{sec:Estimators:RetrainBased:Shapley:Complexity}

\citet{Deng:1994:ComplexityCooperativeGames} prove that computing Shapley values is \#P\=/complete.
Therefore, in the worst case, \sv{} requires exponential time to determine exactly assuming ${\text{P} \ne \text{NP}}$.
There has been significant follow-on work to develop tractable \sv{} estimators, many of which are reviewed in Sec.~\ref{sec:Estimators:RetrainBased:Shapley:RelatedMethods}.

The analysis of \sv{}'s space, storage, and incremental time complexities follows that of \loo{} (Sec.~\ref{sec:Estimators:RetrainBased:LeaveOneOut:Complexity}) with the exception that \sv{} requires up to $2^{\nTr}$~models, not just $\nTr$~models as with \loo{}.

\subsubsection{Strengths and Weaknesses}

Among all influence analysis methods, \sv{} may have the strongest theoretical foundation with the chain of research extending back several decades.
\sv{}'s dynamics and limitations are well understood, providing confidence in the method's quality and reliability.
In addition, \sv{} makes minimal assumptions about the nature of the cooperative game (i.e.,~model to be trained), meaning \sv{} is very flexible.
This simplicity and flexibility allow \sv{} to be applied to many domains beyond dataset influence as discussed in the next section.

\revOne{%
\sv{} has been shown to satisfy multiple appealing mathematical axioms.
First, \sv{} satisfies the \kw{dummy player} axiom where
\begin{equation}%
  \label{eq:Estimators:RetrainBased:Shapley:DummyPlayer}%
  \mathlarger{\forall}_{\trainSetAlt \subseteq \trainSet \setminus \zI}
    \,\,
      \risk{\zTe}%
           {\wFinSubBase{\trainSetAlt \cup \zI}}
    =
      \risk{\zTe}%
           {\wFinSubBase{\trainSetAlt}}
  \implies
      \baseInfFunc{\infSV}%
                  {\zI}%
                  {\zTe}
      =
      0
      \text{.}
\end{equation}
\noindent%
Second, \sv{} is \kw{symmetrical} meaning for any ${\zI, \zIAlt \in \trainSet}$,
\begin{equation}%
  \label{eq:Estimators:RetrainBased:Shapley:Symmetric}%
  \mathlarger{\forall}_{\trainSetAlt \subseteq \trainSet \setminus \myset{\zI, \zIAlt}}
    \,\,
    \risk{\zTe}%
         {\wFinSubBase{\trainSetAlt \cup \zI}}
    =
    \risk{\zTe}%
         {\wFinSubBase{\trainSetAlt \cup \zIAlt}}
  \implies
    \baseInfFunc{\infSV}{\zI}{\zTe}
    =
    \baseInfFunc{\infSV}{\zIAlt}{\zTe}
  \text{.}
\end{equation}
\noindent%
Third, \sv{} is \kw{linear}~\citep{Shapley:1953,Dubey:1975:ShapleyValueUniqueness}, meaning given any two data valuation metrics~${\shapValSym', \shapValSym''}$ and ${\alpha', \alpha'' \in \real}$, it holds that
\begin{equation}%
  \label{eq:Estimators:RetrainBased:Shapley:Linearity}
  \shapValComplete{\trainSetAlt}%
                  {\alpha' \shapValSym' + \alpha'' \shapValSym''}
      =
        \alpha' \, \shapValComplete{\trainSetAlt}{\shapValSym'}
        +
        \alpha'' \, \shapValComplete{\trainSetAlt}{\shapValSym''} \text{.}
\end{equation}
\noindent
\sv{}'s linearity axiom makes it possible to estimate both pointwise and joint \sv{} influences without repeating any data collection~\citep{Ghorbani:2019:DataShapley}.
Any data value satisfying the dummy player, symmetry, and linearity axioms is referred to as a \kw{semivalue}~\citep{Dubey:1981:Semivalues,Kwon:2022:BetaShapley}.
Additional semivalues include leave-one-out (Sec.~\ref{sec:Estimators:RetrainBased:LeaveOneOut}) and Banzhaf value (Sec.~\ref{sec:Estimators:RetrainBased:Shapley:RelatedMethods})~\citep{Banzhaf:1965:WeightedVoting}.%
}

Furthermore, by evaluating training sets of different sizes, \sv{} can detect subtle influence behavior that is missed by methods like \feldman{} and \loo{}, which evaluate a single training-set size.
\citet{Lin:2022:InfluenceRandomizedExperiments} evidence this phenomenon empirically showing that \revOne{adversarial training instances (i.e., poison)} can sometimes be better detected with small \sv{} training subsets.

Concerning weaknesses, \sv{}'s computational intractability is catastrophic for non-trivial dataset sizes~\citep{Kwon:2022:BetaShapley}.
For that reason, numerous (heuristic) \sv{} speed-ups have been proposed, with the most prominent ones detailed next.

\subsubsection{Related Methods}%
\label{sec:Estimators:RetrainBased:Shapley:RelatedMethods}

\relatedHeading{Monte Carlo Shapley}
\citet{Ghorbani:2019:DataShapley} propose two \sv{} estimators.
First, \kw{truncated Monte Carlo Shapley} (\mcShap{}) relies on randomized subset sampling from training set~$\trainSet$.\footnote{The term ``truncated'' in \mcShap{} refers to a speed-up heuristic used when estimating the pointwise influences of multiple training instances at the same time. Truncation is not a necessary component of the method and is not described here.}
As a simplified description of the algorithm, \mcShap{} relies on random permutations of $\trainSet$; for simplicity, denote the permutation ordering ${\zOne, \ldots, \zFin}$.
For each permutation, $\nTr$~models are trained where the $\trIdx$\=/th model's training set is instances $\myset{\zOne, \ldots, \zI}$.
To measure each $\zI$'s marginal contribution for a given permutation, \mcShap{} compares the performance of the ${(\trIdx - 1)}$\=/th and $\trIdx$\=/th models, i.e., the models trained on datasets $\myset{\zOne, \ldots, \zIOne}$ and $\myset{\zOne, \ldots, \zI}$, respectively.
\mcShap{} generates additional training-set permutations and trains new models until the \sv{} estimates converge.
\citeauthor{Ghorbani:2019:DataShapley} state that \mcShap{} convergence usually
requires analyzing on the order of $\nTr$~training-set permutations.
Given training each model has time complexity \bigOT{} (Remark~\ref{rem:Estimators:RetrainBased:TimeComplexity}), \mcShap{}'s full time complexity is in general \bigOnSqT{}.

\relatedHeading{Gradient Shapley}
\mcShap{} may be feasible for simple models that are fast to train.
For more complex systems, \bigOnSq{} complete retrainings are impractical.
\citet{Ghorbani:2019:DataShapley} also propose \kw{Gradient Shapley}~(\gShap{}), an even faster \sv{} estimator which follows the same basic procedure as \mcShap{} with one critical change.
Rather than taking $\nItr$~iterations to train each model, \gShap{} assumes models are trained in just \textit{one} gradient step.
This means that \gShap{}'s full and incremental time complexity is only \bigOnSq{} -- a substantial speed-up over \mcShap{}'s full \bigOnSqT{} complexity.\footnote{%
  \gShap{} and \mcShap{} have the same incremental time complexity -- \bigOnSq{}.
  Both estimators' storage complexity is $\bigO{\nTr^2 \dimW}$.%
}
There is no free lunch, and \gShap{}'s speed-up is usually at the expense of lower influence estimation accuracy.

\relatedHeading{Efficient Nearest-Neighbor Shapley}
Since \mcShap{} and \gShap{} rely on heuristics and assumptions to achieve tractability, neither method provides approximation guarantees.
In contrast,
\citet{Jia:2019:KnnShapley} prove that for $\kNeigh$\=/nearest neighbors classification, \sv{} pointwise influence can be calculated exactly in $\bigO{\nTr \lg \nTr}$~time.
Formally, \sv{}'s characteristic function for \knn{} classification is
\begin{equation}\label{eq:Estimators:RetrainBased:Shapley:Knn:CharacteristicFunc}
  \shapValSym_{\kNeigh\text{NN}}(\trainSetAlt)
    \defeq
      -
      \frac{1}%
           {\kNeigh}
      \sum_{\yI \in \neighAlt{\xTe}{\trainSetAlt}}
      \ind{\yI = \yTe}
  \text{,}
\end{equation}
where $\neighAlt{\xTe}{\trainSetAlt}$ is the set of $\kNeigh$~neighbors in~$\trainSetAlt$ nearest to~$\xTe$ and $\ind{\cdot}$ is the indicator function. %
Each training instance either has no effect on Eq.~\eqref{eq:Estimators:RetrainBased:Shapley:Knn:CharacteristicFunc}'s value (${\zI \notin \neighAlt{\xTe}{\trainSetAlt}}$ or ${\yI \ne \yTe}$).
Otherwise, the training instance increases the value by one (${\zI \in \neighAlt{\xTe}{\trainSetAlt}}$ and ${\yI = \yTe}$).

Assuming the training instances are sorted by increasing distance from $\xTe$ (i.e.,~$\xI[1]$ is closest to~$\xTe$ and $\xI[\nTr]$ is furthest), then $\zI$'s pointwise \kw{\knnShap{} influence} is
\begin{equation}
  \labelAndRemember{eq:Estimators:RetrainBased:Shapley:Knn:Influence}{%
    \baseInfFunc{\infKnnSV}{\zI}{\zTe}
        \defeq
          \frac{\ind{\yI[\nTr] = \yTe}}%
               {\nTr}%
          + %
          \sum_{\trIdxAlt = \trIdx}^{\nTr - 1}%
          \frac{\ind{\yI[\trIdxAlt] = \yTe} - \ind{\yI[\trIdxAlt + 1] = \yTe}}%
                   {\kNeigh}
                   \frac{\min\myset{\kNeigh,\trIdxAlt}}
                   {\trIdxAlt}
  }
  \text{.}
\end{equation}
\noindent
Observe that $\infKnnSV$'s closed form is linear in~$\nTr$ and requires no retraining at all.
In fact, \knnShap{}'s most computationally expensive component is sorting the training instances by distance from~$\xTe$.
Similar to \knnLOO{} above, \citet{Jia:2019:KnnShapley} propose using \knnShap{} as a surrogate \sv{} estimator for more complex model classes.
For example, \knnShap{} could be applied to the feature representations generated by a deep neural network.

\relatedHeading{Beta Shapley}
Recent work has also questioned the optimality of \sv{} assigning uniform weight to each training subset size (see Eq.~\eqref{eq:Estimators:RetrainBased:Shapley:Base}).
Counterintuitively, \citet{Kwon:2022:BetaShapley} show theoretically and empirically that influence estimates on larger training subsets are \textit{more} affected by training noise than influence estimates on smaller subsets.
As such, rather than assigning all data subset sizes~($\abs{\trainSetAlt}$) uniform weight, \citeauthor{Kwon:2022:BetaShapley} argue that smaller training subsets should be prioritized.
Specifically, \citeauthor{Kwon:2022:BetaShapley} propose \kw{\betaShap}, which modifies vanilla \sv{} by weighting the training-set sizes according to a positive skew (i.e., left-leaning) beta distribution.

\sv{} has also been applied to study other types of influence beyond training set membership.
For example, \kw{Neuron Shapley} applies \sv{} to identify the model neurons that are most critical for a given prediction~\citep{Ghorbani:2020:NeuronShapley}.
\citepos{Lundberg:2017:SHAP} \SHAP{} is a very well-known tool that applies \sv{} to measure feature importance.
For a comprehensive survey of Shapley value applications beyond training data influence, see the work of \citet{Sundararajan:2020:ManyShapleyValues} and a more recent update by \citet{Rozemberczki:2022:ShapleyValueSurvey}.

\relatedHeading{Banzhaf Value}
\revOne{%
  Also a semivalue~\citep{Dubey:1981:Semivalues}, \kw{Banzhaf value}~\citep{Banzhaf:1965:WeightedVoting} is closely related to Shapley value.
Formally, the Banzhaf value influence of ${\zI \in \trainSet}$ on test instance $\zTe$ is
\begin{equation}%
  \labelAndRemember{eq:Estimators:RetrainBased:Shapley:Banzhaf}{%
    \baseInfFunc{\infBanzhaf}{\zI}{\zTe}
      \defeq
        \frac{1}%
             {2^{\nTr - 1}}
        \sum_{\trainSetAlt \subseteq \trainSetNoZi}
          \risk{\zTe}%
               {\wFinSubBase{\trainSetAlt}}
          -
          \risk{\zTe}%
               {\wFinSubBase{\trainSetAltPlusZi}}
  }
  \text{.}
\end{equation}
\noindent
Intuitively, the primary differences between Eqs.~\eqref{eq:Estimators:RetrainBased:Shapley:Base} and~\eqref{eq:Estimators:RetrainBased:Shapley:Banzhaf} is that Shapley value assigns each subset \textit{size}~($\abs{\trainSetAlt}$) equal weight while Banzhaf value assigns each \textit{subset}~($\trainSetAlt$) equal weight.
\citet{Wang:2023:DataBanzhaf} prove that influence rankings based on Banzhaf value are more robust to training variance than both leave\=/one\=/out and Shapley value.
\citeauthor{Wang:2023:DataBanzhaf} also empirically demonstrate that Banzhaf value can (significantly) outperform \sv{} in practice.

Like \sv{}, Banzhaf value has exponential time complexity.
Uniform, Monte Carlo sampling from power set~$\powerSet{\trainSet \setminus \zI}$ provides an unbiased estimate of Eq.~\eqref{eq:Estimators:RetrainBased:Shapley:Banzhaf}.
\citet{Wang:2023:DataBanzhaf} provide a more sophisticated Banzhaf value Monte Carlo sampling strategy they term \kw{maximum sample reuse}~(MSR).
MSR improves the estimates' \kw{sample complexity} by a factor of \bigOn{} over uniform Monte Carlo.
}
\section{Gradient-Based Influence Estimation}%
\label{sec:Estimators:GradientBased}

For modern models, retraining even a few times to tune hyperparameters is very expensive.
In such cases, it is prohibitive to retrain an entire model just to gain insight into a single training instance's influence.

For models trained using gradient descent, training instances only influence a model through training gradients.
Intuitively then, training data influence should be measurable when the right training gradients are analyzed.
This basic idea forms the basis of gradient-based influence estimation.
As detailed below, gradient-based influence estimators rely on Taylor-series approximations or risk stationarity.
These estimators also assume some degree of differentiability -- either of just the loss function~\citep{Yeh:2018:Representer} or both the model and loss~\citep{Koh:2017:Understanding,Pruthi:2020:TracIn,Chen:2021:Hydra}.

The exact analytical framework each gradient-based method employs depends on the set of model parameters considered~\citep{Hammoudeh:2022:GAS}.
\kw{Static, gradient-based methods} -- discussed first -- estimate the effect of retraining by studying gradients w.r.t.\ final model parameters~$\wFin$.
Obviously, a single set of model parameters provide limited insight into the entire optimization landscape, meaning static methods generally must make stronger assumptions.
In contrast, \kw{dynamic, gradient-based influence estimators} reconstruct the training data's influence by studying model parameters throughout training, e.g.,~${\wZero, \ldots, \wFin}$.
Analyzing these intermediary model parameters makes dynamic methods more computationally expensive in general, but it enables dynamic methods to make fewer assumptions.

This section concludes with a discussion of a critical limitation common to all existing gradient-based influence estimators -- both static and dynamic.
This common weakness can cause gradient-based estimators to systematically overlook highly influential (groups of) training instances.

\subsection{Static, Gradient-Based Influence Estimation}%
\label{sec:Estimators:GradientBased:Static}

As mentioned above, static estimators are so named because they measure influence using only final model parameters~$\wFin$.
Static estimators' theoretical formulations assume \kw{stationarity} (i.e.,~the model parameters have converged to a risk minimizer) and \kw{convexity}.

Below we focus on two static estimators -- \kw{influence functions}~\citep{Koh:2017:Understanding} and \kw{representer point}~\citep{Yeh:2018:Representer}.
Each method takes very different approaches to influence estimation with the former being more general and the latter more scalable.
Both estimators' underlying assumptions are generally violated in deep networks.%

\subsubsection{Influence Functions}%
\label{sec:Estimators:GradientBased:Static:IF}

Along with Shapley value (Sec.~\ref{sec:Estimators:RetrainBased:Shapley}),
\citepos{Koh:2017:Understanding} \kw{influence functions} is one of the best-known influence estimators.
The estimator derives its name from influence functions (also known as \kw{infinitesimal jackknife}~\citep{Jaeckel:1972:InfinitesimalJackknife}) in robust statistics~\citep{Hampel:1974:InfluenceCurve}.
These early statistical analyses consider how a model changes if training instance~$\zI$'s weight is infinitesimally perturbed by~${\epsI}$.
More formally, consider the change in the \kw{empirical risk minimizer} from
\begin{equation}\label{eq:Estimators:GradientBased:Static:IF:ErmBase}
  \wFin
    =
    \argmin_{\W}
    \frac{1}{\nTr}
    \sum_{\zSym \in \trainSet}
      \risk{\zSym}{\W}
\end{equation}
\noindent
to
\begin{equation}\label{eq:Estimators:GradientBased:Static:IF:ErmPerturb}
  \wFinEpsI
    =
    \argmin_{\W}
    \frac{1}{\nTr}
    \sum_{\zSym \in \trainSet}
      \risk{\zSym}{\W}
    +
    \epsI \risk{\zI}{\W}
  \text{.}
\end{equation}

Under the assumption that model~$\dec$ and loss function~$\loss$ are twice-differentiable and strictly convex, \citet{Cook:1982:InfluenceRegression} prove via a first-order Taylor expansion that
\begin{equation}\label{eq:Estimators:GradientBased:Static:ParamDeriv}
  \frac{d \wFinEpsI}%
       {d \epsI}
  \vertEpsIzero
    =
    -\invHessFin
    \gradW
    \risk{\zI}{\wFin}
  \text{,}
\end{equation}
\noindent
where empirical risk Hessian ${\hessFin \defeq \frac{1}{\nTr}\sum_{\zSym \in \trainSet} \gradWSq \risk{\zSym}{\wFin}}$ is by assumption positive definite.
\citet{Koh:2017:Understanding} extend \citeauthor{Cook:1982:InfluenceRegression}'s result to consider the effect of this infinitesimal perturbation on $\zTe$'s~risk, where
\begin{align}
  \frac{d \risk{\zTe}{\wFin}}
       {d \epsI}
  \vertEpsIzero
  &=
  \frac{d \risk{\zTe}{\wFin}}
       {d \wFinEpsI}
  \transpose
  \frac{d \wFinEpsI}
       {d \epsI}
  \vertEpsIzero
  &
  \text{\algcomment{Chain rule}}
  \\
  &=
    -
    \gradW \risk{\zTe}{\wFin}\transpose
    \invHessFin
    \gradW \risk{\zI}{\wFin}
  \text{.}
  \label{eq:Estimators:GradientBased:Static:LossDerivEps}
\end{align}

Removing training instance~$\zI$ from~$\trainSet$ is equivalent to ${\epsI = -\frac{1}{\nTr}}$ making the \kw{pointwise influence functions estimator}
\begin{equation}%
  \labelAndRemember{eq:Estimators:GradientBased:Static:IF}{%
    \baseInfFunc{\infEstIF}{\zI}{\zTe}
    \defeq
      \frac{1}
           {\nTr}
      \gradW \risk{\zTe}{\wFin}\transpose
      \invHessFin
      \gradW \risk{\zI}{\wFin}
  }%
  \text{.}
\end{equation}
\noindent
More intuitively, Eq.~\eqref{eq:Estimators:GradientBased:Static:IF} is the influence functions' estimate of the leave\=/one\=/out influence of $\zI$ on~$\zTe$.

\paragraph{Time, Space, and Storage Complexity}

Calculating inverse Hessian~\eqsmall{$\invHessFin$} directly requires $\bigO{\nTr\dimW^2 + \dimW^3}$~time and $\bigO{\dimW^2}$ space~\citep{Koh:2017:Understanding}.
For large models, this is clearly prohibitive.
Rather than computing~\eqsmall{$\invHessFin$} directly, \citeauthor{Koh:2017:Understanding} instead estimate \textit{Hessian-vector product} (HVP)
\begin{equation}%
  \label{eq:Estimators:GradientBased:Static:IF:STest}
  \infFuncSTest
    \defeq
    \invHessFin
    \gradW \risk{\zTe}{\wFin}
  \text{,}
\end{equation}
\noindent%
with each training instance's pointwise influence then
\begin{equation}%
  \baseInfFunc{\infEstIF}{\zI}{\zTe}
  =
    \frac{1}
         {\nTr}
    \infFuncSTest\transpose
    \gradW \risk{\zI}{\wFin}
  \text{.}
\end{equation}

\citet{Koh:2017:Understanding} use the stochastic algorithm of \citet{Pearlmutter:1994} and \citet{Agarwal:2017} to estimate~$\infFuncSTest$.
This reduces influence functions' complexity to just \bigOnp{} time and \bigOpPln{} space.
Since $\infFuncSTest$ is specific to test instance~$\zTe$, $\infFuncSTest$ must be \textit{estimated for each test instance individually}.
This increases influence functions' computational cost but also means that influence functions require no additional storage.

\paragraph{Strengths and Weaknesses}%
\label{sec:Estimators:GradientBased:Static:IF:StrengthWeaknesses}

Influence functions' clear advantage over retraining-based methods is that influence functions eliminate the need to retrain any models.
Recall from Table~\ref{tab:InfluenceEstimators:Comparison} that
retraining-based methods have a high upfront complexity of retraining but low incremental complexity.
Influence functions invert this computational trade-off.
While computing $\infFuncSTest$ can be slow (several hours for a single test instance~\citep{Yeh:2018:Representer,Guo:2021:FastIF,Hammoudeh:2022:GAS}), it is still significantly faster than retraining \bigOn{} models.
However, influence functions require that $\infFuncSTest$ is recalculated for each test instance.
Hence, when a large number of test instances need to be analyzed, \textit{retraining-based methods may actually be faster than influence functions} through amortization of the retraining cost.

In addition, while influence functions can be very accurate on convex and some shallow models, the assumption that Hessian $\hessFin$ is positive definite often does not hold for deep models~\citep{Basu:2021:InfluenceFunctionsFragile}.
To ensure inverse $\invHessFin$ exists, \citeauthor{Koh:2017:Understanding} add a small dampening coefficient to the matrix's diagonal;
this dampener is a user-specified hyperparameter that is problematic to tune, since there may not be a ground-truth reference.
When this dampening hyperparameter is set too small, $\infFuncSTest$ estimation can diverge~\citep{Yeh:2018:Representer,Hara:2019:SgdInfluence,Hammoudeh:2022:GAS}.

More generally, \citet{Basu:2021:InfluenceFunctionsFragile} show that training hyperparameters also significantly affect influence functions' performance.
Specifically, \citeauthor{Basu:2021:InfluenceFunctionsFragile} empirically demonstrate that model initialization, model width, model depth, weight-decay strength, and even the test instance being analyzed~($\zTe$) all can negatively affect influence functions' \loo{} estimation accuracy.
Their finding is supported by the analysis of \citet{Zhang:2022:RethinkingInfluenceFunctions} who show that HVP estimation's accuracy depends heavily on the model's training regularizer with HVP accuracy ``pretty low under weak regularization.''

\citet{Bae:2022:InfluenceFunctionsQuestion} also empirically analyze the potential sources of influence functions' fragility on deep models.
\citeauthor{Bae:2022:InfluenceFunctionsQuestion} identify five common error sources, the first three of which are the most important.%
\footnote{%
  \citet[Table~1]{Bae:2022:InfluenceFunctionsQuestion} provide a formal mathematical definition of influence functions' five error sources.%
}
\begin{itemize}
  \item \textit{Warm-start gap}:
    Influence functions more closely resembles performing fine-tuning close to final model parameters~$\wFin$ than retraining from scratch (i.e.,~starting from initial parameters~$\wZero$).
    This difference in starting conditions can have a significant effect on the \loo{} estimate.
  \item \textit{Proximity gap}:
    The error introduced by the dampening term included in the HVP~($\infFuncSTest$) estimation algorithm.
  \item \textit{Non-convergence gap}:
    The error due to final model parameters~$\wFin$ not being a stationary point, i.e.,~${\sum_{\zI \in \trainSet} \gradW \risk{\zI}{\wFin} \ne \zeroVec}$.
  \item \textit{Linearization error}:
    The error induced by considering only a first-order Taylor approximation when deleting~$\zI$ and ignoring the potential effects of curvature on~$\baseInfFunc{\infEstIF}{\zI}{\zTe}$~\citep{Basu:2020:OnSecond}.
  \item \textit{Solver error}:
    General error introduced by the specific solver used to estimate~$\infFuncSTest$.
\end{itemize}
\noindent
Rather than estimating the \loo{} influence,
\citeauthor{Bae:2022:InfluenceFunctionsQuestion} argue that influence functions more closely estimate a different measure they term the \kw{proximal Bregman response function}~(PBRF).
\citeauthor{Bae:2022:InfluenceFunctionsQuestion} provide the intuition that PBRF ``approximates the effect of removing a data point while trying to keep predictions consistent with the \ldots{} trained model''~\citep{Bae:2022:InfluenceFunctionsQuestion}.
Put simply, PBRF mimics a prediction-constrained LOO influence.

\citeauthor{Bae:2022:InfluenceFunctionsQuestion} assert that PBRF can be applied in many of the same situations where \loo{} is useful.
\citeauthor{Bae:2022:InfluenceFunctionsQuestion} further argue that influence functions' fragility reported by earlier works~\citep{Basu:2021:InfluenceFunctionsFragile,Zhang:2022:RethinkingInfluenceFunctions} is primarily due to those works focusing on the ``wrong question'' of \loo{}.
When the ``right question'' is posed and influence functions are evaluated w.r.t.\ PBRF, influence functions give accurate answers.

\paragraph{Related Methods}%
\label{sec:Estimators:GradientBased:Static:IF:RelatedMethods}

Improving influence functions' computational scalability has been a primary focus of follow-on work.
For instance, applying influence functions only to the model's final linear (classification) layer has been considered with at best mixed results~\citep{Barshan:2020:RelatIF,Yeh:2022:FirstBetterLast}.%
\footnote{%
  Section~\ref{sec:Estimators:GradientBased:Static:RepresenterPoint:StrengthWeaknesses} discusses some of the pitfalls associated with estimating influence using only a model's final (linear) layer.%
}
Moreover, \citepos{Guo:2021:FastIF} \kw{fast influence functions} (\fastif) integrate multiple speed-up heuristics.
First, they leverage the inherent parallelizability of \citepos{Pearlmutter:1994} HVP estimation algorithm.
In addition, \fastif{} includes recommended hyperparameters for \citeauthor{Pearlmutter:1994}'s HVP algorithm that reduces its execution time by 50\%~on average.

\relatedHeading{Arnoldi-Based Influence Functions}
More recently, \citet{Schioppa:2022:ScaledUpInfluenceFunctions} show that influence functions can be sped-up by three to four orders of magnitude by using a different algorithm to estimate~$\infFuncSTest$.
Specifically, \citeauthor{Schioppa:2022:ScaledUpInfluenceFunctions} use \citepos{Arnoldi:1951} famous algorithm that quickly finds the dominant (in absolute terms) eigenvalues and eigenvectors of \eqsmall{$\hessFin$}.
These dominant eigenvectors serve as the basis when projecting all gradient vectors to a lower-dimensional subspace.
\citeauthor{Schioppa:2022:ScaledUpInfluenceFunctions} evaluate their revised influence functions estimator on large transformer networks (e.g.,~ViT\=/L32 with 300M parameters~\citep{Dosovitskiy:2021:VisionTransformer}), which are orders of magnitude larger than the simple networks \citet{Koh:2017:Understanding} consider~\citep{Schioppa:2023:TheoreticalPracticalInfFunc}.

\relatedHeading{Influence Functions for Decision Trees}
Another approach to speed up influence functions is to specialize the estimator to model architectures with favorable computational properties.
For example, \citepos{Sharchilev:2018:TreeInfluence} \leafInfluence{} method adapts influence functions to gradient boosted decision tree ensembles.
By assuming a fixed tree structure and then focusing only on the trees' leaves, \leafInfluence{}'s tree-based estimates are significantly faster than influence functions on deep models~\citep{Brophy:2023:TreeInfluence}.

\relatedHeading{Group Influence Functions}
A major strength of influence functions is that it is one of the few influence analysis methods that has been studied beyond the pointwise domain.
For example, %
\citepos{Koh:2019:GroupEffects} follow-on paper analyzes influence functions' empirical performance estimating group influence.
In particular, \citet{Koh:2019:GroupEffects} consider coherent training data subpopulations whose removal is expected to have a large, broad effect on the model.
Even under naive assumptions of (pointwise) influence additivity, \citet{Koh:2019:GroupEffects} observe that simply summing influence functions estimates tends to \textit{underestimate} the true group influence.
More formally, let ${\trainSetAlt \subseteq \trainSet}$ be a coherent training-set subpopulation, then
\begin{equation}
  \sum_{\zI \in \trainSetAlt}
    {\big\lvert
      \baseInfFunc{\infEstIF}%
                  {\zI}%
                  {\zTe}%
    \big\rvert}
  <
    {\big\lvert
      \baseInfFunc{\influence}%
                  {\trainSetAlt}%
                  {\zTe}%
    \big\rvert}
  \text{.}
\end{equation}
\noindent
Nonetheless, influence functions' additive group estimates tend to have strong rank correlation w.r.t.\ subpopulations' true group influence.
In addition, \citet{Basu:2020:OnSecond} extend influence functions to directly account for subpopulation group effects by considering higher-order %
terms in influence functions' Taylor-series approximation.

\subsubsection{Representer Point Methods}%
\label{sec:Estimators:GradientBased:Static:RepresenterPoint}

Unlike this section's other gradient-based estimators, \citepos{Yeh:2018:Representer} \kw{representer point} method does not directly rely on a Taylor-based expansion.
Instead, \citeauthor{Yeh:2018:Representer} build on \citepos{Scholkopf:2001:RepresenterTheorem} \kw{generalized representer theorem}.
The derivation below assumes that model~$\dec$ is linear.
\citeauthor{Yeh:2018:Representer} use a simple ``trick'' to extend linear representer point methods to multilayer models like neural networks.

Representer-based methods rely on \kw{kernels}, which are functions $\func{\kernel}{\domainX \times \domainX}{\real}$ that measure the similarity between two vectors~\citep{Hofmann:2008:KernelMethods}.
\citeauthor{Scholkopf:2001:RepresenterTheorem}'s representer theorem proves that the optimal solution to a class of \lTwo{}~regularized functions can be reformulated as a weighted sum of the training data in kernel form.
Put simply, representer methods decompose the predictions of specific model classes into the individual contributions (i.e.,~influence) of each training instance.
This makes influence estimation a natural application of the representer theorem.

Consider regularized empirical risk minimization where optimal parameters satisfy%
\footnote{%
  Eq.~\eqref{eq:Estimators:GradientBased:Static:RepPt:ERM} is slightly different than \citepos{Yeh:2018:Representer} presentation.
  This alternate formulation requires specifying $\wdecay$ $2\times$~larger.
  We selected this alternate presentation to provide consistency with later work~\citep{Tsai:2023:HighDimRepPt,Chen:2021:Hydra}.
}
\begin{equation}%
  \label{eq:Estimators:GradientBased:Static:RepPt:ERM}
  \wOpt
    \defeq
      \argmin_{\W}
      \frac{1}{\nTr}
      \sum_{\zI \in \trainSet}
        \risk{\zI}%
             {\W}
        +
        \frac{\wdecay}%
             {2}
        \norm{\W}_{2}
  \text{,}
\end{equation}
\noindent
with ${\wdecay > 0}$ the \lTwo{}~regularization strength.
Note that Eq.~\eqref{eq:Estimators:GradientBased:Static:RepPt:ERM} defines minimizer~$\wOpt$ slightly differently than the last section~\eqref{eq:Estimators:GradientBased:Static:IF:ErmBase} since the representer theorem \textit{requires} regularization.

Empirical risk minimizers are \kw{stationary points} meaning
\begin{equation}
  \gradW
  \bigg(
    \frac{1}{\nTr}
    \sum_{\zI \in \trainSet}
      \risk{\zI}{\wOpt}
      +
      \wdecay \norm{\wOpt}_{2}
  \bigg)
  =
  \zeroVec
  \text{,}
\end{equation}
\noindent
where $\zeroVec$~is the $\dimW$\=/dimensional, zero vector.  The above simplifies to
\begin{align}
  \frac{1}%
       {\nTr}
  \sum_{\zI \in \trainSet}
    \frac{\partial \risk{\zI}{\wOpt}}%
         {\partial \W}
    + \wdecay \wOpt
  &=
    \zeroVec
  \\
  \wOpt
  &=
    -
    \frac{1}%
         {\wdecay \nTr}
    \sum_{\zI \in \trainSet}
      \frac{\partial \risk{\zI}{\wOpt}}%
           {\partial \W}
  \text{.}
  \label{eq:Estimators:GradientBased:Static:RepresenterPoint:WoptOne}
\end{align}

For a linear model where ${\decFunc{\X}{\W} = \W\transpose \X \fedeq \yHat}$, Eq.~\eqref{eq:Estimators:GradientBased:Static:RepresenterPoint:WoptOne} further simplifies via the chain rule to
{\small%
  \begin{align}
    \wOpt &=
      -\frac{1}{\wdecay \nTr}
      \sum_{(\xI,\yI) \in \trainSet}
      \frac{\partial \lDecFunc{\xI}{\wOpt}{\yI}}{\partial \W}
    \\
    &=
      -\frac{1}{\wdecay \nTr}
      \sum_{(\xI,\yI) \in \trainSet}
      \rpLossDerivI
      \,
      \frac{\partial \yHatI}{\partial \W}
    &
    \text{\algcomment{Chain Rule}}
    \\
    &=
      -\frac{1}{\wdecay \nTr}
      \sum_{(\xI,\yI) \in \trainSet}
      \rpLossDerivI
      \,
      \xI
    &
    \text{\algcomment{$\yHat = \W\transpose \X$}}
    \\
    &=
      \sum_{(\xI,\yI) \in \trainSet}
      \rpValI
      \xI
    \text{,}
  \end{align}%
}%
\noindent
where $\loss$~is any once-differentiable loss function, \eqsmall{$\rpLossDerivI$} is the gradient of \textit{just the loss function} itself w.r.t.\ model output~$\yHat$,%
\footnote{%
  In the case of classification, \eqsmall{$\rpLossDerivI$} has $\abs{\domainY}$\=/dimensions, i.e.,~its dimension equals the number of classes.%
}
 and \eqsmall{${\rpValI \defeq -\frac{1}{\wdecay\nTr} \rpLossDerivI}$} is the $\trIdx$\=/th training instance's \kw{representer value}.
\citet{Yeh:2018:Representer} provide the intuition that a larger magnitude~$\rpValI$ indicates that training instance~$\zI$ has larger influence on the final model parameters~$\wOpt$.

Following \citepos{Scholkopf:2001:RepresenterTheorem} representer theorem kernelized notation, the training set's group influence on test instance~$\zTe$ for any linear model is
\begin{equation}
  \baseInfFunc{\infRepPt}{\trainSet}{\zTe}
  =
    \sum_{\trIdx = 1}^{\nTr}
      \rpValI \xI\transpose \xTe \teDimVert
  =
    \sum_{\trIdx = 1}^{\nTr}
      \kernelITe
  =
    \sum_{\trIdx = 1}^{\nTr}
    \baseInfFunc{\infRepPt}{\zI}{\zTe}
  \text{,}
\end{equation}
\noindent
where kernel function~${\kernelFunc{\xI}{\xTe}{\rpValI} \defeq \rpValI \xI\transpose \xTe}$ returns a vector.
${\kernelITe}$~denotes the kernel value's $\yTe$\=/th dimension.
Then, $\zI$'s \kw{pointwise linear representer point influence} on~$\zTe$ is
\begin{equation}
  \labelAndRemember{eq:Estimators:GradientBased:Static:RepresenterPoint:LinearIdeal}{%
    \baseInfFunc{\infRepPt}{\zI}{\zTe}
    =
      \rpValI \xI\transpose \xTe \teDimVert
  }
  =
    \kernelITe
  \text{.}
\end{equation}

\noindent
\textbf{Extending Representer Point to Multilayer Models}
Often, linear models are insufficiently expressive, with multilayer models used instead.
In such cases, the representer theorem above does not directly apply.
To workaround this limitation, \citet{Yeh:2018:Representer} rely on what they (later) term \kw{last layer similarity}~\citep{Yeh:2022:FirstBetterLast}.

Formally, \citet{Yeh:2018:Representer} partition the model parameters~%
\eqsmall{${\wFin = \lbrack~\wFinBegin~~\wFinLast~\rbrack}$}
into two subsets, where $\wFinLast$~is the last linear (i.e.,~classification) layer's parameters and \eqsmall{${\wFinBegin \defeq \wFin \setminus \wFinLast}$}~is all other model parameters.
Since $\wFinLast$~is simply a linear function, the representer theorem analysis above still applies to it.
\citet{Yeh:2018:Representer} treat the other parameters,~$\wFinBegin$, as a fixed feature extractor and ignore them in their influence analysis.

To use \citepos{Yeh:2018:Representer} multilayer trick,
one small change to Eq.~\eqref{eq:Estimators:GradientBased:Static:RepresenterPoint:LinearIdeal} is required.
In multilayer models, the final (linear) layer does not operate over feature vectors $\xI$ and $\xTe$ directly.
Instead, the final layer only sees an intermediate feature representation.
For arbitrary feature vector~$\X$, let $\featSym$~be the feature representation generated by model parameters~$\wFinBegin$, i.e.,~vector $\featSym$ is the input to the model's last linear layer given input~$\X$.
Then the \kw{representer point influence estimator} for a multilayer model is
\begin{equation}
  \labelAndRemember{eq:Estimators:GradientBased:Static:RepresenterPoint:Estimate}{%
    \baseInfFunc{\infEstRepPt}{\zI}{\zTe}
    \defeq
      \rpValI
      \featI\transpose \featTe \teDimVert
  }
  =
  \kernelFunc{\featI}{\featTe}{\rpValI}\teDimVert \text{.}
\end{equation}

\paragraph{Time, Space, and Storage Complexity}

Treating as constants feature representation dimension~$\abs{\featSym}$ and the overhead to calculate \eqsmall{$\rpLossDerivI$}, estimating the entire training set's representer point influence only requires calculating $\nTr$~dot products. %
This only takes \bigOn{} time and \bigOpPln{} space with no additional storage requirements.

\paragraph{Strengths and Weaknesses}
\label{sec:Estimators:GradientBased:Static:RepresenterPoint:StrengthWeaknesses}

Representer point's primary advantage is its theoretical and computational simplicity.
Eq.~\eqref{eq:Estimators:GradientBased:Static:RepresenterPoint:Estimate} only considers the training and test instances' final feature representations and loss function gradient~\eqsmall{$\rpLossDerivI$}.
Hence, the majority of representer point's computation is forward-pass only and can be sped up using batching.
This translates to representer point being very fast -- several orders of magnitude faster than influence functions and Section~\ref{sec:Estimators:GradientBased:Dynamic}'s dynamic estimators~\citep{Hammoudeh:2021:Simple}.

However, representer points' simplicity comes at a cost.
First, at the end of training, it is uncommon that a model's final linear layer has converged to a stationary point.
Before applying their method, \citet{Yeh:2018:Representer} recommend freezing all model layers except the final one (i.e.,~freezing $\wFinBegin$) and then fine-tuning the classification layer~($\wFinLast$) until convergence/stationarity.
Without this extra fine-tuning, representer point's stationarity assumption does not hold, and poor influence estimation accuracy is expected.
Beyond just complicating the training procedure itself, this extra training procedure also complicates comparison with other influence methods since it may require evaluating the approaches on different parameters.

Moreover, by focusing exclusively on the model's final linear (classification) layer, representer point methods may miss influential behavior that is clearly visible in other layers.
For example, \citet{Hammoudeh:2022:GAS} demonstrate that while some training-set attacks are clearly visible in a network's final layer, other attacks are only visible in a model's first layer -- despite both attacks targeting the same model architecture and dataset.
In their later paper, %
\citet{Yeh:2022:FirstBetterLast} acknowledge the disadvantages of considering only the last layer writing, ``that choice critically affects the similarity component of data influence and leads to inferior results.''
\citet{Yeh:2022:FirstBetterLast} further state that the feature representations in the final layer -- and by extension representer point's influence estimates -- can be ``too reductive.''

In short, \citepos{Yeh:2018:Representer} representer point method is highly scalable and efficient but is only suitable to detect behaviors that are obvious in the model's final linear layer.

\paragraph{Related Methods}\hfill~\linebreak
Given the accuracy limitations of relying on last-layer similarity, limited follow-on work has adapted representer-point methods.

\relatedHeading{Adapting Representer Point to Decision Trees}
\citet{Brophy:2023:TreeInfluence} extend representer point methods to decision forests via their Tree-ensemble Representer Point Examples~(TREX) estimator.  Specifically, they use \kw{supervised tree kernels} -- which provide an encoding of a tree's learned representation structure~\citep{Davies:2014:RandomForestKernel,He:2014:PredictingClicks} -- for similarity comparison.

\relatedHeading{Making Representer Point More Robust}
In addition, \citet{Sui:2021:LocalJacobianRepPt} propose \kw{representer point selection based on a local Jacobian expansion} (\rpLocalJacobian), which can be viewed as a generalizing \citepos{Yeh:2018:Representer} base method.
Rather than relying on \citepos{Scholkopf:2001:RepresenterTheorem} representer theorem, \citeauthor{Sui:2021:LocalJacobianRepPt}'s formulation relies on a first-order Taylor expansion that estimates the difference between the final model parameters and a true stationary point.
\rpLocalJacobian{} still follows vanilla representer point's kernelized decomposition where ${
    \baseInfFunc{\infEstRepPt}{\zI}{\zTe}
    \defeq
      \rpValI
      \featI\transpose \featTe
}$, albeit with $\rpValI$ defined differently.

\rpLocalJacobian{} addresses two weaknesses of \citepos{Yeh:2018:Representer} base approach.
First, \citeauthor{Sui:2021:LocalJacobianRepPt}'s formulation does not presume that $\wFin$ is a stationary point.
Therefore, \rpLocalJacobian{} does not require post-training fine-tuning to enforce stationarity.
Second, as Section~\ref{sec:Estimators:GradientBased:TradeOff} discusses in detail, gradient-based estimators, including vanilla representer point, tend to mark as most influential those training instances with the largest loss values.
This leads to all test instances from a given class having near identical top\=/k influence rankings.
\rpLocalJacobian{}'s alternate definition of $\rpValI$ is less influenced by a training instance's loss value, which enables \rpLocalJacobian{} to generate more semantically meaningful influence rankings.

\newcommand{\subspaceA}{\mathcal{U}}
\newcommand{\subspaceAVec}{u}
\newcommand{\subspaceB}{\mathcal{V}}
\newcommand{\subspaceBVec}{v}

\revOne{%
\relatedHeading{Extending Representer Point to Other Regularizers}
\citepos{Yeh:2018:Representer} representer point formulation exclusively considers \lTwo{}\=/regularized models.
Intuitively, regularization's role is to encourage the model parameters to meet certain desired properties, which may necessitate the use of alternate regularizers.
For example, \lOne{}~regularization is often used to induce sparse minimizers.

Recently,
\citet{Tsai:2023:HighDimRepPt} propose \kw{high-dimensional representers}, a novel extension of \citepos{Yeh:2018:Representer} representer theorem to additional types of regularization.
Specifically, \citeauthor{Tsai:2023:HighDimRepPt} consider \kw{decomposable} regularization functions~\citep{Negahban:2012:DecomposableRegularizers}.
Formally, a regularization function $\func{\regularizerSym}{\domainW}{\realnn}$ is decomposable w.r.t.\ two subspaces ${\subspaceA, \subspaceB \subseteq \domainW}$ if ${\forall\, \subspaceAVec \in \subspaceA}$ and ${\forall\, \subspaceBVec \in \subspaceB}$,
\begin{equation}%
  \label{eq:Estimators:GradientBased:Static:RepPt:Decomposable}%
  \regFunc{\subspaceAVec + \subspaceBVec}
  =
  \regFunc{\subspaceAVec}
  +
  \regFunc{\subspaceBVec}
  \text{.}
\end{equation}
Examples of decomposable regularizers include \lOne{}\=/norm \citep{Tibshirani:1996:LassoRegression} and the matrix nuclear norm~\citep{Yuan:2007:MultivariateRegression,Recht:2011:SimpleMatrixCompletion,Yang:2017:NuclearNormMatrixRegression}.

High-dimensional representers follow Eq.~\eqref{eq:Estimators:GradientBased:Static:RepresenterPoint:Estimate}'s kernelized form.
Representer value~${\rpValI \defeq -\frac{1}{\wdecay\nTr} \rpLossDerivI}$ still quantifies the global importance of each training instance.
Moreover, the similarity between training instance~$\xI$ and test instance~$\xTe$ is still measured via a kernel function~($\kernel$).
The only difference is that the kernels are specialized for these alternate regularizers; specifically the kernels are based on the decomposable regularization function's \kw{sub\=/differential}~\citep{Negahban:2012:DecomposableRegularizers}.%
}

\subsection{Dynamic, Gradient-Based Influence Estimation}%
\label{sec:Estimators:GradientBased:Dynamic}

All preceding influence methods -- static, gradient-based and retraining-based -- define and estimate influence using only final model parameters,~$\wFinSubBase{\trainSetAlt}$, where ${\trainSetAlt \subseteq \trainSet}$.
These final parameters only provide a snapshot into a training instance's possible effect.
Since neural network training is NP-complete~\citep{Blum:1992:TrainingNpComplete}, it can be provably difficult to reconstruct how each training instance affected the training process.

As an intuition, an influence estimator that only considers the final model parameters is akin to only reading the ending of a book.
One \textit{might} be able to draw some big-picture insights, but the finer details of the story are most likely lost.
Applying a dynamic influence estimator is like reading a book from beginning to end.
By comprehending the whole influence ``story,'' dynamic methods can observe training data relationships -- both fine-grained and general -- that other estimators miss.

Since test instance~$\zTe$ may not be known before model training, in-situ influence analysis may not be possible.
Instead, as shown in Alg.~\ref{alg:GradientBased:Dynamic:Training}, intermediate model parameters ${\trainParams \subseteq \myset{\wZero, \ldots, \wT[\nItr - 1]}}$ are stored during training for post hoc influence analysis.%
\footnote{%
  In practice, only a subset of~$\myset{\wZero, \ldots, \wT[\nItr - 1]}$ is actually stored.
  Heuristics are then applied to this subset to achieve acceptable influence estimation error~\citep{Pruthi:2020:TracIn,Hammoudeh:2022:GAS}.%
}

Below we examine two divergent approaches to dynamic influence estimation -- the first defines a novel definition of influence while the second estimates leave\=/one\=/out influence with fewer assumptions than influence functions.

\begin{figure}[t]
  \centering
  \begin{minipage}{\algMiniPageWidth}
    \begin{algorithm}[H]
      {%
        \AlgFontSize%
        \caption{Dynamic influence estimation's training phase}\label{alg:GradientBased:Dynamic:Training}
\begin{algorithmic}[1]
  \algSetStretch%
  \Require Training set~$\trainSet$\algInputDelim{}
           iteration count~$\nItr$\algInputDelim{}
           learning rates ${\lrOne, \ldots, \lrFin}$\algInputDelim{}
           batch sizes~${\batchSizeOne, \ldots, \batchSizeFin}$\algInputDelim{}
           and
           initial parameters~$\wZero$

  \Ensure Final parameters~$\wFin$ and stored parameter set~$\trainParams$

  \State $\trainParams \gets \emptyset$
  \For{$\itr \gets 1 \textbf{ to } \nItr$}
    \State $\trainParams \gets \trainParams \cup \myset{\wTOne}$ \algcomment{Store intermediate params.}
    \State $\batchT \,\simN{\batchSizeTSym}\, \trainSet$
    \State $\wT \gets \textsc{Update}(\lrT, \wTOne, \batchT)$
  \EndFor
  \State \Return $\wFin$, $\trainParams$
\end{algorithmic}
       }%
    \end{algorithm}
  \end{minipage}
\end{figure}

\subsubsection{\tracin{} -- Tracing Gradient Descent}%
\label{sec:Estimators:GradientBased:Dynamic:TracIn}

  Fundamentally, all preceding methods define influence w.r.t.\ changes to the training set.
  \citet{Pruthi:2020:TracIn} take an orthogonal perspective.
  They treat training set~$\trainSet$ as fixed, and consider the \textit{change in model parameters as a function of time}, or more precisely, the training iterations.

  Vacuously, the training set's group influence on test instance~$\zTe$ is
\begin{equation}\label{eq:Estimators:GradientBased:Dynamic:TracIn:DatasetGroup}
  \infFunc{\trainSet}%
          {\zTe}
  =
    \risk{\zTe}{\wZero}
    -
    \risk{\zTe}{\wFin}
  \text{.}
\end{equation}
\noindent
In words, training set~$\trainSet$ causes the entire change in test loss between random initial parameters~$\wZero$ and final parameters~$\wFin$.
Eq.~\eqref{eq:Estimators:GradientBased:Dynamic:TracIn:DatasetGroup} decomposes by training iteration~$\itr$ as
\begin{equation}\label{eq:Estimators:GradientBased:Dynamic:Tracin:DatasetGroup}
  \infFunc{\trainSet}{\zTe}
  =
    \sum_{\itr = 1}^{\nItr}
      \left(
        \risk{\zTe}{\wTOne}
        -
        \risk{\zTe}{\wT}
      \right)
  \text{.}
\end{equation}

Consider training a model with vanilla stochastic gradient descent, where each training minibatch~$\batchT$ is a single instance and gradient updates have no momentum~\citep{Rumelhart:1986:BackProp}.
Here, each iteration~$\itr$ has no effect on any other iteration beyond the model parameters themselves.
Combining this with singleton batches enables attribution of each parameter change to a single training instance, namely whichever instance was in~$\batchT$.
Under this regime, \citet{Pruthi:2020:TracIn} define the \kw{ideal \tracin{} pointwise influence} as
\begin{equation}
  \labelAndRemember{eq:Estimators:GradientBased:Dynamic:TracInIdeal}{%
    \baseInfFunc{\infTracInIdeal}{\zI}{\zTe}
        \defeq
        \sum_{\substack{\itr \\ \zI = \batchT}}
          \left(
            \risk{\zTe}{\wTOne}
            -
            \risk{\zTe}{\wT}
          \right)
  }
  \text{,}
\end{equation}
where the name ``\tracin{}'' derives from ``tracing gradient descent influence.''
Eq.~\eqref{eq:Estimators:GradientBased:Dynamic:Tracin:DatasetGroup} under vanilla stochastic gradient descent decomposes into the sum of all pointwise influences
\begin{equation}\label{eq:Estimators:GradientBased:Dynamic:Tracin:DatasetGroup:IndividualInstance}
  \infFunc{\trainSet}{\zTe}
  =
    \sum_{\trIdx = 1}^{\nTr}
      \Bigg(
        \sum_{\substack{\itr \\ \zI = \batchT}}
          \risk{\zTe}{\wTOne}
          -
          \risk{\zTe}{\wT}
      \Bigg)
  =
    \sum_{\trIdx = 1}^{\nTr}
      \baseInfFunc{\infTracInIdeal}{\zI}{\zTe}
  \text{.}
\end{equation}

While the ideal \tracin{} influence has a strong theoretical motivation, its assumption of singleton batches and vanilla stochastic gradient descent is unrealistic in practice.
To achieve reasonable training times, modern models train on batches of up to hundreds of thousands or millions of instances.
Training on a single instance at a time would be far too slow~\citep{You:2017:LargeBatchTraining,Goyal:2017:LargeMinibatchSGD,Brown:2020:GPT3}.

A naive fix to Eq.~\eqref{eq:Estimators:GradientBased:Dynamic:TracInIdeal} to support non-singleton batches assigns the same influence to all instances in the minibatch, or more formally, divide the change in loss ${\risk{\zTe}{\wTOne} - \risk{\zTe}{\wT}}$ by batch size~$\abs{\batchT}$ for each ${\zI \in \batchT}$.
This naive approach does not differentiate those instances in batch~$\batchT$ that had positive influence on the prediction from those that made the prediction worse.

Instead, \citet{Pruthi:2020:TracIn} estimate the contribution of each training instance within a minibatch via a first-order Taylor approximation. Formally,
\begin{equation}\label{eq:Estimators:GradientBased:Dynamic:TracIn:TaylorExpansion}
  \risk{\zTe}{\wT}
  \approx
  \risk{\zTe}{\wTOne}
  +
  \dotprod{\gradTe}
          {(\wT - \wTOne)}
  \text{.}
\end{equation}
\noindent
Under gradient descent without momentum, the change in model parameters is directly determined by the batch instances' gradients,~i.e.,
\begin{equation}\label{eq:Estimators:GradientBased:Dynamic:TracIn:ParamChangeInstance}
  \wT - \wTOne
  =
    -
    \frac{\lrT}{\abs{\batchT}}
    \sum_{\zI \in \batchT}
      \gradI
  \text{,}
\end{equation}
\noindent
where $\lrT$ is iteration~$\itr$'s learning rate.

Combining Eqs.~\eqref{eq:Estimators:GradientBased:Dynamic:Tracin:DatasetGroup:IndividualInstance} to~\eqref{eq:Estimators:GradientBased:Dynamic:TracIn:ParamChangeInstance}, the \kw{\tracin{} pointwise influence estimator} is
\begin{equation}%
  \labelAndRemember{eq:Estimators:GradientBased:TracIn}{%
    \baseInfFunc{\infEstTracIn}{\zI}{\zTe}
      \defeq
      \sum_{\substack{\itr \\ \zI \in \batchT}}
        \frac{\lrT}
             {\batchSizeT}
        \,
        \dotprod{\gradI}
                {\gradTe}
  }
  \text{,}
\end{equation}
\noindent
with the complete \tracin{} influence estimation procedure shown in Alg.~\ref{alg:GradientBased:Dynamic:TracIn}.

\noindent
\textbf{A More ``Practical'' \tracin{}}
Training's stochasticity can negatively affect the performance of both ideal \tracin{}~\eqref{eq:Estimators:GradientBased:Dynamic:TracInIdeal} and the \tracin{} influence estimator~\eqref{eq:Estimators:GradientBased:TracIn}.
As an intuition, consider when the training set contains two identical copies of some instance.
All preceding gradient-based methods assign those two identical instances the same influence score.
However, it is unlikely that those two training instances will always appear together in the same minibatch.
Therefore, ideal \tracin{} almost certainly assigns these identical training instances different influence scores.
These assigned scores may even be vastly different -- by up to several orders of magnitude~\citep{Hammoudeh:2022:GAS}.
This is despite identical training instances always having the same \textit{expected} \tracin{} influence.

\citet{Pruthi:2020:TracIn} recognize randomness's effect on \tracin{} and propose the \kw{\tracin{} Checkpoint influence estimator}~(\tracinCP{}) as a ``practical'' alternative.
Rather than retrace all of gradient descent, \tracinCP{} considers only a subset of the training iterations (i.e.,~checkpoints) ${\subsetItr \subseteq \setint{\nItr}}$.
More importantly, at each ${\itr \in \subsetItr}$, \textit{all} training instances are analyzed -- not just those in recent batches.
Eq.~\eqref{eq:Estimators:GradientBased:TracInCP} formalizes \tracinCP{}, with its modified influence estimation procedure shown in Alg.~\ref{alg:GradientBased:Dynamic:TracInCP}.
\begin{equation}%
  \labelAndRemember{eq:Estimators:GradientBased:TracInCP}{%
    \baseInfFunc{\infEstTracInCP}{\zI}{\zTe}
      \defeq
      \sum_{\itr \in \subsetItr}
        \lrT \, \dotprod{\gradI}{\gradTe}
  }
\end{equation}

Observe that, unlike \tracin{}, \tracinCP{} assigns identical training instances the same influence estimate.
Therefore, \tracinCP{} more closely estimates expected influence than \tracin{}.
\citet{Pruthi:2020:TracIn} use \tracinCP{} over \tracin{} in much of their empirical evaluation.
Other work has also shown that \tracinCP{} routinely outperforms \tracin{} on many tasks~\citep{Hammoudeh:2022:GAS}.

\begin{figure}[t]
  \centering
  \begin{minipage}[t]{0.49\textwidth}
    \begin{algorithm}[H]
      {%
        \AlgFontSize%
        \caption{\tracin{} influence estimation}\label{alg:GradientBased:Dynamic:TracIn}
\begin{algorithmic}[1]
  \algSetStretch%
  \Require Training param.\ set~$\trainParams$\algInputDelim{}
           iteration count~$\nItr$\algInputDelim{}
           batches ${\batchOne, \ldots, \batchFin}$\algInputDelim{}
           learning rates ${\lrOne, \ldots, \lrFin}$\algInputDelim{}
           training instance~$\zI$\algInputDelim{}
           and
           test example~$\zTe$

  \Ensure \tracin{} influence estimate~$\baseInfFunc{\infEstTracIn}{\zI}{\zTe}$

    \State $\infEstSym \gets 0$
    \For{$\itr \gets 1 \textbf{ to } \nItr$}
      \If{${\zI \in \batchT}$}
        \State ${\wTOne \gets \trainParams\sbrack{\itr}}$
        \State {%
          \scriptsize%
          $\infEstSym \gets \infEstSym %
                            + \frac{\lrT}{\abs{\batchT}} %
                              \dotprod{\gradI}{\gradTe}$%
        }%
      \EndIf%
    \EndFor%
    \State \Return $\infEstSym$%
\end{algorithmic}
       }%
    \end{algorithm}
  \end{minipage}
  \hfill
  \begin{minipage}[t]{0.455\textwidth}
    \begin{algorithm}[H]
      {%
        \AlgFontSize%
        \caption{\tracinCP{} influence estimation}\label{alg:GradientBased:Dynamic:TracInCP}
\begin{algorithmic}[1]
  \algSetStretch%
  \Require Training param.\ set~$\trainParams$\algInputDelim{}
           iteration subset~$\subsetItr$\algInputDelim{}
           learning rates ${\lrOne, \ldots, \lrFin}$\algInputDelim{}
           training instance~$\zI$\algInputDelim{}
           and
           test example~$\zTe$

  \Ensure \tracinCP{} influence est.~$\baseInfFunc{\infEstTracInCP}{\zI}{\zTe}$

    \State $\infEstSym \gets 0$
    \For{\textbf{each} $\itr \in \subsetItr$}
      \State ${\wTOne \gets \trainParams\sbrack{\itr}}$
      \State {%
        \scriptsize%
        $\infEstSym \gets \infEstSym %
                          + \lrT
                            \dotprod{\gradI}{\gradTe}$%
      }
    \EndFor
    \State \Return $\infEstSym$
\end{algorithmic}
       }%
    \end{algorithm}
  \end{minipage}
\end{figure}

\begin{remark}
  \citeauthor{Pruthi:2020:TracIn}'s empirical evaluation uses ${\abs{\subsetItr} \ll \nItr}$.
  For example, when identifying mislabeled examples using \tracinCP{}, \citeauthor{Pruthi:2020:TracIn} evaluate every 30th~iteration.
  Furthermore, \citeauthor{Pruthi:2020:TracIn} note that prioritizing the small number of checkpoints where $\zTe$'s loss changes significantly generally outperforms evaluating a larger number of evenly-spaced checkpoints.
\end{remark}

\paragraph{Time, Space, and Storage Complexity}
Below we derive the time complexity of both versions of \tracin{}.
We then discuss their space and storage complexities.

Consider first \tracinCP{}'s time complexity since it is simpler to derive.
From Alg.~\ref{alg:GradientBased:Dynamic:TracInCP}, each checkpoint in ${\subsetItr}$ requires $\nTr$, $\dimW$\=/dimensional dot products making \tracinCP{}'s complexity $\bigO{\nTr \dimW \abs{\subsetItr}}$.
For vanilla \tracin{}, consider Alg.~\ref{alg:GradientBased:Dynamic:TracIn}.
For each iteration ${\itr \in \setint{\nItr}}$, a $\dimW$\=/dimensional dot product is performed for each instance in $\batchT$.
Let ${b \defeq \max_{\itr} \abs{\batchT}}$ denote the maximum batch size, then \tracin{}'s time complexity is $\bigO{b \dimW \nItr}$.
In the worst case where ${\forall_{\itr} \, \batchT = \trainSet}$ (full-batch gradient descent), \tracin{} time complexity is \bigOnpT{}.

Recall that, by definition, ${\abs{\subsetItr} \leq \nItr}$ meaning \tracinCP{} is \textit{asymptotically} faster than \tracin{}.
However, this is misleading.
In practice, \tracinCP{} is generally \textit{slower} than \tracin{} as \citeauthor{Pruthi:2020:TracIn} note.

Since each gradient calculation is independent, \tracin{} and \tracinCP{} are fully parallelizable.
Table~\ref{tab:InfluenceEstimators:Comparison} treats the level of concurrency as a constant factor, making the space complexity of both \tracin{} and \tracinCP{} \bigOpPln{}.

Lastly, as detailed in Alg.~\ref{alg:GradientBased:Dynamic:Training}, dynamic influence estimators require that intermediate model parameters~$\trainParams$ be saved during training for post hoc influence estimation.
In the worst case, each training iteration's parameters are stored resulting in a storage complexity of~\bigOpT{}.
In practice however, \tracin{} only considers a small fraction of these $\nItr$~training parameter vectors, meaning \tracin{}'s actual storage complexity is generally (much) lower than the worst case.

\paragraph{Strengths and Weaknesses}%
\label{sec:Estimators:GradientBased:Dynamic:TracIn:StrengthsWeaknesses}

\tracin{} and \tracinCP{} avoid many of the primary pitfalls associated with static, gradient-based estimators.

First, recall from Section~\ref{sec:Estimators:GradientBased:Static:IF} that Hessian-vector product~$\infFuncSTest$ significantly increases the computational overhead and potential inaccuracy of influence functions.
\tracin{}'s theoretical simplicity avoids the need to compute any Hessian.

Second, representer point's theoretical formulation necessitated considering only a model's final linear layer,
at the risk of (significantly) worse performance.
\tracin{} has the flexibility to use only the final linear layer for scenarios where that provides sufficient accuracy%
\footnote{%
  Last layer only \tracin{} is also referred to as \tracinLast{}.
  \citet{Yeh:2022:FirstBetterLast} evaluate \tracinLast{}'s effectiveness.%
}
as well as the option to use the full model gradient when needed.

Third, by measuring influence during the training process, \tracin{} requires no assumptions about stationarity or convergence.
In fact, \tracin{} can be applied to a model that is only partially trained.
\tracin{} can also be used to study \textit{when} during training an instance is most influential.
For example, \tracin{} can identify whether a training instance is most influential early or late in training.

Fourth, due to how gradient-based methods estimate influence, highly influential instances can actually appear \textit{uninfluential} at the end of training.
Unlike static estimators, dynamic methods like \tracin{} may still be able to detect these instances.
See Section~\ref{sec:Estimators:GradientBased:TradeOff} for more details.

In terms of weaknesses,
\tracin{}'s theoretical motivation assumes stochastic gradient descent without momentum.
However, momentum and adaptive optimization (e.g., Adam~\citep{Kingma:2015:Adam}) significantly accelerate model convergence~\citep{Qian:1999:MomentumGradientDescent,Duchi:2011:Adagrad,Kingma:2015:Adam}.
To align more closely with these sophisticated optimizers, Eq.~\eqref{eq:Estimators:GradientBased:TracIn} and Alg.~\ref{alg:GradientBased:Dynamic:TracIn} would need to change significantly.
For context, Section~\ref{sec:Estimators:GradientBased:Dynamic:HyDRA} details another dynamic estimator, \hydra{}, which incorporates support for just momentum with the resulting increase in estimator complexity substantial.

\paragraph{Related Methods}

\tracin{} has been adapted by numerous derivative/heuristic variants.
For example, \kw{\tracinLast{}} is identical to vanilla \tracin{} except gradient vectors $\gradI$ and $\gradTe$ only consider the model's final linear layer~\citep{Pruthi:2020:TracIn}.
This can make \tracin{} significantly faster at the risk of (significantly) worse accuracy~\citep{Yeh:2022:FirstBetterLast}.

\relatedHeading{\tracin{} for Language Models}
As a counter to the disadvantages of solely considering a model's last layer,%
\footnote{See Section~\ref{sec:Estimators:GradientBased:Static:RepresenterPoint:StrengthWeaknesses} for an extended discussion of last-layer similarity.}
\tracin{}'s authors subsequently proposed
\kw{\tracin{} word embeddings}~(\tracinWE), which targets large language models and considers only the gradients in those models' word embedding layer~\citep{Yeh:2022:FirstBetterLast}.
Since language-model word embeddings can still be very large (e.g.,~BERT\=/Base's word embedding layer has 23M~parameters~\citep{Devlin:2019:Bert}), the authors specifically use the gradients of only those tokens that appear in both training instance~$\zI$ and test instance~$\zTe$.

\relatedHeading{Low Dimensional \tracin{}}
\citet{Pruthi:2020:TracIn} also propose \kw{\tracin{} Random Projection}~(\tracinRP{}) -- a low-memory version of \tracin{} that provides unbiased estimates of~$\infEstTracIn$ (i.e.,~an estimate of an estimate).
Intuitively, \tracinRP{} maps gradient vectors into a $\tracinRpDim$\=/dimensional subspace (${\tracinRpDim \ll \dimW}$) via multiplication by a ${\tracinRpDim \times \dimW}$ random matrix where each entry is sampled i.i.d.\ from Gaussian distribution~\eqsmall{${\mathcal{N}\big(0,\frac{1}{\tracinRpDim}\big)}$}.
These low-memory gradient ``sketches'' are used in place of the full gradient vectors in Eq.~\eqref{eq:Estimators:GradientBased:TracIn} \citep{Woodruff:2014:SketchingLinearAlgebra}.
\tracinRP{} is primarily targeted at applications where $\dimW$ is sufficiently large that storing the full training set's gradient vectors~(${\forall_{\itr,\trIdx}~\gradI}$) is prohibitive.

\relatedHeading{\tracin{} for Generative Models}
\tracin{} has also been used outside of supervised settings.
For example, \citet{Kong:2021:VaeTracIn} apply \tracin{} to unsupervised learning, in particular density estimation;
they
propose \kw{variational autoencoder \tracin{}}~(\vaeTracIn{}), which quantifies the \tracin{} influence in $\beta$\=/VAEs~\citep{Higgins:2017:BetaVAE}.
Moreover,
\citepos{Thimonier:2022:TracInAD} \kw{\tracin{} anomaly detector}~(\tracinAD) functionally estimates the distribution of influence estimates -- using either \tracinCP{} or \vaeTracIn{}.
\tracinAD{} then marks as anomalous any test instance in the tail of this ``influence distribution''.

Note also that \tracin{} can be applied to any iterative, gradient-based model, including those that are non-parametric.
For example, \citepos{Brophy:2023:TreeInfluence} \kw{\boostin{}} adapts \tracin{} for gradient-boosted decision tree ensembles.

\subsubsection{\hydra{} -- Hypergradient Data Relevance Analysis}%
\label{sec:Estimators:GradientBased:Dynamic:HyDRA}

Unlike \tracin{} which uses a novel definition of influence~\eqref{eq:Estimators:GradientBased:Dynamic:TracInIdeal},
\citepos{Chen:2021:Hydra} \kw{hypergradient data relevance analysis}~(\hydra) estimates the leave\=/one\=/out influence~\eqref{eq:Estimators:RetrainBased:LOO}.
\hydra{} leverages the same Taylor series-based analysis as \citepos{Koh:2017:Understanding} influence functions.
The key difference is that \hydra{} addresses a fundamental mismatch between influence functions' assumptions and deep models.

Section~\ref{sec:Estimators:GradientBased:Static:IF} explains that influence functions consider infinitesimally perturbing the weight of training sample~$\zI$ by~${\epsI}$.
Recall that the change in $\zTe$'s test risk w.r.t.\ to this infinitesimal perturbation is
\begin{equation}
  \label{eq:Estimators:GradientBased:Static:Hydra:LossDerivEps}
  \frac{d \risk{\zTe}{\wFin}}
       {d \epsI}
  =
  \frac{\partial \risk{\zTe}{\wFin}}
       {\partial \wFin}
  \transpose
  \,
  \frac{d \wFin}
       {d \epsI}
  =
  \frac{\partial \risk{\zTe}{\wFin}}
       {\partial \wFin}
  \transpose
  \,
  \hyperGradFinI
\end{equation}
where
\eqsmall{${\hyperGradFinI \defeq \frac{d \wFinEpsI} {d \epsI}}$}
denotes the $\dimW$\=/dimensional \kw{hypergradient} of training instance~$\trIdx$ at the end of training.

\citepos{Koh:2017:Understanding} assumptions of differentiability and strict convexity mean that Eq.~\eqref{eq:Estimators:GradientBased:Static:Hydra:LossDerivEps} has a closed form.
However, deep neural models are not convex.
Under non-convex gradient descent without momentum and with \lTwo{}~regularization, ${\wT \defeq \wTOne - \lrT \gradTOne}$ where gradient%
\begin{equation}\label{eq:Estimators:GradientBased:Dynamic:HyDRA:Grad}
  \gradTOne
  \defeq
    \gradW
      \riskBatchT
    +
    \wdecay \wTOne
  \text{.}
\end{equation}
\noindent%
The exact definition of gradient~$\gradTOne$ depends on the specific contents of batch~$\batchT$ so for simplicity, we encapsulate the batch's contribution to the gradient using catch-all term~${\gradW \riskBatchT}$.

Using Eq.~\eqref{eq:Estimators:GradientBased:Dynamic:HyDRA:Grad},
hypergradient~\eqsmall{$\hyperGradFinI$} can be defined recursively as
\begin{align}
  \hyperGradFinI
  \defeq
  \frac{d \wFinEpsI}
       {d \epsI}
  &=
  \frac{d }
       {d \epsI}
  \left(
    \wTEpsI[\nItr - 1]
    -
    \lrT \gradTOne
  \right)
  \\
  &=
  \hyperGradTI[\nItr - 1]
  -
  \lrT[\nItr]
  \frac{d }
       {d \epsI}
  \left(
    \gradW
    \riskBatchFinOneEpsI
    +
    \wdecay
    \wTEpsI[\nItr - 1]
  \right)
  \\
  &=
  (1 - \lrT[\nItr] \wdecay) \hyperGradTI[\nItr - 1]
  -
  \lrT
  \,
  \frac{d }
       {d \epsI}
  \,
  \gradW
  \riskBatchFinOneEpsI
  \label{eq:Estimators:GradientBased:Dynamic:Hydra:FinUnrolledHypergrad}
\end{align}
\noindent
The recursive definition of hypergradient~$\hyperGradFinI$ needs to be unrolled all the way back to initial parameters~$\wZero$.

The key takeaway from Eq.~\eqref{eq:Estimators:GradientBased:Dynamic:Hydra:FinUnrolledHypergrad} is that training hypergradients affect the model parameters \textit{throughout all of training}.
By assuming a convex model and loss, \citepos{Koh:2017:Understanding} simplified formulation ignores this very real effect.
As \citet{Chen:2021:Hydra} observe, hypergradients often cause non-convex models to converge to a vastly different risk minimizer.
By considering the hypergradients' cumulative effect, \hydra{} can provide more accurate \loo{} estimates than influence functions on non-convex models -- albeit via a significantly more complicated and computationally expensive formulation.

\noindent
\textbf{Unrolling Gradient Descent Hypergradients}
The exact procedure to unroll \hydra{}'s hypergradient~$\hyperGradFinI$ is non-trivial.
For the interested reader, supplemental Section~\ref{sec:App:HypergradientUnrolling} provides hypergradient unrolling's full derivation for vanilla gradient descent without momentum.
Below, we briefly summarize Section~\ref{sec:App:HypergradientUnrolling}'s important takeaways, and Section~\ref{sec:App:HypergradientUnrolling}'s full derivation can be skipped with minimal loss of understanding.

At each training iteration, hypergradient unrolling requires estimating the risk Hessian of each training instance in~$\batchT$.%
\footnote{%
  For clarity, this is \textit{not} the inverse Hessian $\invHessFin$ used by influence functions.%
}
This significantly slows down \hydra{} (by a factor of about 1,000$\times$~\citep{Chen:2021:Hydra}).
As a workaround, \citet{Chen:2021:Hydra} propose treating these risk Hessians as all zeros, proving that, under mild assumptions, the approximation error of this simplified version of \hydra{} is bounded.
Alg.~\ref{alg:GradientBased:Dynamic:Hydra} shows \kw{\hydra{}'s fast approximation algorithm} without Hessians for vanilla gradient descent.%
\footnote{%
  See \hydra{}'s original paper \citep{Chen:2021:Hydra} for the fast approximation algorithm with momentum.%
}
After calculating the final hypergradient, substituting
\eqsmall{${\hyperGradFinI}$}
into Eq.~\eqref{eq:Estimators:GradientBased:Static:Hydra:LossDerivEps} with \eqsmall{${\epsI = -\frac{1}{\nTr}}$}%
\footnote{%
  ${\epsI = -\frac{1}{\nTr}}$ is equivalent to deleting instance $\zI$ from the training set.
  Influence functions follow the same procedure for~$\epsI$.
  See Eqs.~\eqref{eq:Estimators:GradientBased:Static:LossDerivEps} and~\eqref{eq:Estimators:GradientBased:Static:IF}.%
}
yields training instance~$\zI$'s \kw{\hydra{} pointwise influence estimator}
\begin{equation}
  \labelAndRemember{eq:Estimators:GradientBased:Dynamic:HyDRA}{%
    \baseInfFunc{\infEstHydra}{\zI}{\zTe}
    \defeq
      -
      \frac{1}{\nTr}
      \gradW
      \risk{\zTe}{\wFin}
      \transpose
      \,
      \hyperGradFinI
  }
  \text{.}
\end{equation}

\begin{figure}[t]
  \centering
  \begin{minipage}{\algMiniPageWidth}
    \begin{algorithm}[H]
      {%
        \AlgFontSize%
        \caption{Fast \hydra{} influence estimation for gradient descent without momentum}\label{alg:GradientBased:Dynamic:Hydra}
\begin{algorithmic}[1]
  \algSetStretch%
  \Require Training parameter set~$\trainParams$\algInputDelim{}
      final parameters~$\wFin$\algInputDelim{}
      training set size~$\nTr$\algInputDelim{}
      iteration count~$\nItr$\algInputDelim{}
      batches ${\batchOne, \ldots, \batchFin}$\algInputDelim{}
      learning rates ${\lrOne, \ldots, \lrFin}$\algInputDelim{}
      weight decay~$\wdecay$\algInputDelim{}
      training instance~$\zI$\algInputDelim{}
      and
      test example~$\zTe$

  \Ensure \hydra{} influence estimate~$\baseInfFunc{\infEstHydra}{\zI}{\zTe}$

  \State $\hyperGradZeroI \gets \zeroVec$
          \algcomment{Initialize to zero vector}
  \For{$\itr \gets 1 \textbf{ to } \nItr$}
    \If{${\zI \in \batchT}$}
      \State ${\wTOne \gets \trainParams\sbrack{\itr}}$
      \State $\hyperGradTI \gets (1 - \lrT \wdecay)\hyperGradTOneI - \frac{\lrT\nTr}{\abs{\batchT}}\gradW \risk{\zI}{\wTOne}$
    \Else
      \State $\hyperGradTI \gets (1 - \lrT \wdecay)\hyperGradTOneI$
    \EndIf
  \EndFor
  \State $\infEstSym \gets -\frac{1}{\nTr} \gradW \risk{\zTe}{\wFin} \transpose \, \hyperGradFinI$
          \algcomment{Influence estimate}
  \State \Return$\infEstSym$
\end{algorithmic}
       }%
    \end{algorithm}
  \end{minipage}
\end{figure}
 
\noindent
\textbf{Relating \hydra{} and \tracin{}}
When ${\wdecay = 0}$ or weight decay's effects are ignored (as done by \tracin{}), \hydra{}'s fast approximation for vanilla gradient descent simplifies to
\begin{align}
  \baseInfFunc{\infEstHydra}{\zI}{\zTe}
    &\approx
      -\frac{1}{\nTr}
      \gradW \risk{\zTe}{\wFin}\transpose
        \sum_{\substack{\itr \\ \zI \in \batchT}}
          - \frac{\lrT\nTr}
                 {\abs{\batchT}}
            \gradW \risk{\zI}{\wTOne}
    \\
    &=
      \sum_{\substack{\itr \\ \zI \in \batchT}}
        \frac{\lrT}
             {\abs{\batchT}}
        \gradW \risk{\zTe}{\wFin}\transpose
        \gradW \risk{\zI}{\wTOne}
    \label{eq:Estimators:GradientBased:Hydra:FastNoWeightDecay}
    \\
    &=
      \gradW \risk{\zTe}{\wFin}\transpose
        \sum_{\substack{\itr \\ \zI \in \batchT}}
          \frac{\lrT}
               {\abs{\batchT}}
          \gradW \risk{\zI}{\wTOne}
    \text{.}
\end{align}
\noindent%
Eq.~\eqref{eq:Estimators:GradientBased:Hydra:FastNoWeightDecay} is very similar to \tracin{}'s definition in Eq.~\eqref{eq:Estimators:GradientBased:TracIn}, despite the two methods \textit{estimating different definitions of influence} (\loo{} vs.\ ideal \tracin{}~\eqref{eq:Estimators:GradientBased:Dynamic:TracInIdeal}).
The only difference between~\eqref{eq:Estimators:GradientBased:Hydra:FastNoWeightDecay} and~\eqref{eq:Estimators:GradientBased:TracIn} is that \hydra{} always uses final test gradient ${\gradW \risk{\zTe}{\wFin}}$ while \tracin{} uses each iteration's test gradient ${\gradW \risk{\zTe}{\wTOne}}$

The \textit{key takeaway} is that while theoretically different, \hydra{} and \tracin{} are in practice very similar where \hydra{} can be viewed as trading (incremental) speed for lower precision w.r.t.~$\zTe$.

\paragraph{Time, Space, and Storage Complexity}

When unrolling the $\nItr$~training iterations,
\hydra{}'s fast approximation performs a $\dimW$\=/dimensional (hyper)gradient calculation for each of the $\nTr$~training instances.
If \hydra{}'s full version with Hessian vector products is used, \citepos{Agarwal:2017} Hessian approximation algorithm estimates each HVP in \bigOp{} time and space.
Therefore, the fast and standard versions of \hydra{} both have full time complexity \bigOnpT{} -- same as \tracin{}, albeit with potentially much worse constant factors.

Observe that each hypergradient~$\hyperGradFinI$ only needs to be computed once and can be reused for each test instance.
Therefore, the fast and standard version of \hydra{} have incremental time complexity of just $\nTr$~gradient dot products -- \bigOnp{} complexity total.
This incremental complexity is much faster than \tracin{} and asymptotically equivalent to influence functions.
In practice though, \hydra{}'s incremental cost is much lower than that of influence functions.

Alg.~\ref{alg:GradientBased:Dynamic:Hydra} requires storing vector~$\hyperGradTI$ throughout \hydra{}'s entire unrolling procedure, where each training instance's hypergradient takes \bigOp{} space.
To analyze all training instances simultaneously, \hydra{} requires \bigOnp{} total space.
In contrast, \tracin{} only requires \bigOpPln{} space to analyze all instances simultaneously.
This difference is substantial for large models and training sets.
In cases where the fully-parallelized space complexity is prohibitive, each training instance's hypergradient can be analyzed separately resulting in a reduced space complexity of \bigOp{} for both fast and standard \hydra{}.

Like \tracin{}, \hydra{} requires storing model parameters ${\trainParams \subseteq  \myset{\wZero, \ldots, \wT[\nItr - 1]}}$ making its minimum storage complexity~\bigOpT{}.
Since hypergradients are reused for each test instance, they can be stored to eliminate the need to recalculate them; this introduces an additional storage complexity of~\bigOnp{}.
This makes \hydra{}'s total storage complexity \bigOpTpn{}.

\begin{remark}
  Storing both the training checkpoints and hypergradients is unnecessary.
  Once all hypergradients have been calculated, serialized training parameters~$\trainParams$ are no longer needed and can be discarded.
  Therefore, a more typical storage complexity is~\bigOpT{} or~\bigOnp{} -- both of which are still substantial.
\end{remark}

\paragraph{Strengths and Weaknesses}

\hydra{} and \tracin{} share many of the same strengths.
For example, \hydra{} does not require assumptions of convexity or stationarity.
Moreover, as a dynamic method, \hydra{} may be able to detect influential examples that are missed by static methods -- in particular when those instances have low loss at the end of training (see Section~\ref{sec:Estimators:GradientBased:TradeOff} for more discussion).

\hydra{} also has some advantages over \tracin{}.
For example,
as shown in Alg.~\ref{alg:GradientBased:Dynamic:TracIn}, \tracin{} requires that each \textit{test} instance be retraced through the entire training process.
This significantly increases \tracin{}'s incremental time complexity.
In contrast, \hydra{} only unrolls gradient descent for the \textit{training} instances, i.e.,~not the test instances.
Hypergradient unrolling is a one-time cost for each training instance; this upfront cost is amortized over all test instances.
Once the hypergradients have been calculated, \hydra{} is much faster than \tracin{} -- potentially by orders of magnitude.
In addition,
\hydra{}'s overall design allows it to natively support momentum with few additional changes.
Integrating momentum into \tracin{}, while theoretically possible, requires substantial algorithmic changes and makes \tracin{} substantially more complicated.
This would mitigate a core strength of \tracin{} -- its simplicity.

\hydra{} does have two weaknesses in comparison to \tracin{}.
First, \hydra{}'s standard (i.e.,~non-fast) algorithm requires calculating many HVPs.
Second, \hydra{}'s \bigOnp{} space complexity is much larger than the \bigOpPln{} space complexity of other influence analysis methods (see Table~\ref{tab:InfluenceEstimators:Comparison}).
For large models, this significantly worse space complexity may be prohibitive.

\paragraph{Related Methods}

The method most closely related to \hydra{} is \citepos{Hara:2019:SgdInfluence} \kw{SGD\=/influence}.
Both approaches estimate the leave\=/one\=/out influence by unrolling gradient descent using empirical risk Hessians.
There are, however, a few key differences.
First, unlike \hydra{}, \citeauthor{Hara:2019:SgdInfluence} assume that the model and loss function are convex.
Next, SGD\=/influence primarily applies unrolling to quantify the Cook's distance, \eqsmall{${\wFin - \wFinSub}$}.
To better align their approach with dataset influence, \citeauthor{Hara:2019:SgdInfluence} propose a surrogate (linear) influence estimator which they incrementally update throughout unrolling.
This means the full training process must be unrolled for each test instance individually, significantly increasing SGD-influence's incremental time complexity.

\citet{Terashita:2021:InfluenceGAN} adapt the ideas of SGD-influence to estimate training data influence in generative adversarial networks (GANs).

Although proposed exclusively in the context of influence functions (Sec.~\ref{sec:Estimators:GradientBased:Static:IF:RelatedMethods}), \citepos{Schioppa:2022:ScaledUpInfluenceFunctions} basic approach to scale up influence functions via faster Hessian calculation could similarly be applied to speed up \hydra{}'s standard (non-fast) algorithm.

\subsection{Trade-off between Gradient Magnitude and Direction}%
\label{sec:Estimators:GradientBased:TradeOff}

This section details a limitation common to existing gradient-based influence estimators that can cause these estimators to systematically overlook highly influential (groups of) training instances.

Observe that all gradient-based methods in this section rely on some vector dot product.
For a dot product to be large, one of two criteria must be met:

\noindent
(1)~The vector directions align (i.e.,~have high cosine similarity).
More specifically, for influence analysis, vectors pointing in similar directions are expected to encode similar information.
This is the ideal case.

\noindent
(2)~Either vector has a large magnitude, e.g.,~\eqsmall{$\norm{\gradW \risk{\zSym}{\W}}$}.
Large gradient magnitudes can occur for many reasons, but the most common cause is that the instance is either incorrectly or not confidently predicted.

Across the training set, gradient magnitudes can vary by several orders of magnitude~\citep{Sui:2021:LocalJacobianRepPt}.
To overcome such a magnitude imbalance, training instances that actually influence a specific prediction may need to have orders of magnitude better vector alignment.
In reality, what commonly happens is that incorrectly predicted or abnormal training instances appear highly influential to all test instances~\citep{Sui:2021:LocalJacobianRepPt}.
\citet{Barshan:2020:RelatIF} describe such training instances as \kw{globally influential}.
However, globally influential training instances provide very limited insight into individual model predictions.
As \citeauthor{Barshan:2020:RelatIF} note, \kw{locally influential} training instances are generally much more relevant and insightful when analyzing specific predictions.

\relatedHeading{Relative Influence}
To yield a more semantically meaningful influence ranking, \citeauthor{Barshan:2020:RelatIF} propose the \kw{$\W$\=/relative influence functions estimator}~(\relatif), which normalizes \citepos{Koh:2017:Understanding} influence functions' estimator by HVP magnitude
\eqsmall{${\big\lVert \relatIfTerm \big\rVert }$}.%
\footnote{%
  Note that this HVP is different than ${ \infFuncSTest \defeq \invHessFin \gradW \risk{\zTe}{\wFin} }$ in Eq.~\eqref{eq:Estimators:GradientBased:Static:IF:STest}.%
}
Formally,
\begin{equation}%
  \labelAndRemember{eq:Estimators:GradientBased:TradeOff:RelatIF}{%
    \baseInfFunc{\infEstRelatIF}{\zI}{\zTe}
      \defeq
      \frac{\baseInfFunc{\infEstIF}{\zI}{\zTe}}
           {\norm{\relatIfTerm}}
  }
    =
    \frac{1}
         {\nTr}
    \,
    \frac{%
            \gradW \risk{\zTe}{\wFin}\transpose
            \invHessFin
            \gradW \risk{\zI}{\wFin}
         }
         {\norm{\relatIfTerm}}
  \text{.}
\end{equation}
\noindent%
\relatif{}'s normalization inhibits training gradient magnitude
${\big\lVert \gradW \risk{\zI}{\wFin} \big\rVert}$
dominating the influence estimate.

\relatif{}'s biggest limitation is the need to estimate an HVP \textit{for every training instance}.
As discussed in Section~\ref{sec:Estimators:GradientBased:Static:IF:StrengthWeaknesses}, HVP estimation is expensive and often highly inaccurate in deep models.
To work around these issues in their evaluation of \relatif{}, \citeauthor{Barshan:2020:RelatIF} use either very small neural models or just consider a large model's final layer, both of which can be problematic.

\relatedHeading{Renormalized Influence}
\citet{Hammoudeh:2022:GAS} make a similar observation as \citeauthor{Barshan:2020:RelatIF} but motivate it differently.
By the chain rule, gradient vectors decompose as
\begin{equation}\label{eq:Estimators:GradientBased:TradeOff:ChainRule}
  \gradW \risk{\zSym}{\W}
    \defeq
    \frac{\partial \lDecFunc{\X}{\W}{\Y}}
         {\partial \W}
    =
    \frac{\partial \lDecFunc{\X}{\W}{\Y}}
         {\partial \decFunc{\X}{\W}}
    \,
    \frac{\partial \decFunc{\X}{\W}}
         {\partial \W}
  \text{.}
\end{equation}
\noindent
\citeauthor{Hammoudeh:2022:GAS} note that for many common loss functions (e.g.,~squared, binary cross-entropy), loss value \eqsmall{$\lDecFunc{\X}{\W}{\Y}$}
induces a strict ordering over loss norm
\eqsmall{$%
  \big\lVert
    \frac{\partial \lDecFunc{\X}{\W}{\Y}}
         {\partial \decFunc{\X}{\W}}
  \big\rVert%
$}.
\citeauthor{Hammoudeh:2022:GAS} term this phenomenon a \kw{low-loss penalty}, where confidently predicted training instances have smaller gradient magnitudes and by consequence consistently appear uninfluential to gradient-based influence estimators.

To account for the low-loss penalty, \citet{Hammoudeh:2022:GAS} propose \kw{renormalized influence} which replaces all gradient vectors -- both training and test -- in an influence estimator with the corresponding \kw{unit vector}.
Renormalization can be applied to any gradient-based estimator.
For example, \citeauthor{Hammoudeh:2022:GAS} observe that renormalized \tracinCP{}, which they term \kw{gradient aggregated similarity}~(\gas),
\begin{equation}%
  \labelAndRemember{eq:Estimators:GradientBased:TradeOff:GAS}{%
    \baseInfFunc{\infEstGas}{\zI}{\zTe}
    \defeq
    \sum_{\itr \in \subsetItr}
      \lrT
      \,
      \frac{\dotprod{\gradI}{\gradTe}}
           {%
             \norm{\gradI}
             \,
             \norm{\gradTe}%
           }%
  }
  \text{,}
\end{equation}
\noindent
is particularly effective at generating influence rankings.
\citeauthor{Hammoudeh:2022:GAS} also provide a renormalized version of influence functions,%
\begin{equation}%
  \labelAndRemember{eq:Estimators:GradientBased:TradeOff:RenormIF}{%
    \baseInfFunc{\infEstRenormIF}{\zI}{\zTe}
      \defeq
      \frac{\baseInfFunc{\infEstIF}{\zI}{\zTe}}
           {\norm{\gradW \riskIFin}}
  }
    =
    \frac{1}
         {\nTr}
    \,
    \frac{%
            \gradW \risk{\zTe}{\wFin}\transpose
            \invHessFin
            \gradW \risk{\zI}{\wFin}
         }
         {\norm{\gradW \riskIFin}}
  \text{.}
\end{equation}
\noindent
Since renormalized influence functions do not require estimating additional HVPs, it is considerably faster than \relatif{}.
Renormalized influence functions also do not have the additional error associated with estimating \relatif{}'s additional HVPs.%
\footnote{%
  \citet{Hammoudeh:2022:GAS} also provide renormalized versions of representer point and \tracin{}, which are omitted here.%
}

This section should not be interpreted to mean that gradient magnitude is unimportant for influence analysis.
On the contrary, gradient magnitude has a \textit{significant} effect on training.
However, the approximations made by existing influence estimators often overemphasize gradient magnitude leading to influence rankings that are not semantically meaningful.

\begin{remark}
  \citeauthor{Barshan:2020:RelatIF}'s \relatif{} and \citeauthor{Hammoudeh:2022:GAS}'s renormalization do not change the corresponding influence estimators' time and space complexities.
\end{remark}

\section{Applications of Influence Analysis}%
\label{sec:Applications}

\revTwo{%
Section~\ref{sec:EstimatorOverview:RelatedTopics} discusses how a few topics (formally) relate to influence analysis.
This section provides an extended discussion of influence analysis's applications.
Specifically, this section focuses}
on higher-level learning tasks as opposed to the specific application environments where influence analysis has been used including: toxic speech detection~\citep{Han:2020:InfluenceToxicSpeech}, social network graph labeling~\citep{Zhang:2021:ActiveInfluenceGraphs}, user engagement detection~\citep{Liang:2021:UserEngagementDetection}, medical imaging annotation~\citep{Braun:2022:InfluenceMedicalImaging}, etc.

First,
\kw{data cleaning} aims to improve a machine learning model's overall performance by removing ``bad'' training data.
These ``bad'' instances arise due to disparate non-malicious causes including human/algorithmic labeling error, non-representative instances, noisy features, missing features, etc.~\citep{Krishnan:2016:ActiveClean,Liu:2018:FinePruning,Krishnan:2019:AlphaClean}.
Intuitively, ``bad'' training instances are generally anomalous, and their features clash with the feature distribution of typical ``clean'' data~\citep{Wojnowicz:2016:InfluenceSketching}.
In practice, overparameterized neural networks commonly memorize these ``bad'' instances to achieve zero training loss~\citep{Hara:2019:SgdInfluence,Feldman:2020:InfluenceEstimation,Pruthi:2020:TracIn,Thimonier:2022:TracInAD}.
As explained in Section~\ref{sec:EstimatorOverview:AlternativeDefinitions}, memorization can be viewed as the influence of a training instance on itself.
Therefore, influence analysis can be used to detect these highly memorized training instances.
These memorized ``bad'' instances are then either removed from the training data or simply relabeled~\citep{Kong:2022:InfluenceDataRelabeling} and the model retrained.

\kw{Poisoning and backdoor attacks} craft malicious training instances that manipulate a model to align with some attacker objective.
For example, a company may attempt to trick a spam filter so all emails sent by a competitor are erroneously classified as spam~\citep{Shafahi:2018:PoisonFrogs}.
Obviously, only influential (malicious) training instances affect a model's prediction.
Some training set attacks rely on influence analysis to craft better (i.e.,~more influential) poison instances~\citep{Fang:2020:InfluenceFunctionsPoisoning,Jagielski:2021:Subpopulation,Oh:2022:RecommenderInteractionPerturbations}.
Since most training set attacks do not assume the adversary knows training's random seed or even necessarily the target model's architecture, poison instances are crafted to maximize their expected group influence~\citep{Chen:2017:Targeted}.

Influence and memorization analysis have also been used to improve \kw{membership inference attacks}, where the adversary attempts to extract sensitive training data provided only a pretrained (language) model~\citep{Demontis:2019:WhyAttacksTransfer,Cohen:2022:InfluenceMembershipInferenceAttack}.

\kw{Training set attack defenses} detect and mitigate poisoning and backdoor attacks~\citep{Li:2022:Survey}.
Since malicious training instances must be influential to achieve the attacker's objective, defending against adversarial attacks reduces to identifying abnormally influential training instances.
If attackers are constrained in the number of training instances they can insert~\citep{Wallace:2021,You:2023:LlmBackdoor}, the target of a training set attack can be identified by searching for test instances that have a few exceptionally influential training instances~\citep{Hammoudeh:2022:GAS}.
The training set attack mitigation removes these anomalously influential instances from the training data and then retrains the model~\citep{Wang:2019:NeuralCleanse}.
In addition, influence estimation has been applied to the related task of \kw{evasion attack detection}, where the training set is pristine and only test instances are perturbed~\citep{Cohen:2020:DetectInfluenceFunctions}.

\kw{Algorithmic fairness} promotes techniques that enable machine learning models to make decisions free of prejudices and biases based on inherited characteristics such as race, religion, and gender~\citep{Mehrabi:2022:SurveyBiasFairness}.
A classic example of model unfairness is the COMPAS software tool, which estimated the recidivism risk of incarcerated individuals.
COMPAS was shown to be biased against black defendants, falsely flagging them as future criminals at twice the rate of white defendants~\citep{Angwin:2016:CompasPropublica}.
Widespread adoption of algorithmic decision making in domains critical to human safety and well-being is predicated on the public's perception and understanding of the algorithms' inherent ethical principles and fairness~\citep{Awad:2018:MoralMachine}.
Yet, how to quantify the extent to which an algorithm is ``fair'' remains an area of active study~\citep{Dwork:2012:FairnessThroughAwareness,Glymour:2019:MeasuringBiases,Saxena:2019:FairnessFare}.
\citet{Black:2021:LeaveOneOut} propose \kw{leave\=/one\=/out unfairness} as a measure of a prediction's fairness.
Intuitively, when a model's decision (e.g.,~not granting a loan, hiring an employee) is fundamentally changed by the inclusion of a \textit{single} instance in a large training set, such a decision may be viewed as unfair or even capricious.
Leave-one-out influence is therefore useful to measure and improve a model's robustness and fairness.

\kw{Explainability} attempts to make a black-box model's decisions understandable by humans~\citep{Burkart:2021:ExplainableSupervisedLearning}.
Transparent explanations are critical to achieving user trust of and satisfaction with ML systems~\citep{Lim:2009:ExplainationsIntelligibility,Kizilcec:2016:EffectTransperancyTrust,Zhou:2019:EffectInfluenceUserTrust}.
\kw{Example-based explanations} communicate why a model made a particular prediction via visual examples~\citep{Cai:2019,Sui:2021:LocalJacobianRepPt} -- e.g.,~training images --  as social science research has shown that humans can understand complex ideas using only examples~\citep{Renkl:2009:ExampleBasedLearning,Renkl:2014:ExampleBasedLearning}.
Influence estimation can assist in the selection of canonical training instances that are particularly important for a given class in general or a single test prediction specifically.
Similarly, \kw{normative explanations} -- which collectively establish a ``standard'' for a given class~\citep{Cai:2019} -- can be selected from those training instances with the highest average influence on a held-out validation set.
In cases where a test instance is misclassified, influence analysis can identify those training instances that most influenced the misprediction.

\kw{Subsampling} reduces the computational requirements of large datasets by training models using only a subset of the training data~\citep{Ting:2018:OptimalSubsampling}.
Existing work has shown that high-quality training subsets can be created by greedily selecting training instances based on their overall influence~\citep{Khanna:2019:FisherKernels,Wang:2020:UnweightedSubsampling}.
Under mild assumptions, \citet{Wang:2020:UnweightedSubsampling} even show that, in expectation, influence-based subsampling performs at least as well as training on the full training set.

\revOne{%
Annotating unlabeled data can be expensive -- in particular for domains like medical imaging where the annotators must be domain experts~\cite{Braun:2022:InfluenceMedicalImaging}.
Compared to labeling instances u.a.r., \kw{active learning} reduces labeling costs by prioritizing annotation of particularly salient unlabeled data.
In practice, active learning often simplifies to maximizing the \kw{add-one-in influence} where each unlabeled instance's marginal influence must be estimated.
Obviously, retraining for each possible unlabeled instance combination has exponential complexity and is intractable.
Instead, a greedy strategy can be used where the influence of each unlabeled instance is estimated to identify the next candidate to label~\citep{Liu:2021:InfluenceActiveLearning,Jia:2021:ScalabilityVsUtility,Zhang:2021:ActiveInfluenceGraphs}.
}

To enhance the benefit of limited labeled data, influence analysis has been used to create better augmented training data~\citep{Lee:2020:InfluenceAugmentationNetwork,Oh:2021:InfluenceTensorCompletion}.
These influence-guided \kw{data augmentation} methods outperform traditional random augmentations, albeit with a higher computational cost.
\newcommand{\futureHeading}[1]{%
  \noindent
  \textit{#1}:%
}

\section{Future Directions}%
\label{sec:FutureDirections}

The trend of consistently increasing model complexity and opacity will likely continue for the foreseeable future.
Simultaneously, there are increased societal and regulatory demands for algorithmic transparency and explainability.
Influence analysis sits at the nexus of these competing trajectories~\citep{Zhou:2019:EffectInfluenceUserTrust}, which points to the field growing in importance and relevance.
This section identifies important directions we believe influence analysis research should take going forward.

\futureHeading{Emphasizing Group Influence over Pointwise Influence}
Most existing methods target pointwise influence, which apportions credit for a prediction to training instances individually.
However, for overparameterized models trained on large datasets, only the tails of the data distribution are heavily influenced by an individual instance~\citep{Feldman:2020:LearningRequiresMemorization}.
Instead, most predictions are moderately influenced by multiple training instances working in concert~\citep{Feldman:2020:InfluenceEstimation,Das:2021:DataSubsetInfluence,Basu:2020:OnSecond}.

As an additional complicating factor, pointwise influence within data-distribution modes is often approximately \kw{supermodular} where the marginal effect of a training instance's deletion \textit{increases} as more instances from a group are removed~\citep{Hammoudeh:2022:GAS}.
This makes pointwise influence a particularly poor choice for understanding most model behavior.
To date, very limited work has systematically studied group influence~\citep{Koh:2019:GroupEffects,Basu:2020:OnSecond,Hammoudeh:2022:GAS}.
Better group influence estimators could be immediately applied in various domains such as poisoning attacks, coreset selection, and model explainability.

\futureHeading{Certified Influence Estimation}
Certified defenses against poisoning and backdoor attacks guarantee that deleting a fixed number of instances from the training data will not change a model's prediction~\citep{Steinhardt:2017,Levine:2021:DPA,Jia:2022:CertifiedKNN,Wang:2022:DeterministicAggregation,Hammoudeh:2023:CertifiedRegression}.
These methods can be viewed as upper bounding the training data's group influence -- albeit very coarsely.
Most certified poisoning defenses achieve their bounds by leveraging ``tricks'' associated with particular model architectures (e.g., instance-based learners~\citep{Jia:2022:CertifiedKNN} and ensembles~\citep{Levine:2021:DPA,Wang:2022:DeterministicAggregation,Hammoudeh:2023:FeaturePartition,Rezaei:2023:RunOffElection}) as opposed to a detailed analysis of a prediction's stability~\citep{Hammoudeh:2023:CertifiedRegression}.
With limited exception~\citep{Jia:2019:KnnShapley}, today's influence estimators do not provide any meaningful guarantee of their accuracy.
Rather, most influence estimates should be viewed as only providing -- at best -- guidance on an instance's ``possible influence.''
Guaranteed or even probabilistic bounds on an instance's influence would enable influence estimation to be applied in settings where more than a ``heuristic approximation'' is required~\citep{Hammoudeh:2023:CertifiedRegression}.

\futureHeading{Improved Scalability}
Influence estimation is slow.
Analyzing each training instance's influence on a single test instance can take several hours or more~\citep{Barshan:2020:RelatIF,Kobayashi:2020:EfficientInfluenceEstimation,Guo:2021:FastIF,Hammoudeh:2022:GAS}.
For influence estimation to be a practical tool, it must be at least an order of magnitude faster.
Heuristic influence analysis speed-ups could prove very useful~\citep{Guo:2021:FastIF,Schioppa:2022:ScaledUpInfluenceFunctions}.
However, the consequences (and limitations) of any empirical shortcuts need to be thoroughly tested, verified, and understood.
Similarly, limited existing work has specialized influence methods to particular model classes~\citep{Jia:2021:ScalabilityVsUtility} or data modalities~\citep{Yeh:2022:FirstBetterLast}.
While application-agnostic influence estimators are useful, their flexibility limits their scalability and accuracy.
Both of these performance metrics may significantly improve via increased influence estimator specialization.

\futureHeading{Surrogate Influence and Influence Transferability}
An underexplored opportunity to improve influence analysis lies in the use of surrogate models~\citep{Sharchilev:2018:TreeInfluence,Jia:2019:TowardsEfficientShapley,Jia:2021:ScalabilityVsUtility,Brophy:2023:TreeInfluence}.
For example, linear surrogates have proven quite useful for model explainability~\citep{Lundberg:2017:SHAP}.
While using only a model's linear layer as a surrogate may be ``too reductive''~\citep{Yeh:2022:FirstBetterLast}, it remains an open question whether other compact models remain an option.
Any surrogate method must be accompanied by rigorous empirical evaluation to identify any risks and ``blind spots'' the surrogate may introduce~\citep{Rudin:2018:StopExplainingBlackBox}.

\futureHeading{Increased Evaluation Diversity}
Influence analysis has the capability to provide salient insights into why models behave as they do~\citep{Feldman:2020:InfluenceEstimation}.
As an example, \citet{Black:2021:LeaveOneOut} demonstrate how influence analysis can identify potential unfairness in an algorithmic decision.
However, influence estimation evaluation is too often superficial and focuses on a very small subset of possible applications.
For instance, most influence estimation evaluation focuses primarily on contrived data cleaning and mislabeled training data experiments~\citep{Wojnowicz:2016:InfluenceSketching,Koh:2017:Understanding,Khanna:2019:FisherKernels,Ghorbani:2019:DataShapley,Yeh:2018:Representer,Pruthi:2020:TracIn,Chen:2021:Hydra,Terashita:2021:InfluenceGAN,K:2021:RevisitingInfluentialExamples,Sui:2021:LocalJacobianRepPt,Brophy:2023:TreeInfluence,Yeh:2022:FirstBetterLast,Kong:2022:InfluenceDataRelabeling,Kwon:2022:BetaShapley}.
It is unclear how these experiments translate into real-world or adversarial settings, with recent work pointing to generalization fragility~\citep{Basu:2021:InfluenceFunctionsFragile,Bae:2022:InfluenceFunctionsQuestion,Schioppa:2023:TheoreticalPracticalInfFunc}.
We question whether these data cleaning experiments -- where specialized methods already exist~\citep{Krishnan:2016:ActiveClean,Krishnan:2019:AlphaClean,Wang:2019:NeuralCleanse} -- adequately satisfy influence analysis's stated promise of providing ``understanding [of] black-box predictions''~\citep{Koh:2017:Understanding}.

\futureHeading{Objective Over Subjective Evaluation Criteria}
A common trope when evaluating an influence analysis method is to provide a test example and display training instances the estimator identified as most similar or dissimilar.
These ``eye test'' evaluations are generally applied to vision datasets~\citep{Koh:2017:Understanding,Yeh:2018:Representer,Jia:2019:KnnShapley,Pruthi:2020:TracIn,Feldman:2020:InfluenceEstimation} and to a limited extent other modalities.
Such experiments are unscientific.
They provide limited meaningful insight given the lack of a ground truth by which to judge the results.
Most readers do not have detailed enough knowledge of a dataset to know whether the selected instances are especially representative. %
Rather, there may exist numerous training instances that are much more similar to the target that the influence estimator overlooked.
Moreover, such visual assessments are known to be susceptible to confirmation and expectancy biases~\citep{Mahoney:1977:PublicationPrejudices,Nakhaeizadeh:2013:ConfirmationBias,Kassin:2013:ForensicConfirmationBias}.

Influence analysis evaluation should focus on experiments that are quantifiable and verifiable w.r.t.\ a ground truth.
\section{Conclusions}\label{sec:Conclusions}

While influence analysis has received increased attention in recent years, significant progress remains to be made.
Influence estimation is computationally expensive and can be prone to inaccuracy.
Going forward, fast certified influence estimators are needed.
Nonetheless, despite these shortcomings, existing applications already demonstrate influence estimation's capabilities and promise.

This work reviews numerous methods with different perspectives on -- and even definitions of -- training data influence.
It would be a mistake to view this diversity of approaches as a negative.
While no single influence analysis method can be applied to all situations, most use cases should have at least one method that fits well.
An obvious consequence then is the need for researchers and practitioners to understand the strengths and limitations of the various methods so as to know which method best fits their individual use case.
This survey is intended to provide that insight from both empirical and theoretical viewpoints.

\section*{Acknowledgments}
  This work was supported by a grant from the Air Force Research Laboratory and the Defense Advanced Research Projects Agency (DARPA) — agreement number FA8750\=/16\=/C\=/0166, subcontract K001892\=/00\=/S05, as well as a second grant from DARPA, agreement number HR00112090135.
\newrefcontext[sorting=nyt]
\renewcommand*{\bibfont}{\small}
\printbibliography%
\newrefcontext[sorting=ynt]

\startcontents  %
\newpage
\onecolumn
\thispagestyle{empty}
\pagenumbering{arabic}%
\renewcommand*{\thepage}{A\arabic{page}}
\appendix
\SupplementaryMaterialsTitle{}

\begin{center}
  \textbf{\large Organization of the Appendix}
\end{center}
\printcontents{Appendix}{1}[2]{}

\clearpage
\pagebreak
\section{Nomenclature}%
\label{sec:App:Nomenclature}

Table~\ref{tab:App:Nomenclature:General} provides a general nomenclature reference that applies throughout this document, including for all influence analysis methods.
Table~\ref{tab:App:Nomenclature:Training} summarizes the nomenclature related to model training.
Table~\ref{tab:App:Nomenclature:MethodParams} details nomenclature symbols that are specific to an individual influence analysis method.

\begin{table}[h]
  \centering
  \caption{General nomenclature reference}%
  \label{tab:App:Nomenclature:General}
  {%
    \nomenclatureTableFontSize%
\begin{tabular}{lp{5.0in}}
  \toprule
  $\setint{r}$      & Set $\myset{1, \ldots, r}$ for arbitrary positive integer~$r$ \\
  ${A \simN{m} B}$  & Set $A$ is a u.a.r.\ subset of size~$m$ from set~$B$ \\
  $\powerSet{A}$    & Power set of~$A$ \\
  $\ind{a}$         & Indicator function where ${\ind{a} = 1}$ if predicate~$a$ is true and 0~otherwise \\
  $\zeroVec$        & Zero vector \\
  $\X$              & Feature vector \\
  $\domainX$        & Feature domain where ${\domainX \subseteq \real^{\dimX}}$ and ${\forall_{\X} \, \X \in \domainX}$ \\
  $\dimX$           & Feature dimension where ${\dimX \defeq \abs{\X}}$ \\
  $\Y$              & Dependent/target value, e.g.,~label \\
  $\domainY$        & Dependent value domain, i.e., ${\forall_{\Y} \, \Y \in \domainY}$.  Generally ${\domainY \subseteq \real}$ \\
  $\zSym$           & Feature vector-dependent value tuple where ${\zSym \defeq (\X,\Y)}$ \\
  $\domainZ$        & Instance domain where ${\domainZ \defeq \domainX \times \domainY}$ and ${\forall_{\zSym} \, \zSym \in \domainZ}$ \\
  $\trainDataDist$  & Instance data distribution where $\func{\trainDataDist}{\domainZ}{\realnn}$ \\
  $\trainSet$       & Training set where ${\trainSet \defeq \myset{\zI}_{\trIdx = 1}^{\nTr}}$  \\
  $\nTr$            & Size of the training set where ${\nTr \defeq \abs{\trainSet}}$ \\
  $\trainSetAlt$    & Arbitrary training subset where ${\trainSetAlt \subseteq \trainSet}$ \\
  $\trIdx$          & Arbitrary training example index where ${\trIdx \in \setint{\nTr}}$ \\
  $\zTe$            & Arbitrary test instance where ${\zTe \defeq (\xTe, \yTe)}$ \\
  $\dec$            & Model where $\func{\dec}{\domainX}{\domainY}$ \\
  $\W$              & Model parameters where ${\W \in \real^{\dimW}}$ \\
  $\dimW$           & Parameter dimension where ${\dimW \defeq \abs{\W}}$ \\
  $\loss$           & Loss function where $\func{\loss}{\domainY \times \domainY}{\real}$ \\
  $\risk{\zSym}{\W}$
                    & Empirical risk of example ${\zSym = (\X,\Y)}$ w.r.t.\ $\W$, where ${\risk{\zSym}{\W} \defeq \lDecFunc{\X}{\W}{\Y}}$\\
  $\infFunc{\zI}{\zTe}$
                    & Exact pointwise influence of training instance $\zI$ on test instance $\zTe$ \\
  $\baseInfFunc{\infEstSym}{\zI}{\zTe}$
                    & Estimate of training instance $\zI$'s pointwise influence on test instance~$\zTe$ \\
  $\infFunc{\trainSetAlt}{\zTe}$
                    & Group influence of training subset ${\trainSetAlt \subseteq \trainSet}$ on test instance $\zTe$ \\
  $\baseInfFunc{\infEstSym}{\trainSetAlt}{\zTe}$
                    & Estimate of the group influence of training subset ${\trainSetAlt \subseteq \trainSet}$ on test instance $\zTe$ \\
  $\trainSetCore$   & A coreset (Sec.~\ref{sec:EstimatorOverview:RelatedTopics}) \\
  \bottomrule
\end{tabular}
   }%
\end{table}

\begin{table}[ht]
  \centering
  \caption{%
    Training related nomenclature reference.%
  }%
  \label{tab:App:Nomenclature:Training}
  {%
    \nomenclatureTableFontSize%
\renewcommand{\arraystretch}{1.25}

\begin{tabular}{lp{5.0in}}
  \toprule
  $\nItr$           & Number of training iterations \\
  $\itr$            & Training iteration number where ${\itr \in \myset{0, 1, \ldots, \nItr}}$.  ${\itr = 0}$ denotes initial conditions. \\
  $\wT$             & Model parameters at the end of iteration~$\itr$. \\
  $\wZero$          & Model parameters at the start of training\\
  $\wFin$           & Model parameters at the end of training \\
  $\wOpt$           & Optimal model parameters \\
  $\wFinSubBase{\trainSetAlt}$
                    & Final model parameters trained on training data subset ${\trainSetAlt \subseteq \trainSet}$ \\
  $\wdecay$         & \lTwo{}~regularization (i.e.,~weight decay) hyperparameter \\
  $\batchT$         & (Mini)batch used during training iteration~$\itr$ where ${\batchT \subseteq \trainSet}$ \\
  $\batchSizeTSym$  & Batch size for iteration~$\itr$, where ${\batchSizeTSym \defeq \abs{\batchT}}$ \\
  $\lrT$            & Learning rate at training iteration~$\itr$ \\
  $\trainParams$    & Serialized training parameters where \eqsmall{${\trainParams \subseteq \myset{\wZero, \ldots, \wT[\nItr - 1]}}$} \\
  $\gradW \riskIT$
                    & Training instance $\zI$'s risk gradient for iteration~$\itr$ \\
  $\gradWSq \riskIT$
                    & Training instance $\zI$'s risk Hessian for iteration~$\itr$ \\
  $\hessT$          & Empirical risk Hessian the entire training set where \eqsmall{${\hessT \defeq \frac{1}{\nTr} \sum_{\trIdx = 1}^{\nTr} \gradWSq \riskIT}$} \\
  $\invHessT$       & Inverse of the empirical risk Hessian \\
  \bottomrule
\end{tabular}
   }%
\end{table}

\begin{table}[ht]
  \centering
  \caption{Influence (estimator) specific hyperparameters and nomenclature}%
  \label{tab:App:Nomenclature:MethodParams}
  {%
    \nomenclatureTableFontSize%
\renewcommand{\arraystretch}{1.25}
\setlength{\dashlinedash}{0.4pt}
\setlength{\dashlinegap}{1.5pt}
\setlength{\arrayrulewidth}{0.3pt}

\newcommand{\GrpSep}{\cdashline{1-2}}

\begin{tabular}{lp{5.2in}}
  \toprule
  $\trainSetNoZi$   & Leave-one-out training set where instance~$\zI$ is held out \\
  $\kNeigh$         &
        $\kNeigh$\=/nearest neighbors neighborhood size \\
  $\neighAlt{\xTe}{\trainSetAlt}$    &
        \knn{} neighborhood for test feature vector~$\xTe$ from training set ${\trainSetAlt \subseteq \trainSet}$ \\
  \GrpSep
  $\nFeldModel$     &
        Number of submodels trained by the \feldman{} estimator \\
  $\feldDatasetK$   &
        Training set used by the $\feldIdx$\=/th \feldman{} submodel \\
  $\wFeldK$         &
        Final model parameters used by the $\feldIdx$\=/th \feldman{} submodel \\
  $\nFeldModelI$    &
        Number of \feldman{} submodels trained using $\zI$, where ${\nFeldModelI \defeq \sum_{\feldIdx = 1}^{\nFeldModel} \ind{\zI \in \feldDatasetK} }$ \\
  $\feldDatasetSize$ &
                    \feldman{} submodel training-set size where ${\forall_{\feldIdx}\, \abs{\feldDatasetK} = \feldDatasetSize < \nTr}$ \\
  \GrpSep
  $\shapValSym$     & Shapley value characteristic function $\func{\shapValSym}{\powerSet{A}}{\real}$ for arbitrary set~$A$. \\
  \GrpSep
  $\epsI$           & Training instance~$\trIdx$ weight perturbation \\
  $\wTEpsI$         & Model parameters trained on a training set perturbed by~$\epsI$ \\
  \GrpSep
  $\rpValI$         & Training instance~$\zI$'s representer value, where \eqsmall{${\rpValI \defeq -\frac{1}{\wdecay\nTr} \rpLossDerivI}$} \\
  $\wFinLast$       & Model $\dec$'s final linear layer parameters \\
  $\wFinBegin$      & All model $\dec$'s parameters except the final layer, where ${\wFinBegin := \wFin \setminus \wFinLast}$ \\
  $\featI$          & Training instance~$\zI$'s feature representation input into model $\dec$'s final linear layer \\
  $\featTe$         & Test instance~$\zTe$'s feature representation input into model $\dec$'s final linear layer \\
  $\kernel$         & Kernel (similarity) function between two (feature) vectors \\
  $\regFunc{\W}$    & Regularizer function where $\func{\regularizerSym}{\domainW}{\realnn}$ \\
  \GrpSep
  $\subsetItr$      & Subset of the training iterations considered by \tracinCP{}, where ${\subsetItr \subset \setint{\nItr}}$. \\
  \GrpSep
  $\hyperGradTI$    & Training hypergradient where \eqsmall{${\hyperGradTI \defeq \frac{d \wTEpsI}{d \epsI} \in \domainW}$} \\
  \bottomrule
\end{tabular}
   }%
\end{table}
 
\clearpage
\pagebreak
\section{Influence Analysis Method Definition Reference}%
\label{sec:App:EstimatorReference}

\newcommand{\eqWidth}{5.0in}

\newcommand{\baseInfluenceRow}[2]{%
  {\scriptsize #1}
  &
  \adjustbox{valign=c}{%
    \begin{minipage}[t]{\eqWidth}%
      \vspace{4pt}%
      {%
        \EqFontSize%
        \begin{equation*}%
          #2%
        \end{equation*}%
      }%
    \end{minipage}
  }
  \\
}
\newcommand{\InfluenceRow}[2]{%
  \baseInfluenceRow{#1}%
                   {\recallLabel{#2}}%
}

\newcommand{\DefSep}{\cdashline{1-2}}

\newcolumntype{D}{@{}>{\centering\arraybackslash} m{3.90cm}@{}}

\setlength{\dashlinedash}{0.4pt}%
\setlength{\dashlinegap}{1.5pt}%
\setlength{\arrayrulewidth}{0.3pt}%
\renewcommand*{\arraystretch}{1.25}%

\setlength{\LTcapwidth}{\textwidth}
\begin{longtable}{Dp{5.1in}}%
  \caption{%
      Influence analysis method formal definitions including equation numbers and citations.
  }%
  \label{tab:App:Nomenclature:InfDef}
  \\\toprule
  \endfirsthead
  \caption{%
      Influence analysis method formal definitions including equation numbers \& citations (continued).
  }
  \\\toprule
  \endhead
    \bottomrule
    \multicolumn{2}{r@{}}{(Continued \ldots)}\\
  \endfoot
    \bottomrule
  \endlastfoot
  \InfluenceRow{Memorization \citep{Feldman:2020:InfluenceEstimation,Pruthi:2020:TracIn}}%
               {eq:EstimatorOverview:Memorization}
  \DefSep
  \InfluenceRow{Cook's Distance\linebreak\citep[Eq.~(5)]{Cook:1977:DetectionInfluence}}%
               {eq:EstimatorOverview:CooksDistance:Pointwise}
  \DefSep
  \InfluenceRow{Leave-One-Out Influence \citep{Cook:1982:InfluenceRegression}}%
               {eq:Estimators:RetrainBased:LOO}
  \DefSep
  \InfluenceRow{\feldman{} Influence Estimator \citep{Feldman:2020:InfluenceEstimation}}%
               {eq:Estimators:RetrainBased:Feldman}
  \DefSep
  \baseInfluenceRow{Consistency Score \citep{Jiang:2021:CScore}}%
                   {%
    \cScore
    \defeq
      \expectS{\cTrSize \,\simOne\, \setint{\nTr}}%
              {%
                - \expectS{\trainSetAlt \,\simN{\cTrSize}\, \trainSet}
                          {\risk{\zTe}{\wFinSubBase{\trainSetAlt}}}
              }%
  }
  \DefSep
  \InfluenceRow{Shapley Value Pointwise Influence \citep{Shapley:1953,Ghorbani:2019:DataShapley}}%
               {eq:Estimators:RetrainBased:ShapleyInfluence}
  \DefSep
  \InfluenceRow{Shapley Interaction Index \citep{Grabisch:1999:InteractionAmongPlayers}}%
               {eq:Estimators:RetrainBased:Shapley:InteractionIndex}
  \DefSep
  \InfluenceRow{Shapley-Taylor Interaction Index \citep{Sundararajan:2020:ShapleyTaylorInteraction}}%
               {eq:Estimators:RetrainBased:Shapley:Taylor}
  \DefSep
  \InfluenceRow{$\kNeigh$\=/Nearest Neighbors Shapley Influence \citep{Jia:2019:KnnShapley}}%
               {eq:Estimators:RetrainBased:Shapley:Knn:Influence}
  \DefSep
  \InfluenceRow{Banzhaf Value~\citep{Banzhaf:1965:WeightedVoting}}%
               {eq:Estimators:RetrainBased:Shapley:Banzhaf}
  \DefSep
  \InfluenceRow{Influence Functions Estimator \citep{Koh:2017:Understanding}}%
               {eq:Estimators:GradientBased:Static:IF}
  \DefSep
  \InfluenceRow{Linear Model Representer Point Influence \citep{Yeh:2018:Representer}}%
               {eq:Estimators:GradientBased:Static:RepresenterPoint:LinearIdeal}
  \DefSep
  \InfluenceRow{Representer Point Influence Estimator \citep{Yeh:2018:Representer}}%
               {eq:Estimators:GradientBased:Static:RepresenterPoint:Estimate}
  \DefSep
  \InfluenceRow{\tracin{} Ideal Pointwise Influence \citep{Pruthi:2020:TracIn}}%
               {eq:Estimators:GradientBased:Dynamic:TracInIdeal}
  \DefSep
  \InfluenceRow{\tracin{} Influence Estimator \citep{Pruthi:2020:TracIn}}%
               {eq:Estimators:GradientBased:TracIn}
  \DefSep
  \InfluenceRow{\tracinCP{} Influence Estimator \citep{Pruthi:2020:TracIn}}%
               {eq:Estimators:GradientBased:TracInCP}
  \pagebreak%
  \InfluenceRow{\hydra{} Influence Estimator \citep{Chen:2021:Hydra}}%
               {eq:Estimators:GradientBased:Dynamic:HyDRA}
  \DefSep
  \InfluenceRow{$\W$\=/Relative Influence Estimator \citep{Barshan:2020:RelatIF}}%
               {eq:Estimators:GradientBased:TradeOff:RelatIF}
  \DefSep
  \InfluenceRow{\gas{} Renormalized Influence Estimator \citep{Hammoudeh:2022:GAS}}%
               {eq:Estimators:GradientBased:TradeOff:GAS}
  \DefSep
  \InfluenceRow{Renormalized Influence Functions Estimator \citep{Hammoudeh:2022:GAS}}%
               {eq:Estimators:GradientBased:TradeOff:RenormIF}
\end{longtable}

\begin{table}[ht]
  \centering
  \caption{\textbf{Influence Analysis Method Abbreviations}: Related methods are grouped together as in Figure~\ref{fig:Estimator:Taxonomy}. Each method includes its corresponding source reference.}%
  \label{tab:App:Nomenclature:Methods}
  {%
    \nomenclatureTableFontSize%
\renewcommand{\arraystretch}{1.2}
\setlength{\dashlinedash}{0.4pt}
\setlength{\dashlinegap}{1.5pt}
\setlength{\arrayrulewidth}{0.3pt}

\newcommand{\GrpSep}{\cdashline{1-2}}

\begin{tabular}{p{1.8in}p{3.85in}}
  \toprule
  \loo{}            & Leave-One-Out Influence~\citep{Cook:1982:InfluenceRegression,Koh:2017:Understanding} \\
  \knnLOO{}         & $\kNeigh$\=/Nearest-Neighbors Leave-One-Out~\citep{Jia:2021:ScalabilityVsUtility} \\
  \leafRefit{}      & Decision Forest Leaf Refitting~\citep{Sharchilev:2018:TreeInfluence} \\
  Influence Sketching
                    & Least Squares Influence Sketching~\citep{Wojnowicz:2016:InfluenceSketching} \\
  \GrpSep
  \feldman{}        & Downsampled Leave-One-Out~\citep{Feldman:2020:InfluenceEstimation} \\
  \cScoreName{}     & Consistency Score~\citep{Jiang:2021:CScore} \\
  Generative \feldman{}
                    & \feldman{} for Generative Density Models~\citep{Van:2021:Memorization} \\
  \GrpSep
  \sv{}             & Shapley Value~\citep{Shapley:1953} \\
  Interaction Index & Shapley Interaction Index~\citep{Grabisch:1999:InteractionAmongPlayers} \\
  Shapley-Taylor    & Shapley-Taylor Interaction Index~\citep{Sundararajan:2020:ShapleyTaylorInteraction} \\
  \mcShap{}         & Truncated Monte Carlo Shapley~\citep{Ghorbani:2019:DataShapley} \\
  \gShap{}          & Gradient Shapley~\citep{Ghorbani:2019:DataShapley} \\
  \knnShap{}        & $\kNeigh$\=/Nearest-Neighbors Shapley~\citep{Jia:2019:KnnShapley} \\
  \betaShap{}       & Beta Distribution-Weighted Shapley Value~\citep{Kwon:2022:BetaShapley} \\
  Banzhaf Value     & Banzhaf Value~\citep{Banzhaf:1965:WeightedVoting,Wang:2023:DataBanzhaf} \\
  \ame{}            & Average Marginal Effect~\citep{Lin:2022:InfluenceRandomizedExperiments} \\
  \SHAP{}           & \underline{Sh}apley \underline{A}dditive Ex\underline{p}lanations~\citep{Lundberg:2017:SHAP} \\
  Neuron Shapley    & Shapley Value-Based Neural Explanations~\citep{Ghorbani:2020:NeuronShapley} \\
  \GrpSep
  IF                & Influence Functions~\citep{Koh:2017:Understanding} \\
  \fastif{}         & Fast Influence Functions~\citep{Guo:2021:FastIF} \\
  Arnoldi IF        & Arnoldi-Based Influence Functions~\citep{Schioppa:2022:ScaledUpInfluenceFunctions} \\
  \leafInfluence{}  & Decision Forest Leaf Influence~\citep{Sharchilev:2018:TreeInfluence} \\
  Group~IF          & Group Influence Functions~\citep{Koh:2019:GroupEffects} \\
  Second\=/Order~IF & Second-Order Group Influence Functions~\citep{Basu:2020:OnSecond} \\
  \relatif{}        & Relative Influence (Functions)~\citep{Barshan:2020:RelatIF} \\
  Renorm.\ IF       & Renormalized Influence Functions~\citep{Hammoudeh:2022:GAS} \\
  \GrpSep
  RP                & Representer Point~\citep{Yeh:2018:Representer} \\
  High Dim.\ Rep.   & High-Dimensional Representers~\citep{Tsai:2023:HighDimRepPt} \\
  \rpLocalJacobian  & Representer Point Based on Local Jacobian Expansion~\citep{Sui:2021:LocalJacobianRepPt} \\
  \trex{}           & \underline{T}ree-Ensemble \underline{Re}presenter-Point E\underline{x}planations~\citep{Brophy:2023:TreeInfluence} \\
  \GrpSep
  \tracin{}         & Traced Gradient Descent Influence~\citep{Pruthi:2020:TracIn} \\
  \tracinCP{}       & \tracin{} Checkpoint~\citep{Pruthi:2020:TracIn} \\
  \tracinRP{}       & \tracin{} Random Projection~\citep{Pruthi:2020:TracIn} \\
  \tracinLast{}     & \tracin{} Last Layer Only~\citep{Pruthi:2020:TracIn,Yeh:2022:FirstBetterLast} \\
  \vaeTracIn{}      & Variational Autoencoder \tracin{}~\citep{Kong:2021:VaeTracIn} \\
  \tracinAD{}       & \tracin{} Anomaly Detection~\citep{Thimonier:2022:TracInAD} \\
  \tracinWE{}       & \tracin{} Word Embeddings~\citep{Yeh:2022:FirstBetterLast} \\
  \boostin{}        & Boosted (Tree) Influence~\citep{Brophy:2023:TreeInfluence} \\
  \gas{}            & Gradient Aggregated Similarity~\citep{Hammoudeh:2021:Simple,Hammoudeh:2022:GAS}   \\
  \GrpSep
  \hydra{}          & Hypergradient Data Relevance Analysis~\citep{Chen:2021:Hydra} \\
  SGD\=/Influence   & Stochastic Gradient Descent Influence~\citep{Hara:2019:SgdInfluence} \\
  \bottomrule
\end{tabular}
   }%
\end{table}
 
\clearpage
\pagebreak
\newcommand{\riskBatchTOneUnroll}{\riskBatchTOne}
\newcommand{\wTOneUnroll}{\wTOne}

\section{Unrolling Gradient Descent Hypergradients}%
\label{sec:App:HypergradientUnrolling}

This section provides a formal derivation of how to unroll \hydra{}'s training hypergradients.
We provide this reference for the interested reader to understand unrolling's complexity.\footnote{%
  Similar complexity would be required if \tracin{}~\citep{Pruthi:2020:TracIn} were extended to support momentum or adaptive optimization.%
}
Readers do not need to understand this section's details to understand how \hydra{} relates to other influence analysis methods.

Recall that Eq.~\eqref{eq:Estimators:GradientBased:Dynamic:Hydra:FinUnrolledHypergrad} does not describe how the exact contents of batch~$\batchFin$ affect unrolling.
Eq.~\eqref{eq:Estimators:GradientBased:Static:IF:ErmPerturb} defines the effect that infinitesimally perturbing the weight of training instance~$\zI$ has on a model's empirical risk minimizer.
Formally, the perturbed empirical risk is
\begin{equation}%
  \label{eq:Estimators:GradientBased:Dynamic:Hydra:TotalRisk}
  \risk{\trainSet}%
       {\W}
  \defeq
    \frac{1}%
         {\nTr}
    \sum_{\zSym \in \trainSet}
        \risk{\zSym}%
             {\W}
    +
    \epsI
      \risk{\zI}%
           {\W}
  \text{,}
\end{equation}
\noindent
Observe that ${\epsI = -\frac{1}{\nTr}}$ is the same as removing instance $\zI$ from the training set.

We now extend this idea to the effect of $\epsI$ on a single minibatch.
For any iteration ${\itr \in \setint{\nItr}}$,
Formally, batch~$\batchT$'s risk under a training set perturbation by $\epsI$ is
\begin{equation}\label{eq:Estimators:GradientBased:Dynamic:Hydra:BatchRisk}
  \riskBatchTOneUnroll
  \defeq
    \frac{1}%
         {\abs{\batchT}}
    \sum_{\zSym \in \batchT}
      \risk{\zSym}%
           {\wTOneUnroll}
    +
    \ind{\zI \in \batchT}
    \left(
      \frac{\nTr\epsI}%
           {\abs{\batchT}}
      \,
      \risk{\zI}%
           {\wTOneUnroll}
    \right)
  \text{,}
\end{equation}
\noindent
where indicator function~$\ind{\zI \in \batchT}$ checks whether instance~$\zI$ is in batch~$\batchT$.
Observe that
$\risk{\zI}{\wTOneUnroll}$
is scaled by $\nTr$.
Without this multiplicative factor, then when ${\epsI = -\frac{1}{\nTr}}$, $\zI$'s effect is not completely removed from the batch, i.e.,
\begin{equation}
  \frac{1}%
       {\abs{\batchT}}
  \,
  \risk{\zI}%
       {\wTOneUnroll}
  -
  \frac{1}%
  {\nTr\abs{\batchT}}
  \,
  \risk{\zI}%
       {\wTOneUnroll}
  \ne
  0
  \text{.}
\end{equation}

Eq.~\eqref{eq:Estimators:GradientBased:Dynamic:Hydra:BatchRisk}'s gradient w.r.t.\ $\W$ is
\begin{equation}\label{eq:Estimators:GradientBased:Dynamic:Hydra:BatchRiskGrad}
  \gradW
  \riskBatchTOneUnroll
  =
    \frac{1}%
         {\abs{\batchT}}
    \sum_{\zSym \in \batchT}
    \gradW
      \risk{\zSym}{\wTOneUnroll}
    +
    \ind{\zI \in \batchT}
    \left(
      \frac{\nTr\epsI}%
           {\abs{\batchT}}
      \,
      \gradW
      \risk{\zI}{\wTOneUnroll}
    \right)
  \text{.}
\end{equation}
\noindent%
\hydra{}'s hypergradients specify that the derivative is taken w.r.t.\ $\epsI$.
Consider the simpler case first where ${\zI \notin \batchT}$, then
{\small%
  \begin{align}
    \frac{d}
         {d \epsI}
    \sbrack{%
      \gradW
      \riskBatchTOneEpsI
    }
    &=
      \frac{1}
           {\abs{\batchT}}
      \sum_{\zSym \in \batchT}
      \frac{d}
           {d \epsI}
      \,
      \frac{\partial}
           {\partial \W}
          \risk{\zSym}{\wTOneUnroll}
    \\
    &=
      \frac{1}
           {\abs{\batchT}}
      \sum_{\zSym \in \batchT}
      \frac{\partial}
           {\partial \W^2}
          \risk{\zSym}{\wTOneUnroll}
      \frac{d \wT}
           {d \epsI}
    &
    \text{\algcomment{Chain rule}}
    \\
    &=
      \frac{1}%
           {\abs{\batchT}}
      \sum_{\zSym \in \batchT}
      \gradWSq
          \risk{\zSym}{\wTOneUnroll}
      \,
      \hyperGradTOneI
    \text{.}
    \label{eq:Estimators:GradientBased:Dynamic:Hydra:NoZiBatch}
  \end{align}%
}%
\noindent
Note that
\eqsmall{${%
    \gradWSq
  \risk{\zSym}{\wTOneUnroll}
}$}
is training instance $\zSym$'s risk Hessian w.r.t.\ parameters~$\wTOneUnroll$.

When instance $\zI$ is in $\batchT$,
unrolling hypergradient~$\hyperGradTI$
requires an additional term.
We derive that term below with similar analysis as when ${\zI \notin \batchT}$ with the addition of using the product rule.
Observe that Eq.~\eqref{eq:Estimators:GradientBased:Dynamic:Hydra:EpsIRiskDeriv} below considers both $\zI$'s risk gradient and Hessian.
\begin{align}
  \frac{d}%
       {d \epsI}
  \sbrack{%
    \epsI
    \,
    \gradW
    \risk{\zI}{\wTOneUnroll}
  }
  &=
    \gradW
    \risk{\zI}{\wTOneUnroll}
    +
    \epsI
    \frac{d}%
         {d \epsI}
    \gradW
    \risk{\zI}{\wTOneUnroll}
  &
  \text{\algcomment{Product rule}}
  \\
  &=
    \gradW
    \risk{\zI}{\wTOneUnroll}
    +
    \epsI
      \gradWSq
      \risk{\zI}{\wTOneUnroll}
    \hyperGradTOneI
  &
  \text{\algcomment{Chain rule}}
  \label{eq:Estimators:GradientBased:Dynamic:Hydra:EpsIRiskDeriv}
\end{align}

Combining Eqs.~\eqref{eq:Estimators:GradientBased:Dynamic:Hydra:FinUnrolledHypergrad}, \eqref{eq:Estimators:GradientBased:Dynamic:Hydra:NoZiBatch}, and~\eqref{eq:Estimators:GradientBased:Dynamic:Hydra:EpsIRiskDeriv}, the hypergradient update rule for vanilla gradient descent without momentum is
\begin{align}
  \hyperGradTI
  =
  ~
  &
  (1 - \lrT \wdecay)
  \,
  \hyperGradTOneI
  \nonumber
  \\
  &-
  \frac{\lrT}{\abs{\batchT}}
  \sum_{\zSym \in \batchT}
    \gradWSq
    \risk{\zSym}{\wTOneUnroll}
    \,
    \hyperGradTOneI
  \nonumber
  \\
  &- \frac{\lrT\nTr}{\abs{\batchT}}
  \ind{\zI \in \batchT}
  \left(
    \gradW
    \risk{\zI}{\wTOneUnroll}
    +
    \epsI
    \gradWSq
    \risk{\zI}{\wTOneUnroll}
    \,
    \hyperGradTOneI
  \right)
  \label{eq:Estimators:GradientBased:Dynamic:Hydra:GeneralUnrolledHypergrad}
  \text{.}
\end{align}
\noindent
In \citepos{Chen:2021:Hydra} fast approximation of \hydra{}, all Hessians (e.g., \eqsmall{${\gradWSq \risk{\zSym}{\wTOneUnroll}}$}) in Eq.~\eqref{eq:Estimators:GradientBased:Dynamic:Hydra:GeneralUnrolledHypergrad} are treated as zeros and the associated terms dropped.
The resulting simplified equation,
\begin{equation}%
  \label{eq:Estimators:GradientBased:Dynamic:Hydra:SimplifiedUnrolledHypergrad}
  \hyperGradTI
  =
  (1 - \lrT \wdecay)
  \,
  \hyperGradTOneI
  -
  \frac{\lrT\nTr}{\abs{\batchT}}
  \ind{\zI \in \batchT}
    \gradW
    \risk{\zI}{\wTOneUnroll}
  \text{,}
\end{equation}
\noindent
is the basis of the fast-approximation update rule in Alg.~\ref{alg:GradientBased:Dynamic:Hydra}.

  \stopcontents  %

\end{document}